\UseRawInputEncoding
\pdfoutput=1
\documentclass[journal]{IEEEtran}
\usepackage{amsmath,amsfonts}
\usepackage{algorithmic}
\usepackage{algorithm}
\usepackage{array}
\usepackage[colorlinks=True,linkcolor=red,anchorcolor=blue,citecolor=blue]{hyperref}

\usepackage[caption=false,font=footnotesize,labelfont=rm,textfont=rm]{subfig}

\usepackage{textcomp}
\usepackage{stfloats}
\usepackage{url}
\usepackage{verbatim}
\usepackage{graphicx}
\usepackage{cite}
\usepackage{multirow}

\begin{document}

\title{\huge{A Universal Knowledge Embedded Contrastive Learning Framework for Hyperspectral Image Classification}}

\author{Quanwei Liu, Yanni Dong, Tao Huang,~Lefei Zhang, and Bo Du%, \IEEEmembership{Senior member, IEEE}

        % <-this % stops a space
\thanks{
%This study was jointly supported by the National Natural Science Foundation of China under Grants 62222116 and 62171417. (\textit{Corresponding author: Yanni Dong})

Quanwei Liu is with the School of Resource and Environmental Sciences, Wuhan University, Wuhan 430079, China and the College of Science and Engineering, James Cook University, Cairns QLD 4878, Australia  (e-mail: quanwei.liu@my.jcu.edu.au).

Yanni Dong is with the School of Resource and Environmental Sciences, Wuhan University, Wuhan 430079, China (e-mail: dongyanni@whu.edu.cn).

Tao Huang is with the College of Science and Engineering, James Cook University, Cairns QLD 4878, Australia (e-mail: tao.huang1@jcu.edu.au).

Lefei Zhang and Bo Du work with the School of Computer Science, Wuhan University, Wuhan 430079, China (e-mail: zhanglefei@whu.edu.cn; dubo@whu.edu.cn).}% <-this % stops a space
% \thanks{Manuscript received April 19, 2021; revised August 16, 2021.}
}

% The paper headers
\markboth{  }%IEEE TRANSACTIONS ON PATTERN ANALYSIS AND MACHINE INTELLIGENCE}%
{Shell \MakeLowercase{\textit{et al.}}: A Sample Article Using IEEEtran.cls for IEEE Journals}

\IEEEpubid{}
% Remember, if you use this you must call \IEEEpubidadjcol in the second
% column for its text to clear the IEEEpubid mark.

\maketitle

\begin{abstract}

Hyperspectral image (HSI) classification techniques have been intensively studied and a variety of models have been developed. However, these HSI classification models are confined to pocket models and unrealistic ways of dataset partitioning. The former limits the generalization performance of the model and the latter is partitioned leading to inflated model evaluation metrics, which results in plummeting model performance in the real world.  Therefore, we propose a universal knowledge embedded contrastive learning framework (KnowCL) for supervised, unsupervised, and semisupervised HSI classification, which largely closes the gap between HSI classification models between pocket models and standard vision backbones. We present a new HSI processing pipeline in conjunction with a range of data transformation and augmentation techniques that provide diverse data representations and realistic data partitioning. The proposed framework based on this pipeline is compatible with all kinds of backbones and can fully exploit labeled and unlabeled samples with the expected training time. Furthermore, we design a new loss function, which can adaptively fuse the supervised loss and unsupervised loss, enhancing the learning performance. This proposed new classification paradigm shows great potential in exploring for HSI classification technology. The code can be accessed at \url{https://github.com/quanweiliu/KnowCL}.

\end{abstract}
% which largely closed the gap of HSI classification models between pocket models and strandard vision backbones.
% which realizes the transformation of HSI classification models from pocket models to standard models.


\section{Introduction}
\label{introduction}

\IEEEPARstart{T}{he}  motion of electrons within matter produces electromagnetic radiation, which results in the formation of an optical spectrum. Objects typically exhibit distinct spectra, and hyperspectral imaging techniques can precisely capture these spectral variations.\cite{shimoni2019hypersectral}. As a result, hyperspectral images (HSIs) have powerful recognition capability and are applied to agriculture, military, energy, and cultural relics, which makes pixel-level HSI classification a hot research topic \cite{al2021spectral, dian2023zero}. Roughly, these methods are divided into two major categories: traditional machine learning methods and deep learning methods\cite{wu2023fully}.  These data-driven methods can extract different levels of features in line with the labelling of the samples. However, labelling of HSIs is expert-dependent and time-consuming, which leads to the scarcity of labeled HSIs, resulting in the current HSI classification techniques stopping at pocket models and unrealistic ways of data partitioning\cite{gao2022unsupervised}.

To fully exploit the information contained in HSIs, previous work focuses on machine learning models that first extract a series of features of HSIs, such as spectral features, texture features, morphological semantic features, etc. Then the extracted features are fed into classifiers such as $k$-nearest neighbor classifier (KNN) to obtain classification results\cite{sohn2002supervised, benediktsson2005classification, huang2016remote}. However, early machine learning methods relied on handcrafted shallow features and empirical assumptions that seriously hindered the model-building efficiency and generalization performance. 

\begin{figure}[!t]
\centering
\includegraphics[width=3in]{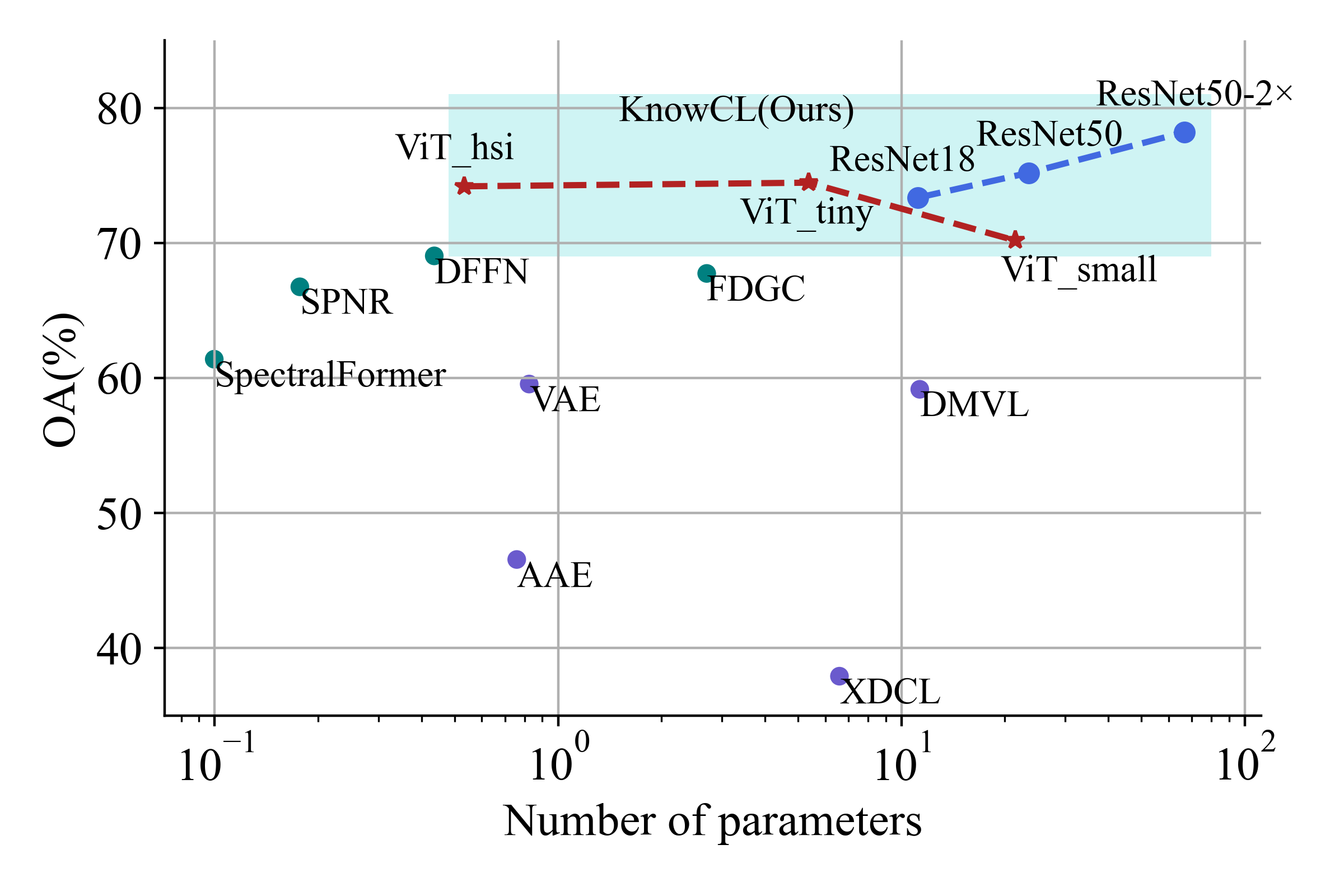}
\caption{Performance of KnowCL on DFC2018 using ViT and ResNet architecture, compared to other unsupervised and supervised baselines.}
\label{fig_1}
\end{figure}
\IEEEpubidadjcol   % 版权生命命令，需要在第一页右栏调用才不会发生文字重叠的现象

Deep learning can extract feature representations end-to-end, which was applied in HSI classification in recent years \cite{cao2018hyperspectral}. Depending on whether labels are involved in feature extraction or not, deep neural networks (DNNs) are categorized as supervised learning, unsupervised learning, and semi-supervised learning \cite{gu2020semi, guo2022deep,hemae}. Minimizing the difference between the labels of the training sample and the ground truth is the learning criterion for supervised learning that often requires a large number of samples with annotations to train the model \cite{zheng2023farseg++}. However, large-scale data labeling, especially for HSIs, is difficult to achieve \cite{gao2022unsupervised}. Unsupervised learning, on the other hand, does not require the involvement of ground truth and it can learn discriminative feature representations from large-scale unlabeled data. Nevertheless, the way training a large amount of unlabeled hyperspectral data will cost a lot of resources. The obtained task-agnostic feature representations often struggle to outperform models dedicated to supervised learning in specific downstream tasks \cite{zhao2023exploring, chen2020big}. For HSI classification tasks, we need to collect the semantic labels of samples before we predict the categories of other regions by the model. Unsupervised learning generally uses a model of unsupervised pre-training plus supervised fine-tuning, which does not fully utilize the precious label information \cite{chen2021exploring}.

To address the above-mentioned problems, semi-supervised learning methods that can handle labeled and unlabeled data simultaneously have been proposed. Semi-supervised deep learning is mainly implemented with the help of complex data processing and classical neural network models \cite{wu2017semi, dong2022weighted, liu2020cnn}. Traditional semi-supervised learning algorithms are acquired to feed the entire image into the network. Although global features can be adequately extracted, their memory footprint and computational complexity grow exponentially with the increase of data size \cite{9785802}. Especially for HSIs consisting of tens of thousands of bands, these methods can only validate the model on some small datasets and cannot be applied to relatively large-size HSIs \cite{dong2022weighted}. More importantly, due to the limited availability of benchmark data and the high cost of labeling, the number of parameters of the aforementioned HSI classification models has been much smaller than that of the conventional image classification models, as shown in Fig. \ref{fig_1}. The gap has been limiting the development of HSI classification technology.

In addition to the learning strategies, another critical step is the evaluation of HSI classification methods, such as benchmark datasets, sampling strategies and metrics, where sampling strategies have become one of the most urgent methods to be revised \cite{audebert2019deep}. A common sampling setting in the HSI classification is to randomly select a fixed number or a fixed proportion of labeled samples from an HSI as a training set and the rest of the samples in this image as a test set \cite{liang2016sampling}.  These selected samples are randomly distributed throughout the image. The spatial autocorrelation of the land cover makes similar features share the same label with a high probability \cite{friedl2000note}. Consequently, the random sampling based on pixel-by-pixel is very inefficient and impractical. In a real environment, most regions rather than pixels are usually identified as ground truth by visual interpretation or field survey. On the other hand, with the development of HSI classification based on CNNs, spectral-spatial classification becomes a concise and efficient method. However, the classification accuracy is appreciably high due to the spatial autocorrelation in the test data set. A considerable number of studies have achieved nearly 100\% correctness in conditions where only a small number of training samples are used \cite{hong2021spectralformer, zheng2020fpga, zhang2021spectral, zhang2022cross}. Cao et. al. \cite{cao2021nonoverlapped} conducted a preliminary performance comparison of HSI classification based on the overlapped and nonoverlapped sampling strategy, where the experimental results also showed that the classification accuracy is significantly reduction using the nonoverlapped sampling strategy.

\begin{figure*}[!t]
\centering
\includegraphics[width=6.5in]{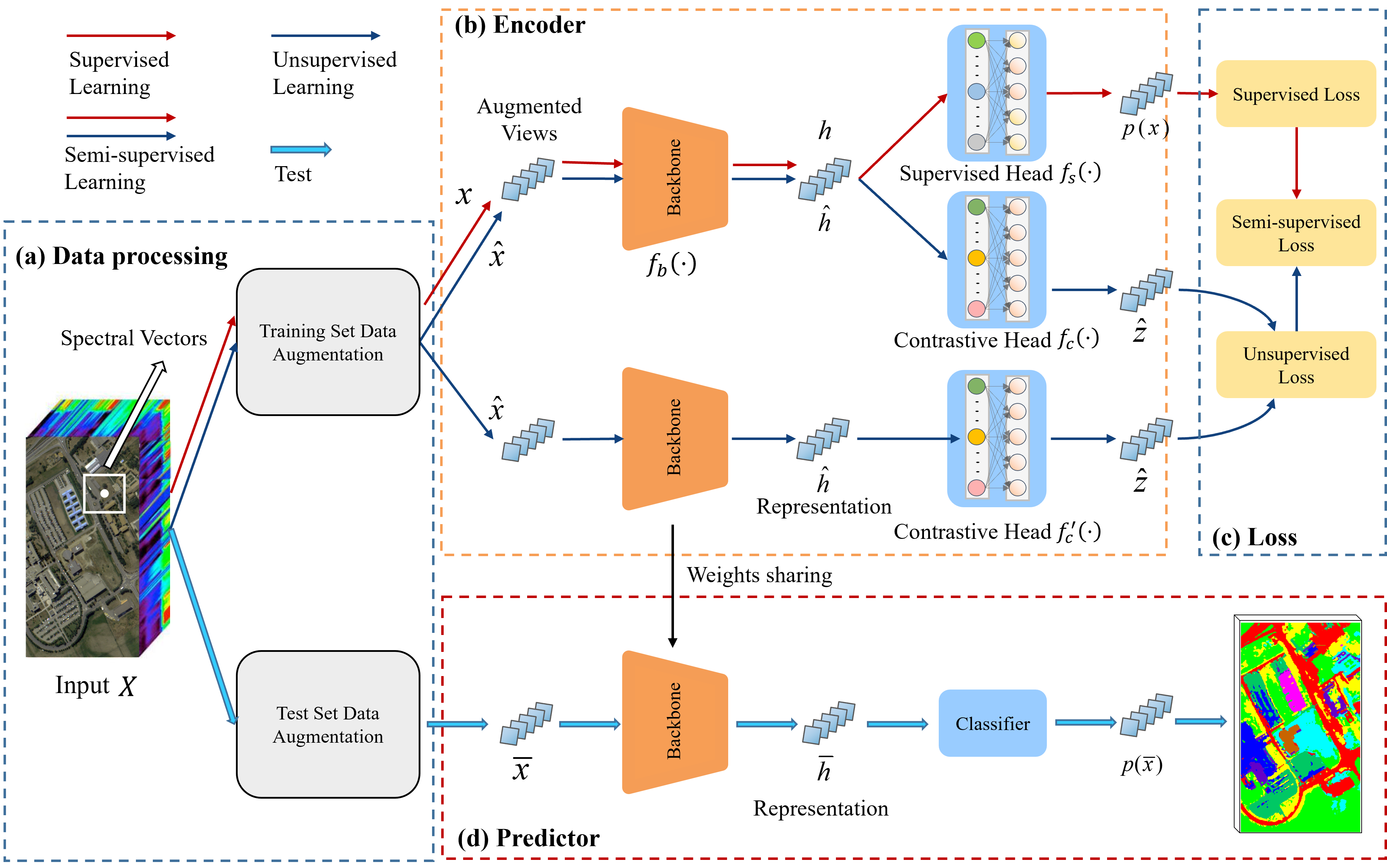}
\caption{The KnowCL framework for HSI classification. The framework is fed with the labeled samples patch $x$, and unlabeled samples patch $\hat x$ in supervised training and unsupervised training, respectively. For semi-supervised training, $x$ and $\hat x$ are fed jointly into the network.}
\label{framework}
\end{figure*}

Given the problems with modeling and evaluation, in this article, we propose a knowledge-embedded contrastive learning framework (KnowCL) for supervised, unsupervised, and semi-supervised HSI classification. The framework diagram is shown in Fig. \ref{framework}. First, we adopt a disjoint approach to divide the HSIs into training and test sets, and this sampling strategy more realistically reproduces the labeling of HSIs. Then, we obtain a diversity of labeled and unlabeled samples through a novel HSI processing pipeline. The supervised classification model is constructed with an iteration optimization of a backbone and a supervised head. The unsupervised model is optimized to maximize the similarity of the two feature extraction branches. We introduced task-specific knowledge into the unsupervised model to construct a novel semi-supervised learning model, which will greatly speed up the network fitting and enhance the model performance. Moreover, the auto-weighted loss that can adaptively fuse supervised learning and contrastive learning is designed for the framework. In addition, we present a light vision transformer (ViT) and test ResNet as backbones to validate the effectiveness of our framework. The main contributions of this research are listed as follows.
 % we present a light vision transformer (ViT) with only four self-attention blocks, which can achieve fast and accurate classification results with fewer hardware requirements.
 
% 本文的创新点有哪些？
% 数据和划分We 
% \begin{itemize}
\begin{enumerate}

\item We combine multiple data augmentation to construct a novel data processing pipeline for HSI classification that well exploits the spectral-spatial information of HSIs and accurately evaluates the performance of the model in realistic scenarios. 
%%%%%%%%%%%%%% 以前的数据增强直接用到了自然图像上，降到了三维，没有充分利用到光谱信息。 %%%%%%%%%%%%%%%%%%%%%%%%%

% 模型
\item We propose a new framework for HSI classification named knowledge embedded contrastive learning (KnowCL), which is strongly scalable and can unify supervised, unsupervised, and semi-supervised HSI classification into one end-to-end framework.
% 2.We propose a new framework for real-world HSI classification. By introducing the necessary labeling knowledge into CL, we are able to extract global feature representations with comparable efficiency to supervised learning.

% 损失函数
\item We develop a new semi-supervised learning paradigm for HSI classification, which is very useful in practical applications. This framework is achieved by a new loss function based on multi-task learning, designed to adaptively combine supervised and contrastive losses. 
% 3. We developed a new loss function based on multi-task learning, designed to adaptively combine supervised and contrastive losses. This fusion of losses enables us to enhance the learning performance of the model.

% 实验
\item	We qualitatively and quantitatively evaluate the classification performance of the proposed model. Extensive experiments indicate that KnowCL achieves performance levels that are on par with or surpass their opponents in supervised and unsupervised methods. Our semi-supervised KnowCL leads all comparison algorithms by a large margin.
% Through extensive experiments, we demonstrate the distinctive characteristics of KnowCL and its superiority over various state-of-the-art feature extraction methods.

% \end{itemize}
\end{enumerate}

The rest of this article is organized as follows. Section \ref{related works} reviews some related works about supervised learning, semi-supervised learning, and unsupervised learning networks. The proposed KnowCL framework is a detailed description in section \ref{proposed approach}. Section IV elaborates qualitatively and quantitatively evaluates the performance of KnowCL on the benchmark datasets. Finally, Section \ref{conclusion} draws comprehensive conclusions and looks forward to further research directions.

\section{Related Works}
\label{related works}

\subsection{Supervised Learning}
In the typical supervised learning classification paradigm, a backbone network is initially employed to extract features from the data. These extracted features are then mapped to a label probability distribution using a softmax layer. The discrepancy between the predicted label and the ground truth is computed by loss functions such as focal loss and cross-entropy loss. The resulting error value is then backpropagated through the network to optimize its parameters. Over the last few years, lots of supervised learning classification methods based on DL have been proposed and become mainstream in HSI classification\cite{sun2021supervised,zhao2022superpixel}. Some classic backbones play a huge role in the DL, such as recursion neural network (RNN), CNN and transformer\cite{li2018independently, he2016deep, liu2021swin}.

RNN can only model the spectral features of HSIs, but due to the phenomenon of the same object with different spectra and the same spectra with different objects, it is difficult to achieve good classification results by only spectral features \cite{fu2021coded}, so spatial information was introduced into the HSI classification. For instance, DFFN\cite{li2019dff} incorporates spectral-spatial features through global and local feature extraction. The CNN network uses the convolution kernel to scan the image, which can extract the spatial features of the image very efficiently. Especially with the emergence of ResNet \cite{he2016deep}, the skip connection greatly improves the depth of the CNN and enhances the classification performance. However, CNN can only rely on multiple patches to extract the local spectral-spatial information, leading to a deep and complex network structure. Therefore, FDGC \cite{9785802} was proposed to extract long-term information in non-European space with GCN, and Hong et. al. \cite{hong2021spectralformer} presented SpectralFormer with the help of Transformer to conduct HSI classification, which breaks through the limitation of local connectivity in CNN models by self-attention mechanism \cite{vaswani2017attention}. They greatly facilitate the development of DL models.  

% The transformer structure often requires more samples for training to achieve performance beyond that of CNNs.

The supervised learning process of DNNs necessitates a substantial volume of labeled training data in order to effectively adjust the model's parameters. As a result, these models are most suitable for situations where abundant labeled data is available\cite{zheng2023label}. 

\subsection{Unsupervised Learning}

The previous unsupervised learning methods \cite{cao2020unsupervised} extract latent feature representations with the help of image reconstruction. This generation process consumes a lot of resources. As an unsupervised learning method, contrastive learning (CL) has achieved unprecedented performance in computer vision tasks \cite{he2020momentum, grill2020bootstrap, caron2021emerging}. CL only needs to conduct discriminative tasks in the feature space, thus avoiding the expensive computational cost of generating elaborate fake samples. Therefore, CL models become simpler, and the generalization ability is stronger \cite{chen2020simple}.

Contrastive learning centers on making positive sample pairs closer to each other and negative sample pairs farther apart \cite{oord2018representation, tian2020makes}. For the image, sample pairs can be constructed using image augmentation methods (e.g., rotation, flip, blur, etc.). A sample undergoing a series of data augmentations produces two samples as a pair of positive samples, and other samples from the same batch are viewed as negative sample pairs. MoCo \cite{he2020momentum} builds a sequence of stored negative samples to expand the number of negative samples. SimCLR  \cite{chen2020simple} uses 128 TPUs directly to train the model to store large batch-size samples. DMVL\cite{liu2020deep} uses the infoNCE loss function similar to the SimCLR framework to construct the HSI classification model.  XDCL\cite{zhang2022cross} learns domain invariance features in the spectral and spatial domains of HSIs.

 % The disadvantage of this method is that it requires a larger batch size to produce a contrasting effect, which will significantly increase the memory burden.  
 
% Recent works have shown that we can learn unsupervised features without discriminating between images, but directly predicts the output of one view from another view. BYOL \cite{grill2020bootstrap} relies only on positive pairs to minimize the similarity loss between the output of one view from another view. 

Unsupervised learning models can directly learn feature representations of HSIs without labels, and it is suitable for large-scale remote sensing image feature extraction. However, unsupervised model training consumes a lot of time, and the extracted task-agnostic features still need labels for specific task fine-tuning.

% For our knowledge embedded contrastive learning framework, we find that a larger batch size is not better. A smaller batch size is able to learn better weights to optimize the network, and the necessary labels to jointly optimize the network are crucial for the quality of the feature representation.

\begin{figure*}[!t]
\centering
\includegraphics[width=6.5in]{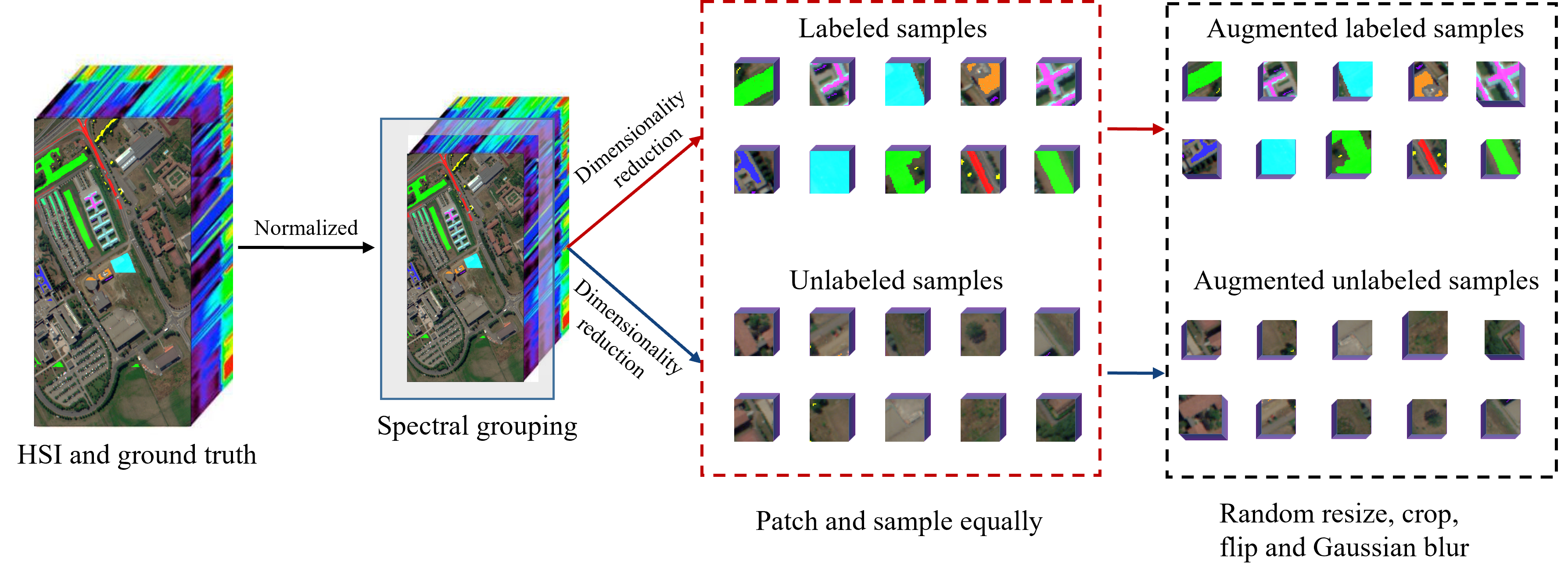}
\caption{Data processing overview.}
\label{fig_3}
\end{figure*}

\subsection{Semi-supervised Learning}

While labeled samples in HSIs are currently expensive to obtain, there are a large number of unlabeled samples that are easily accessible. Utilizing a very limited number of labeled samples and a large number of unlabeled samples simultaneously, the semi-supervised method shows good performance \cite{pan2018mugnet}. However, how to extract labeled and unlabeled sample features is a problem. There are various solutions, such as constructing pseudo-labels or neural network-based methods \cite{wu2017semi, dong2022weighted}.

The method based on pseudo-labels can directly use unlabeled samples. It usually clusters HSIs (both labeled and unlabeled) first. After assigning the cluster labels to all unlabeled data, we can train the model using supervised learning \cite{wu2017semi}. The main problem is that incorrect pseudo-labels cause error accumulation and long training time. Models based on CNNs or GNNs can directly encode unlabeled samples. CNNs \cite{zheng2020fpga, liu2021patch} extracts features of unlabeled samples through convolution kernels, which is an implicit semi-supervised learning method. By converting HSIs into graph data, features can be directly aggregated on all samples through GNNs \cite{dong2022weighted,9745164}. Liu et al. \cite{liu2020cnn} approached HSI classification from a graph perspective, utilizing graph convolutional networks to extract feature representations of irregular image regions in non-Euclidean space. However, these methods necessitate loading the entire HSI into memory, making it unsuitable for large-scale satellite image processing.

% Zheng et al propose a fast patch-free global learning framework, which utilize the encoder-decoder architecture to process the whole image. 

\section{Proposed Approach}
\label{proposed approach}

The framework incorporates four components, as illustrated in Fig. \ref{framework}. (a) data processing module: the data processing module prepares the training and test dataset for training and test; (b) encoder module: the encoder module extracts representation vectors from augmented samples; (c) loss module: the loss module generates supervised loss, unsupervised loss, or adaptively fuses supervised and unsupervised loss; (d) predictor module: we can use the predictor module to accomplish the classification task.

\subsection{Data Processing}
\label{data processing}

We construct a generalized data processing pipeline for real-world HSI classification, which is applicable to supervised learning, unsupervised learning and semi-supervised learning models for HSI classification, including normalization, data splitting, spectral grouping, dimensionality reduction and data augmentation. The structure of this pipeline is illustrated in Fig. \ref{fig_3}.

Following the common data processing procedure, we first normalize the data to be between 0 and 1. Then, as stated in Section \ref{introduction}, the random sampling strategy is an unrealistic use case. We selected a fixed proportion of HSIs from each class, in turn from top to bottom, as training samples. The remaining labeled samples are used as test samples. The visual division results can be found in \ref{Reference Datasets}. This method can ensure that the test set information will not be introduced during spatial feature extraction. The ground truth is regional rather than discrete, which allows us to more accurately measure how well the model generalizes to the new land covers.

CL requires learning invariant feature representations of objects from different views of the same object. Data augmentation is widely used for CL as a simple way to generate multi-view representations. HSIs have multiple spectral bands. Beyond conventional data augmentation, these bands can all be viewed as different observations of the same target. We halve the HSI into two groups for extracting invariant feature representation, as shown in Fig. \ref{fig_3}.  Considering the impact of different sensors and bands on the classification performance, we do not directly reduce each HSI to three dimensions to apply natural image augmentation methods. Instead, we use principal component analysis (PCA) \cite{9785802} to extract the most discriminative features in each HSI group. With each grouped spectral pixel as the center, image blocks $ X \in \mathbb{R} ^{H \times {W} \times C}$ are constructed. The common data augmentation includes random crop and resize, random vertical and horizontal flips, as well as Gaussian blur. It is worth noting that HSIs do not have the concept of color, so data augmentation, such as color jitter, is not adaptive. The input of HSI blocks $X$ is divided into two views after data processing: a labeled sample view $x$ and an unlabeled sample view $\hat{x}$.

\subsection{Encoder Modules}
\begin{figure}[!t]
\centering
\subfloat[]{\includegraphics[width=1.0in]{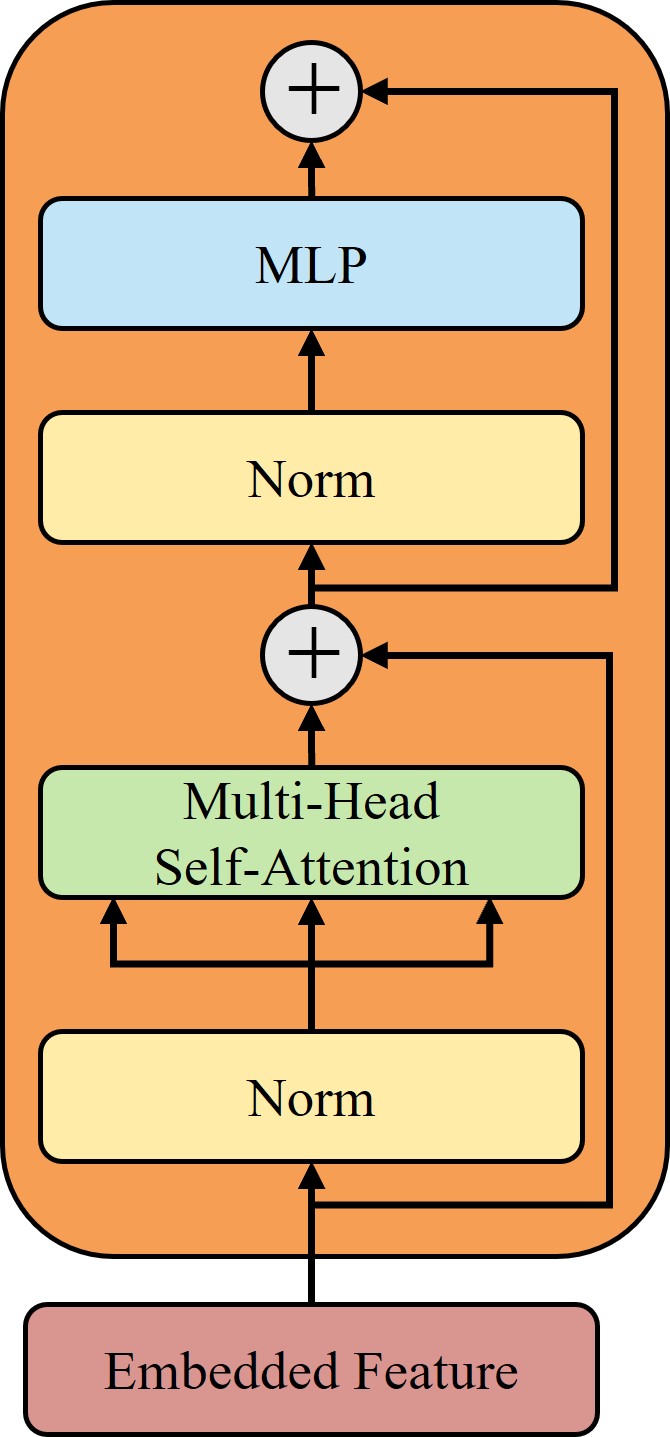}}
\hspace{2mm}
\subfloat[]{\includegraphics[width=0.8in]{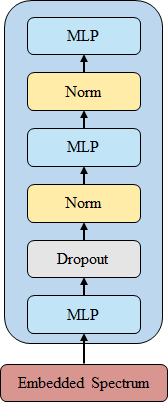}}
\hspace{2mm}
\subfloat[]{\includegraphics[width=0.8in]{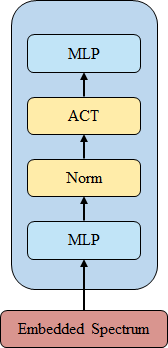}}
\caption{Detailed structures of network components: (a) the attention block, (b) the supervised head, (c) the contrastive head.}
\label{heads}
\end{figure}

\begin{table}
\tabcolsep=1.5mm  % 左右间距
\centering
\caption{Details of the Four Datases.}
\label{tab:table2}
\begin{tabular}{c|cccc}
\hline
& UP& Salinas& Dioni& DFC2018\\
\hline
Region or Country & Pavia& California& Dioni& Houston\\

 Sensor type & ROSIS& AVIRIS& Hyperion& ITRES CASI\\
 Bands  & 103& 204& 176& 48\\
 Wavelength($
 \mu m$)&  0.43-0.86&  0.40-2.5& 0.4-2.5 & 0.38-1.05\\
 GSD($m$) &  1.3& 3.7& 30 & 1.0\\
 Image size     &  610$ \times $340& 512$ \times $217& 250$ \times $1376 & 601$ \times $2384\\
 Labeled samples        & 42776& 54129& 20024&  504712\\
 Classes             & 9& 16& 12& 20\\
 Release time & 2003& 2001& 2018& 2018\\
 \hline
\end{tabular}
\end{table}

The encoder module is composed of a backbone $ f_b (cdot) $ and two projection heads $ f_s (cdot) $ and $ f_c (cdot) $. Our framework allows selecting combinations of network architectures to accomplish supervised classification, unsupervised classification, and semi-supervised classification with an arbitrary network backbone without any restrictions. 

Transformer-based methods have shown excellent performance on an increasing number of tasks. Many ViT-based methods have been proposed to solve HSI classification. Especially when the training samples are sufficient, ViT can achieve better performance. Our framework can be trained by global samples, which greatly expands the sample size. Therefore, we choose a simplified ViT as the backbone, named ViT-hsi, which follows the implementation of DINO \cite{caron2021emerging}. Here, we summarize some different network configurations. To encoder images more efficiently, the ViT architecture takes a sequence of flattened 2D patches as input. So we further reshape the image block into 2D patches, which passed through a trainable linear layer to form a set of patch embeddings. Position embeddings and class tokens have also been added to ensure consistency with previous works. Note that the class token only plays the role of aggregating global information when training the supervision head and is no different from other tokens when training the contrastive head. The embeddings and class tokens are passed into a four-layer standard attention block. An attention block is shown in Fig. \ref{heads}(a). The drop path layer regularization is added after each self-attention layer and multi-layer perceptron (MLP) layer to alleviate over-fitting. The self-attention mechanism updates the feature representation by encoding global image blocks. 

CNN-based models also show excellent performance. We also employed several ResNet variants to test the performance of our framework. The sliced HSI block size is usually much smaller than the natural image, so we set all convolutional kernels of ResNet to 3 and removed the pooling layer and the linear layer. Two projection heads $f_s(\cdot)$ and $f_c(\cdot)$ are shown in Fig \ref{heads}(b) and (c). The supervised head is responsible for mapping the extracted features to the category space. To improve the accuracy of the mapping, we use a 2-layer MLP. The Dropout layer and Normalization layer is used to improve the generalization ability of the model. The main function of the projection head is to map the features of two contrastive branches into the same subspace for computing the contrastive loss. The projection head consists of an MLP layer with hidden dimension 126 followed by ${l}_2$ normalization.

\begin{table*}
\tabcolsep=1.4mm  % 左右间距
% \scriptsize
\centering
\caption{Numbers of Training and Test Samples for UP, Salinas, Dioni, and DFC 2018}
\label{tab:table1}
\begin{tabular}{c|ccc|ccc|ccc|ccc}
\hline
\multicolumn{1}{c|}{}     
& \multicolumn{3}{c|}{UP}  
& \multicolumn{3}{c|}{Salinas}     
& \multicolumn{3}{c|}{Dioni}     
& \multicolumn{3}{c}{DFC2018}  \\ 
\hline
No.     & Class & Train & Test & Class & Train & Test & Class & Train  & Test  & Class & Train & Test\\
\hline
1                           & Asphalt                 & 1996                       & 4635                       & 
Broccoli-weeds-1& 419          & 1590       & 
\begin{tabular}[c]{@{}c@{}}Dense urban\\ fabric\end{tabular} & 632   & 630 & Healthy grass          & 1458 & 8341 \\
2                           & Meadows             & 5619                       & 13030                      & Broccoli-weeds-2                                                                         & 806                        & 2920                       & 
Mineral extraction site& 106           & 98                        & Stressed grass          & 4316     & 28186\\  
3                           & Gravel                   & 644             & 1455         & Fallow  & 425            & 1551              & 
\begin{tabular}[c]{@{}c@{}}Non-irrigated\\      arable land\end{tabular} & 307                        & 307                       & Artificial turf       & 331                        & 353 \\
4                           & Trees                      & 922                        & 2142                       & 
\begin{tabular}[c]{@{}c@{}}Fallow-rough-\\      plow\end{tabular}         & 287            & 1107   & Fruit trees             & 75                        & 75         & Evergreen trees   & 2005      & 11583 \\
5                           & Metal Sheets        & 417                      & 928             & Fallow-smooth           & 540           & 2138               & Olive groves            & 902           & 866                 & 
\begin{tabular}[c]{@{}c@{}}Deciduous trees\end{tabular}         & 676     & 4372 \\
6                           & Bare Soil              & 1545                     & 3484         & Stubble       & 794       & 3165         & 
\begin{tabular}[c]{@{}c@{}}Coniferous \\      forest\end{tabular}           & 181                        & 180                       & Bare earth                     & 1757                       & 2759\\
7                           & Bitumen             & 407        & 923        & Celery           & 746             & 2833      & 
Dense  vegetation& 2540                       & 2495                      & Water                & 147                        & 119\\
8                           & Bricks             & 1106        & 2576        & Grapes-untrained           & 2270             & 9001      & 
Sparce vegetation& 3244                       & 3130                      & \begin{tabular}[c]{@{}c@{}}Residential\\ buildings\end{tabular}                & 3809                        & 35953\\
9                           & Shadow                  & 285                    & 662               & 
\begin{tabular}[c]{@{}c@{}}Soil-senesced-\\ develop\end{tabular}           & 1311      & 4892      & 
\begin{tabular}[c]{@{}c@{}}Sparsely\\ vegetated areas\end{tabular} & 895           & 859         & 
\begin{tabular}[c]{@{}c@{}}Non-residential\\  buildings\end{tabular}        & 2789       & 220895\\
10                     & -    & -       & -            & Corn-weeds          & 684                  & 2594            & Rocks and sand         & 264          & 228                       & Roads       & 3188            & 42622                       \\
11                     & -    & -        & -            & Lettuce-4wk                   & 230           & 838            & Water             & 808             & 804           & Sidewalks       & 2699                       & 31303\\
12                    & -    & -        & -         & Lettuce-5wk        & 414                        & 1513            & Coastal water      & 202       & 196          & Crosswalks              & 225            & 1291\\
13                    & -    & -        & -       & Lettuce-6wk         & 185                & 731           & -       & -         & -       & 
\begin{tabular}[c]{@{}c@{}}Major\\ thoroughfares\end{tabular}            & 5193       & 41165 \\
14                    & -    & -     & -    & Lettuce-7wk      & 228      & 842    & -    & -    & -   & Highways   & 700   & 9149 \\
15                    & -             & -           & -      & Vinyard-untrained      & 1455        & 5813       & -     & -           & -         & Railways   & 1224        & 5713\\
16                    & -             & -           & -      & Vinyard-trellis     & 380        & 1427         & -     & -           & -         & Parking lots1   & 1179          & 10296\\
17                    & -    & -      & -      & -      & -       & -      & -     & -    & -    & 
\begin{tabular}[c]{@{}c@{}} Parking lots2\end{tabular}      & 127    & 22 \\
18                   & -    & -       & -      & -      & -      & -      & -     & -     & -       & Cars   & 848      & 5730 \\
19                   & -    & -       & -      & -      & -      & -      & -     & -     & -      & Trains  & 493    & 4872\\
20                   & -    & -       & -      & -      & -      & -      & -     & -     & -      & Stadium seats    & 1313   & 5511\\ 
\hline
\multicolumn{1}{c|}{Ratio(\%)}    & -   & 0.3    & 0.7   & -    & 0.2  & 0.8   & -   & 0.5  & 0.5   &  - & 0.07  & 0.93\\
\multicolumn{1}{c|}{Total} & \multicolumn{1}{c}{-}   & \multicolumn{1}{c}{12941} & \multicolumn{1}{c|}{29835} & \multicolumn{1}{c}{-}     & \multicolumn{1}{c}{11174} & \multicolumn{1}{c|}{42955} & \multicolumn{1}{c}{-}     & \multicolumn{1}{c}{10156} & \multicolumn{1}{c|}{9868} & \multicolumn{1}{c}{-}       & \multicolumn{1}{c}{34477} & \multicolumn{1}{c}{470235} \\ 
\hline
\end{tabular}
\end{table*}

\begin{figure*}[!t]
\centering
\includegraphics[width=7in]{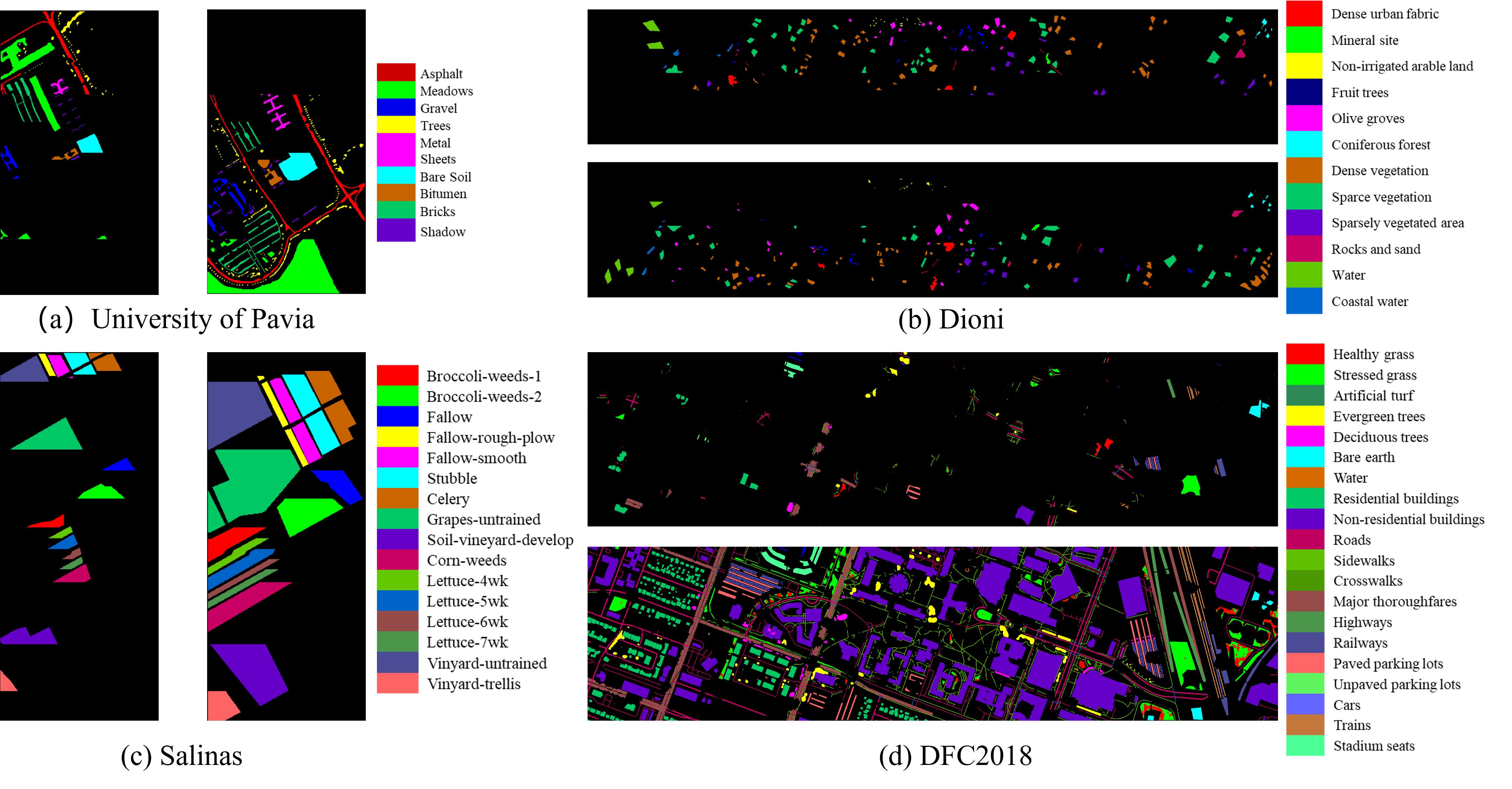}
\caption{The training and test ground-truth maps on the four reference datasets, i.e., (a) UP, (b) Dioni, (c) Salinas, (d) DFC2018.}
\label{fig_5}
\end{figure*}

\subsection{Loss Functions and Fusion Strategy}
The existing HSI classification approaches mainly consider cross entropy to assess how close the actual output is to the desired output to calculate the loss value, which is expressed as:
\begin{equation}
\label{deqn_ex1a}
{\cal L}_{ce}(p, q) =  - \sum\limits_x {p(x)\log q(x)},
\end{equation}
where $p(x)$ is the expected distribution of $x$ and $q(x)$ is the distribution of the corresponding ground truth. This approach requires ground truth to be involved in each iterative training. In order to take advantage of the rich unlabelled samples, we use CL to extract features of unlabelled samples. We randomly sample a batch size of unlabelled samples to define the CL task. Let $N$ be the size of a batch. The number of unlabelled samples constructed by data augmentation is $2N$. Consider the outputs $z$ and ${\hat{z}}$ of the two branches as a pair of positive samples, then $z$ and the remaining $2N-1$ samples constitute negative samples. The loss function is defined as

\begin{equation}\label{equation 2}
{\cal L}({z_i},{\hat{z}_j}) = \log \frac{{{e^{\theta ({z_i},{z_j})/\tau }}}}{{\sum\limits_{m = 1}^{2N} \mathbb{I}{{_{[m \ne i]}}{e^{\theta ({z_i},{z_m})/\tau }}} }},
\end{equation}
\begin{equation}\label{equation 3}
\theta ({z},{\hat{z}}) = z^T\cdot{\hat{z}}/||z^T||\cdot||{\hat{z}}||,
\end{equation}
where ${z_i}$ and ${z_j}$ represent the nonlinear embeddings generated from training samples. $\theta ({z},{\hat{z}}) $  calculate the cosine similarity between  $z$ and ${\hat{z}}$.  $\tau$ is the temperature parameter and $ \mathbb{I}_{[m \ne i]} $ acts as an indicator function. The final optimization goal is to minimize the average over all positive pairs.
\begin{equation}\label{equation 4}
{\cal L}_{cl} = -\frac{1}{{2N}}\sum\limits_{i = 1}^N {[{\cal L}({z_i}, {\hat{z}_j}) + {\cal L}({\hat{z}_j}, {z_i})]}.
\end{equation}
Considering contrastive learning and supervised learning as atomic tasks in a classification task, we can define a multi-task ${\cal T}$, whose loss function ${\cal L}_{T}$ consisting of a cross-entropy loss and an infoNCE loss. This yields a combined loss function
\begin{equation}\label{equation 5}
{\cal {L}_{\cal T}}= \sum\limits_{{t} \in {\cal T}} {{\cal {L}}{_t} \cdot w_t}.
\end{equation}
Due to each of the contributing single loss functions behaving differently, $w_t$ is used to regulate the proportion of atomic tasks contributing to the combined loss. Instead of setting a fixed parameter, inspired by \cite{liebel2018auxiliary}, we add a learnable parameter to weighting each. This final combined loss function is defined as
\begin{equation}\label{equation 6}
{\cal {L}_{\cal T}} = \sum\limits_{t \in T} {\frac{1}{{2 \cdot w_t^2}}\cdot {L_t}}  + \ln (1 + w_t^2),
\end{equation}
where $\ln (1 + w_t^2)$ is a regularization term to avoid trivial solutions.

\subsection{Training and Test Procedure}

 As shown in Fig. \ref{framework}, the training set from the data processing module, encoder module, and loss module is used for training. The supervised loss is computed along the red line. The labeled image block $x$ from the augmented views is passed to the backbone to extract latent features $h$. Then we use a supervised head to obtain predicted labels $p(x)={f_s}(h)$. The supervised loss can be calculated by cross-entropy loss between predicted labels and ground truth. The difference between unsupervised and supervised classification is that we use another identical branch to extract features from unlabeled image block $\hat{x}$. Features from both branches are mapped to a high-dimensional subspace using a contrastive head. We maximize their similarity to train the network, as shown by the dark blue line. We design a new semi-supervised learning model relying on fusing supervised and unsupervised learning. During each iteration, we obtain the supervised loss and the contrastive loss separately and then fuse the two in a learnable way to obtain the semi-supervised loss.

After the convergence of KnowCL, the test set is used to evaluate the model performance.  A combination of the trained backbone and a classifier is used to validate the effectiveness of the model. Supervised classification directly outputs classification results that can be used for testing. Unsupervised classification usually verifies the model's effectiveness with the help of fine-tuning. In semi-supervised learning, due to the inclusion of the supervised head, we can directly output the classification results from the supervised head for evaluation in an end-to-end manner. We can also use the linear evaluation protocol \cite{oord2018representation, bachman2019learning}, which consists in training a linear classifier on top of the frozen representation. Finally, following \cite{ye2019unsupervised}, a simple KNN classifier is used to evaluate the model's performance. In the experimental session, we discuss each classification approach in detail.

\section{Experiments}

\subsection{Reference Datasets}
\label{Reference Datasets}

Four HSI datasets are mainly used in this study which are the University of Pavia dataset (UP), the Salinas dataset, the Dioni dataset and the 2018 IEEE GRSS data fusion contest dataset (DFC2018). These four datasets used four sensors and captured four regions at different times, as shown in Fig. \ref{fig_5}. The DFC2018 has a minimum of 48 bands, while the Salinas has up to 204 bands. The DFC2018 and UP only scan the visible near-infrared spectra. The Salinas and Dioni acquire both visible near-infrared and shortwave infrared spectra. Dioni is located in the middle of the wilderness and has a low spatial resolution, while the remaining three datasets focus on urban and agricultural scenes and have a high spatial resolution. Of these four datasets, the Salinas was collected the earliest, which is relatively small in size. DFC2018 is approximately 12 times larger than the Salinas data and also contains the largest number of labeled samples. The detailed parameters associated with these data are given in Table \ref{tab:table2}.

As stated in the section \ref{data processing}, we adopted a well-disjoint data sampling strategy for UP, Salinas, and Dioni datasets.  For the DFC2018, the training and test set had been defined by the GRSS. The detailed division is shown in Table \ref{tab:table1} and Fig. \ref{fig_5} visualizes the distribution of labels in the training set and test set.

\subsection{Implementation Details}

\subsubsection{Baseline Methods}
% However, it is worth noting that there is no unified model for the HSI classification community. The hyperparameters of various models are sometimes not comprehensively described in the papers. We try to reproduce the state-of-the-art models. The reimplementations are close to the use in the original articles as possible. However, for some unknown hyperparameters, we set them only empirically.

\begin{table}
\tabcolsep=0.9mm  % 左右间距
\centering
\caption{Quantitative Comparison with Various Crop Sizes on the DFC2018 Datasets.}
\label{tab:table3}
\begin{tabular}{c|ccccccccc}
\hline
Crop size & 17    & 18    & 19   & 20    & 21    & 22    & 23    & 24    & 25    \\
\hline
OA        & 73.63 & 72.26 & 73.7& 72.15 & 73.89 & 73.81 & \textbf{73.98} & 73.71 & 73.21   \\
\hline
\end{tabular}
\end{table}

To comprehensively evaluate the performance of the the proposed HSI classification framework in the disjoint data, a traditional machine learning method EMP-SVM and four state-of-the-art models each from supervised learning, semi-supervised learning, and unsupervised learning are selected for comparison, respectively. In the supervised learning scheme, DFFN\cite{li2019dff} incorporates spectral-spatial features through global and local feature extraction, and SPRN\cite{zhang2021spectral} explores the enhancement of spectral partition on the model performance. FDGC\cite{9785802} and SpectralFormer\cite{hong2021spectralformer} explored the application of GCN and transformer modules for HSI classification, respectively. The unsupervised learning AAE and VAE \cite{cao2020unsupervised} extract latent feature representations with the help of HSI reconstruction for downstream classification tasks. Both DMVL\cite{liu2020deep} and XDCL\cite{zhang2022cross} use the infoNCE loss function to construct the CL model.  For semi-supervised learning, CNN-based FPGA\cite{zheng2020fpga} solves the HSI classification problem in terms of semantic segmentation. CEGCN\cite{liu2020cnn}, WFCG\cite{dong2022weighted}, and EMSGCN\cite{9745164} are models belonging to GNN, which convert the HSI into a graph by constructing hyperpixels to extract the global features in the non-Euclidean space.

\subsubsection{Technical details}
This framework unifies supervised learning, unsupervised learning and semi-supervised learning, and we name them KnowCL-SU, KnowCL-US and knowCL, respectively. KnowCL(under the kNN classifier) has the best performance and is mainly used to analyse and discuss the superiority of the proposed algorithm. We adopt a modified ViT as the backbone which has four attention blocks. We first intercept HSI blocks of size $C\times 25 \times 25$ based on each center pixel, where $C$ depends on the input HSI. Following the standard ViT, the image blocks are further flattened with patches size of 4. The input HSI blocks will be encoded into 126-D latent features. These features will be further mapped to 256-D latent features by a 1-layer MLP contrastive head and a 2-layer MLP supervisory head to compute the infoNCE loss and cross-entropy loss, respectively. The temperature $\tau$ is set as 0.5. The data augmentation is available in PyTorch’s torchvision package. The random crop size is set to 23×23-pixel, considering the experimental results. AdamW is used as the optimizer with a cosine annealing learning rate schedule. We train for 30 epochs with batch size 256, weight decay 0.000001, and learning rate 0.001. We follow \cite{ye2019unsupervised, wu2018unsupervised} to evaluate the performance of the model with a weighted kNN classifier, and the $k$ value is set to 5 according to the performance of the experiment. We also further tested our model on the supervised head and linear classification protocol to explore the accuracy limits of the model. All models are implemented based on Pytorch 1.9 and performed with an NVIDIA GeForce RTX 3090 GPU.

% 我们依照主要用到了一个加权knn 分类器评估模型的表现，and tok-k(k=5) 依照实验的表现。我们也进一步测试了我们的模型在线性分类准则和fine-tuning 上的精度表现。

\subsubsection{Metrics}
There is no unified model for the HSI classification community, leading to a wide range of performance, number of parameters, and training time for these models. We evaluate these algorithms using common practice in deep learning. They are the overall accuracy, the average accuracy, the Kappa coefficient, parameters, FLOPs, and execution time.

\begin{table}
\centering
\caption{Quantitative Comparison with Various Batch Size on the DFC2018 Datasets.}
\label{tab:table4}
\begin{tabular}{c|ccccc}
\hline
Batch size & 64    & 128   & 256   & 512   & 1024  \\
\hline
OA         & 73.78 & 73.15 & \textbf{73.98} & 72.91 & 71.57 \\
\hline
\end{tabular}
\end{table}

\begin{table}
\tabcolsep=0.9mm  % 左右间距
\centering
\caption{Quantitative Comparison with Different Bands on the Four Datasets.}
\label{tab:table5}
\begin{tabular}{c|ccccccccc}
\hline
Bands   & 1     & 3     & 5      & 7     & 10    & 15    & 20    & 25    & 30    \\

UP      & -     & 69.33 & 78.23& 78.5  & \textbf{91.36} & 90.11 & 88.55 & 85.2  & 86.52 \\
salina  & -     & 87.74 & 91.96  & \textbf{93.34} & 92.7  & 91.12 & 90.91 & 91.45 & 90.93 \\
Dioni   & 57.44 & \textbf{82.52} & 80.76  & 75.79 & 68.46 & 68.33 & 65.01 & 65.13 & 63.43 \\
DFC2018 & -     & 64.68 & 71.675 & 72.13 & \textbf{73.98} & 70.85 & 71.47 & -     & -    \\
\hline
\end{tabular}
\end{table}

\subsection{Parameter analysis}

\begin{table*}
\centering
\caption{Quantitative Comparison with Other Classification Methods on the UP Dataset.}
\label{tab:table6}
\begin{tabular}{c|c|ccccccc}
\hline
&         Learning        & Parameters & FLOPs   & OA    & AA    & Kappa & Training time & Test time \\ \hline
EMP-SVM        &        -                  & -          & -       & 61.58 & 74.39 & 53.17 & -             & -         \\ 
\hline
DFFN \cite{li2019dff}&  \multirow{5}{*}{Supervised Learning}& 532.7K     & 175.47M & 78.69 & 79.28 & 72.64 & 660.81s        & 63.03s     \\
SPRN \cite{zhang2021spectral}&                          & 174.67K    & 63.43M  & 78.62 & 70.54 & 72.72 & 270.37s& 31.13s     \\ 
FDGC \cite{9785802}&                          & 2.45M      & 14.84M  & 75.63 & 75.28 & 68.88 & 113.98s        & 2.24s      \\
SpectralFormer \cite{hong2021spectralformer}&                         & 99.25K     & 10.29M  & 87.10 & 87.50 & 83.24 & 453.00s        & 3.23s\\
 KnowCL-SU& & 931.43K& 20.09M
& 86.27 & 77.87 & 81.53 & 545.99s&3.70 
\\         \hline
DMVL \cite{liu2020deep}& \multirow{5}{*}{Unsupervised Learning}& 11.3M& 917.5M& 58.75& 47.34 & 45.45 & 197.67min        & 7.37s      \\
VAE \cite{cao2020unsupervised}&                          & 824.03K    & 25.2M   & 68.40 & 63.43 & 59.66 & 110.53min        & 43.74s    \\
AAE \cite{cao2020unsupervised}&                          & 758.37K    & 25.1M   & 73.64 & 72.45 & 66.59 & 125.1min         & 28.26s     \\
XDCL \cite{zhang2022cross}&                          & 6.47M      & 77.11M  & 81.04 & 82.66 & 75.38 & 506.67s        & 2.17s      \\
 KnowCL-US& & 599.77K& 19.76M
& 82.04 & 73.23 & 76.31 & 516.74 &3.16 
\\ \hline
FPGA \cite{zheng2020fpga}& \multirow{6}{*}{Semi-supervised Learning}& 2.56M& 110.42G & 63.18& 70.73 & 55.56 & 83.61s         & 0.07s      \\
CEGCN \cite{liu2020cnn}&                           & 152.85K    & 13.06G  & 83.31 & 87.34 & 78.83 & 47.32s         & 10.10s     \\
WFCG \cite{dong2022weighted}&                          & 63.19K     & 13.23G  & 79.80 & 87.38 & 74.90 & 60.82s         & 7.03s      \\
EMSGCN \cite{9745164}&                          & 201.69k    & 14.68G  & 86.45 & 86.25 & 82.23 & 64.54s         & 1.88s      \\

KnowCL(kNN) & & 997.09K& 20.15M
& \textbf{91.36 }& \textbf{87.75} & \textbf{88.45} & 655.76s        & 3.64s  \\
KnowCL(Linear evaluation)& & 997.09K& 20.15M
& 91.05 & 82.99 & 88.1  & 33.1s          & 2.7s\\ \hline
\end{tabular}
\end{table*}

\begin{figure}[!t]
\centering
\subfloat[]{\includegraphics[width=3.3in]{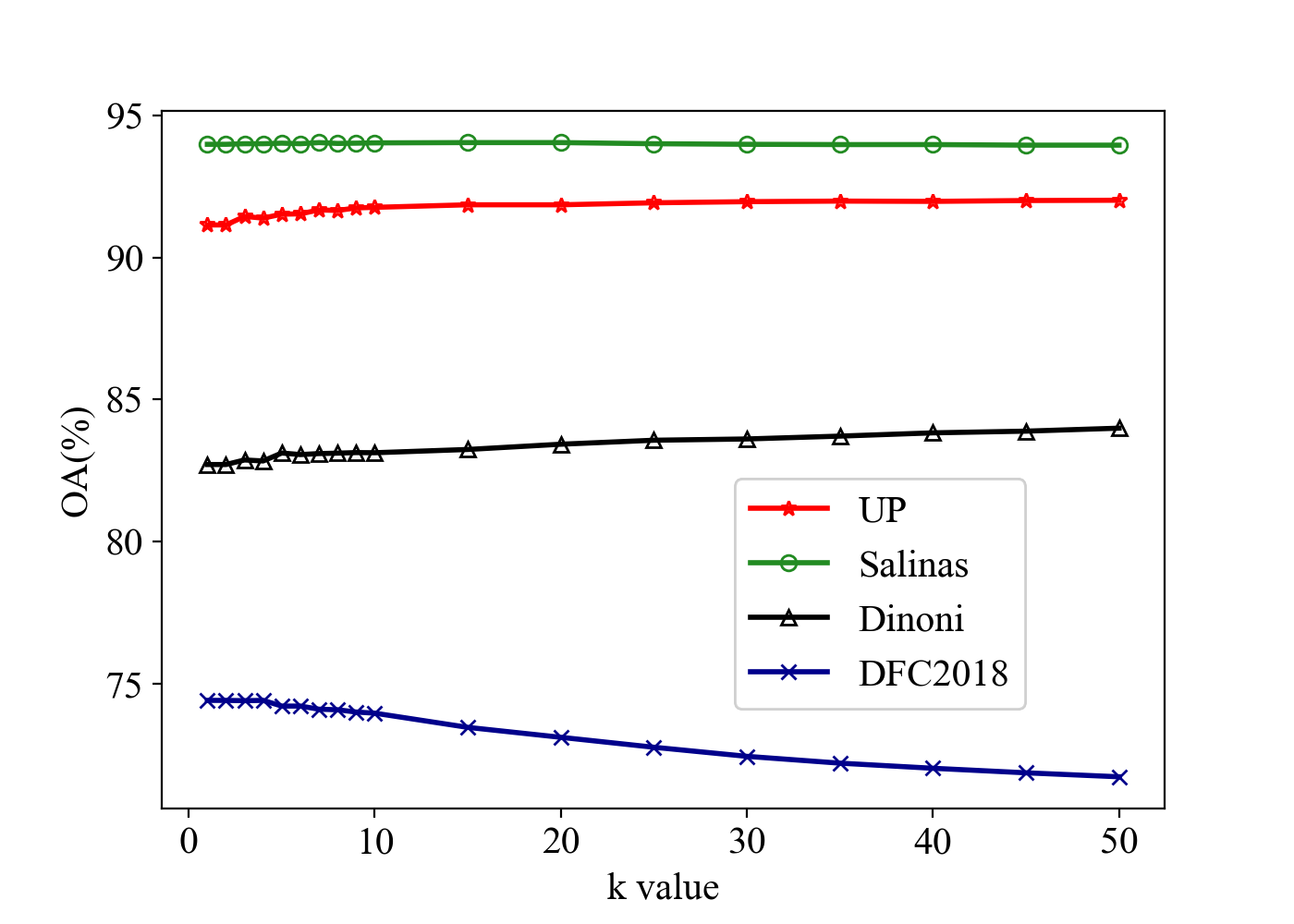}}
\caption{Quantitative comparison with various $k$ values on the four datasets.}
\label{KNN}
\end{figure}

\begin{figure}[!t]
\centering
\subfloat[UP]{\includegraphics[width=1.5in]{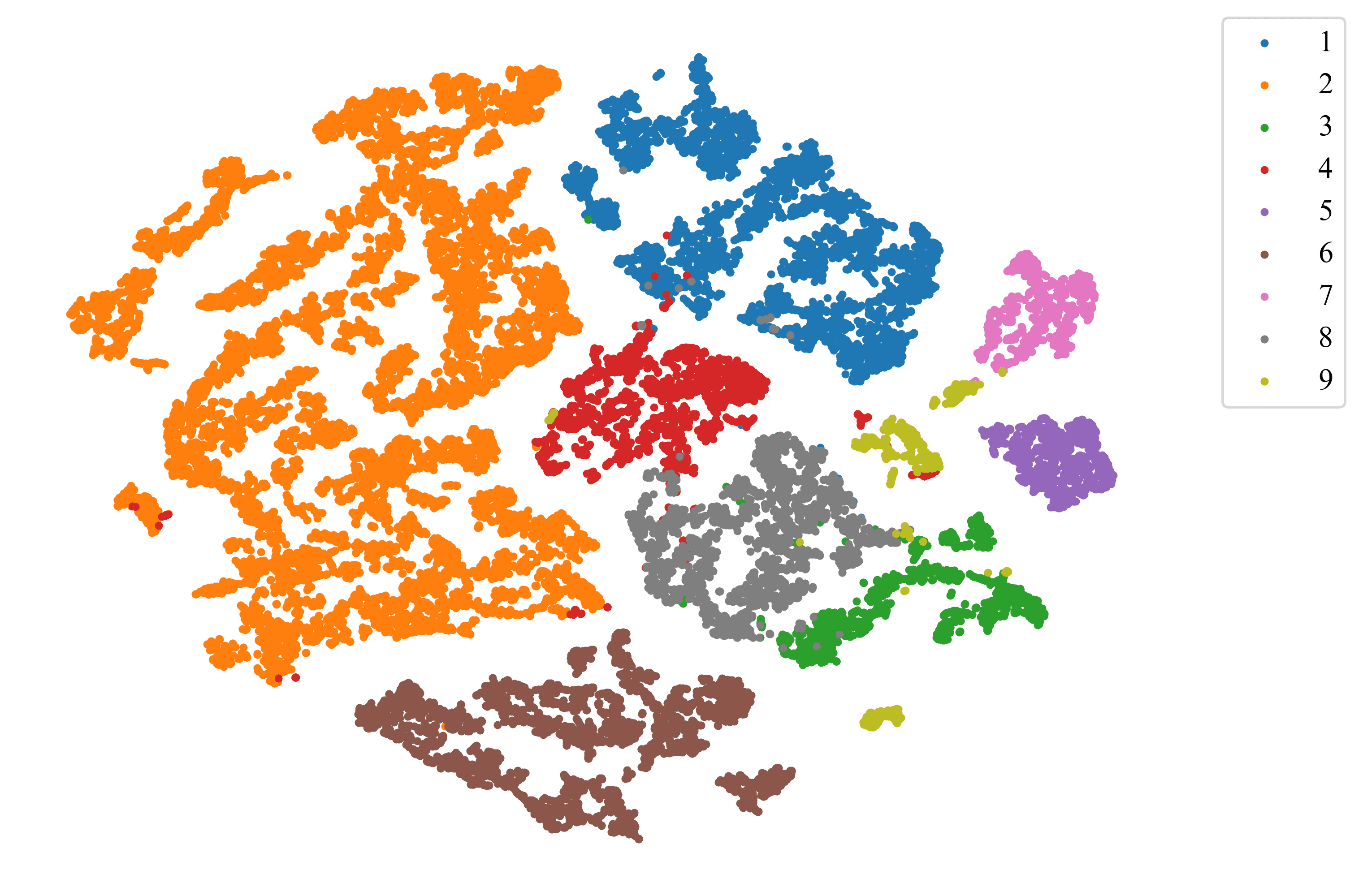}}
\hspace{3mm}
\subfloat[Salinas]{\includegraphics[width=1.5in]{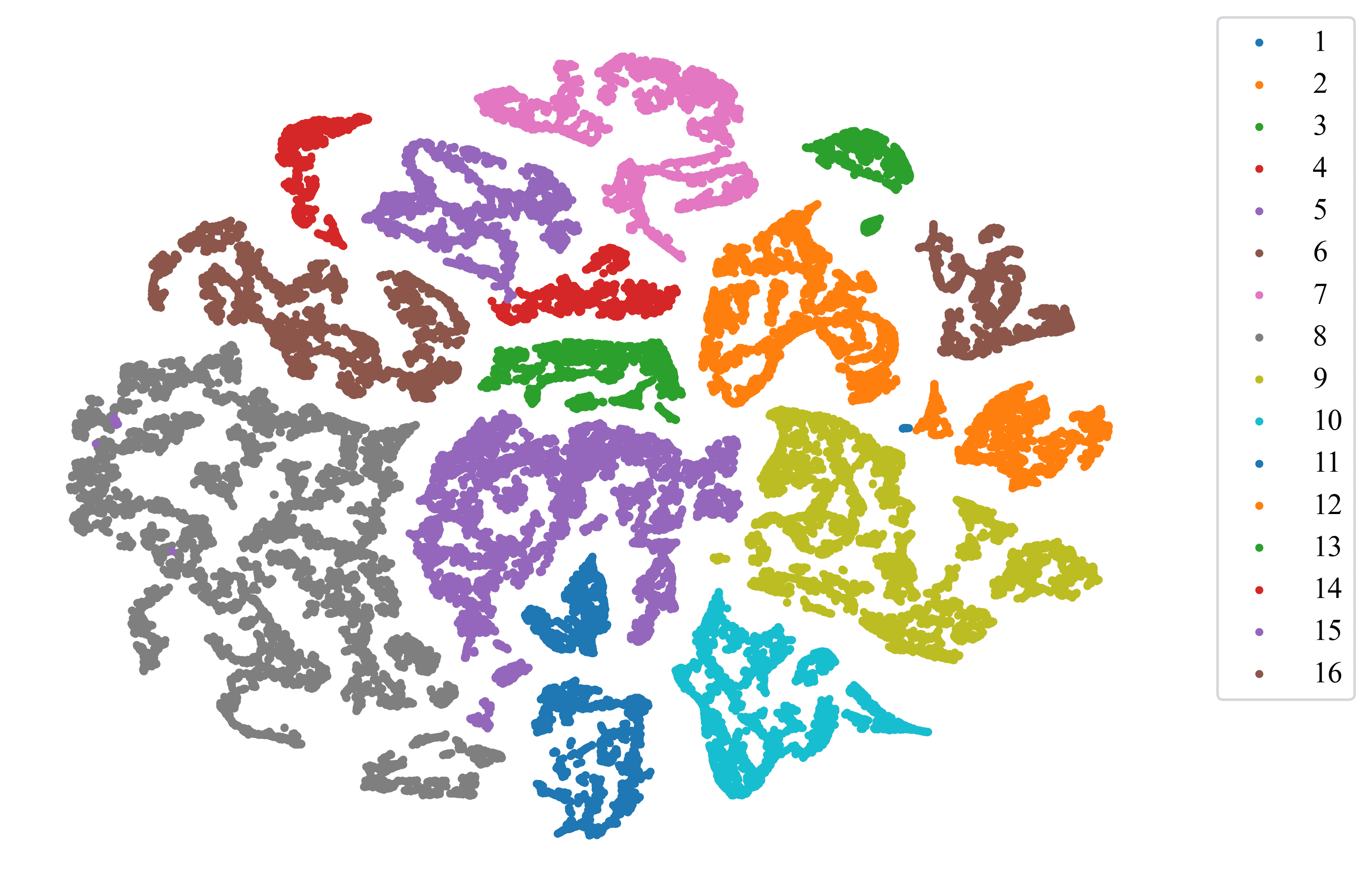}}
\hfil
\subfloat[Dioni]{\includegraphics[width=1.5in]{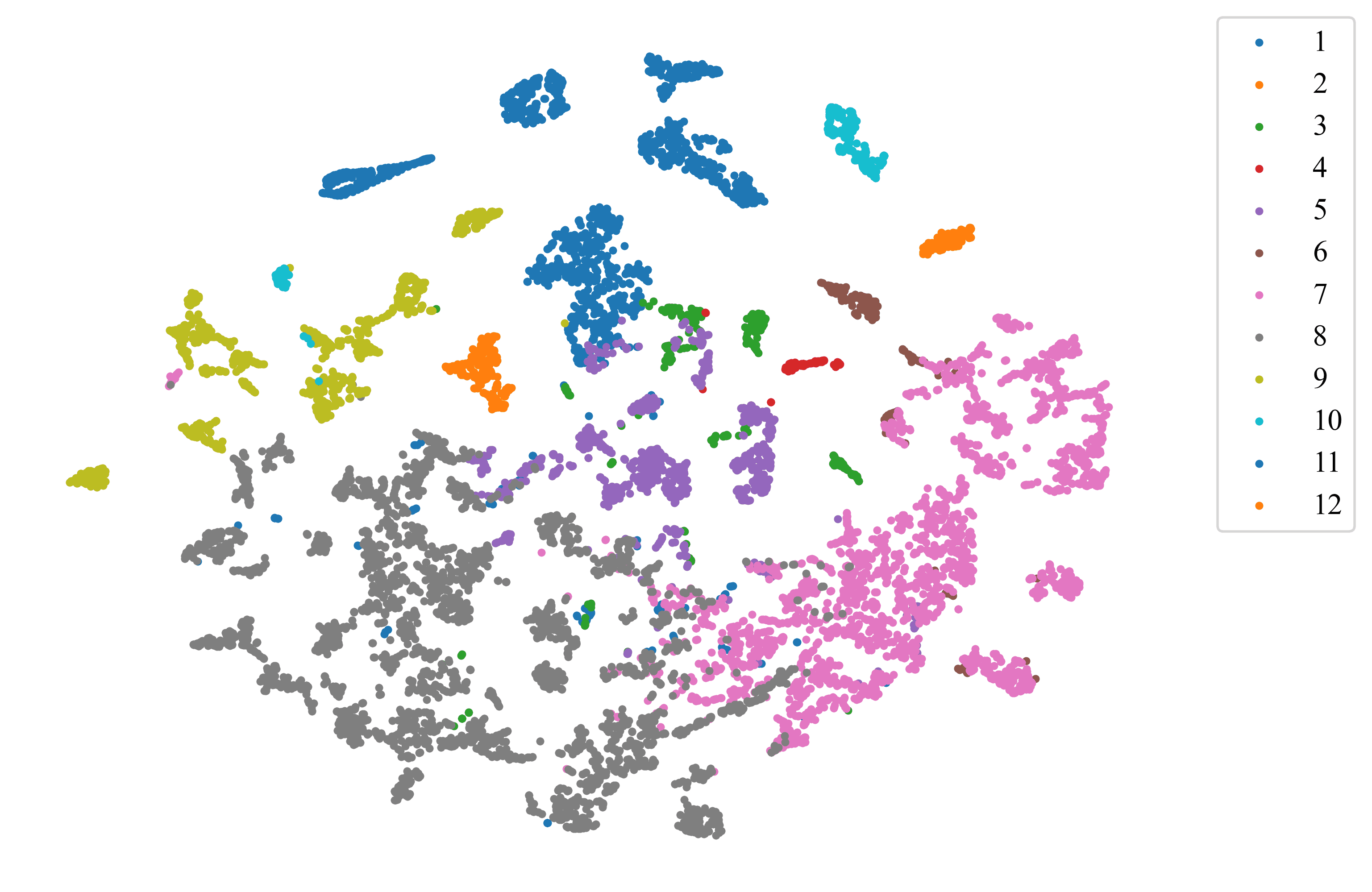}}
\hspace{3mm}
\subfloat[DFC2018]{\includegraphics[width=1.5in]{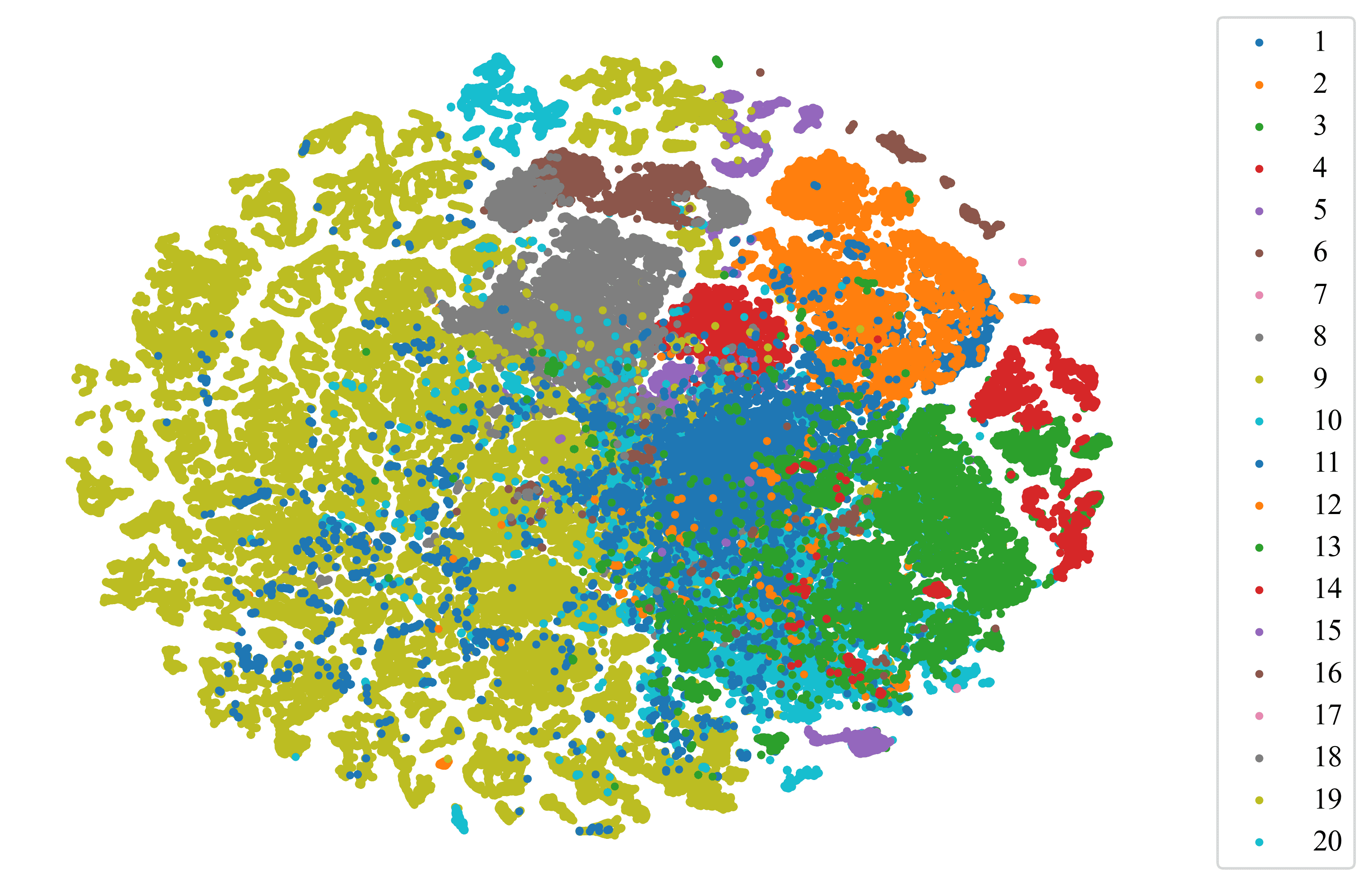}}
\caption{T-SNE isualization of the four datasets using KnowCL.}
\label{tsne}
\end{figure}

\begin{table*}
\centering
\caption{Quantitative Comparison with Other Classification Methods on the Salinas Dataset.}
\label{tab:table17}
\begin{tabular}{c|c|ccccccc}
\hline
&    Learning     & Parameters& FLOPs             & OA & AA & Kappa& Training time & Test time \\
\hline
EMP-SVM               &  -              & -          & -                           & 78.85       & 84.29       & 76.28          & -             & -         \\
\hline
DFFN \cite{li2019dff}
& \multirow{5}{*}{Supervised Learning}& 435.71K    & \multicolumn{1}{l}{140.31M} & 82.06       & 88.95       & 80.03          & 473.18s        & 33.87s      \\
SPRN \cite{zhang2021spectral}
&         & 176.97K    & \multicolumn{1}{l}{63.43M}  & 82.06       & 88.95       & 80.03          & 234.56s         & 51.56s     \\
FDGC \cite{9785802}
&       & 1.98M      & 14.18M      & 85.84       & 91.85       & 84.21          & 108.78s        & 3.92s      \\
SpectralFormer \cite{hong2021spectralformer}
&         & 99.70K     & 20.29M       & 87.62       & 91.90       & 86.20          & 551.55s        & 6.58s      \\
 KnowCL-SU& & 927.19K& 19.87M
& 85.60 & 90.12 & 83.93 & 451.04 &3.69 
\\
\hline
DMVL \cite{liu2020deep}& \multirow{5}{*}{Unsupervised Learning}& 11.31M& 917.5M& 85.06& 87.69   & 83.33     & 103.47min  & 5.19s     \\
VAE \cite{cao2020unsupervised}&      & 824.03K    & 25.2M  & 93.07       & 93.76       & 92.25     & 63.27min      & 38.71s     \\
AAE \cite{cao2020unsupervised} &         & 758.37K    & 25.1M   & 91.15    & 92.36   & 90.12   & 65.30min     & 45.21s\\
XDCL \cite{zhang2022cross}&              & 6.67M     & 99.67M   & 78.84    & 82.73   & 76.54   & 586.67s      & 4.89s\\ 
KnowCL-US& & 593.73K& 19.54M
& 93.45& 96.34& 92.68& 447.70 &3.68
\\  \hline
FPGA \cite{zheng2020fpga}& \multirow{6}{*}{Semi-supervised Learning}& 2.65M& 67.77G         & 78.61& 79.76       & 77.35          & 81.64s         & 0.08s      \\
CEGCN \cite{liu2020cnn}&       & 166.89K    & 8.58G     & 85.08    & 88.66     & 83.26      & 32.91s      & 12.19s     \\
WFCG \cite{dong2022weighted}&  & 76.78K     & 8.62G& 85.68    & 87.84     & 83.97      & 26.32s      & 4.15 s     \\
EMSGCN \cite{9745164}&         & 215.72K    & 9.45G     & 86.92    & 90.20     & 85.36      & 39.37s      & 1.57s  \\
KnowCL(kNN)  & & 992.85K& 19.94M
& \textbf{93.62}    & \textbf{96.78}   & \textbf{93.11}   & 585.05s      & 4.47s\\
KnowCL(Linear evaluation)& & 992.85K& 19.94M
& 93.53   & 95.3     & 92.77   & 38.3s   & 3.8s\\
\hline
\end{tabular}
\end{table*}

\begin{figure*}[!t]
\centering
\subfloat[Ground Truth]{\includegraphics[width=1.0in]{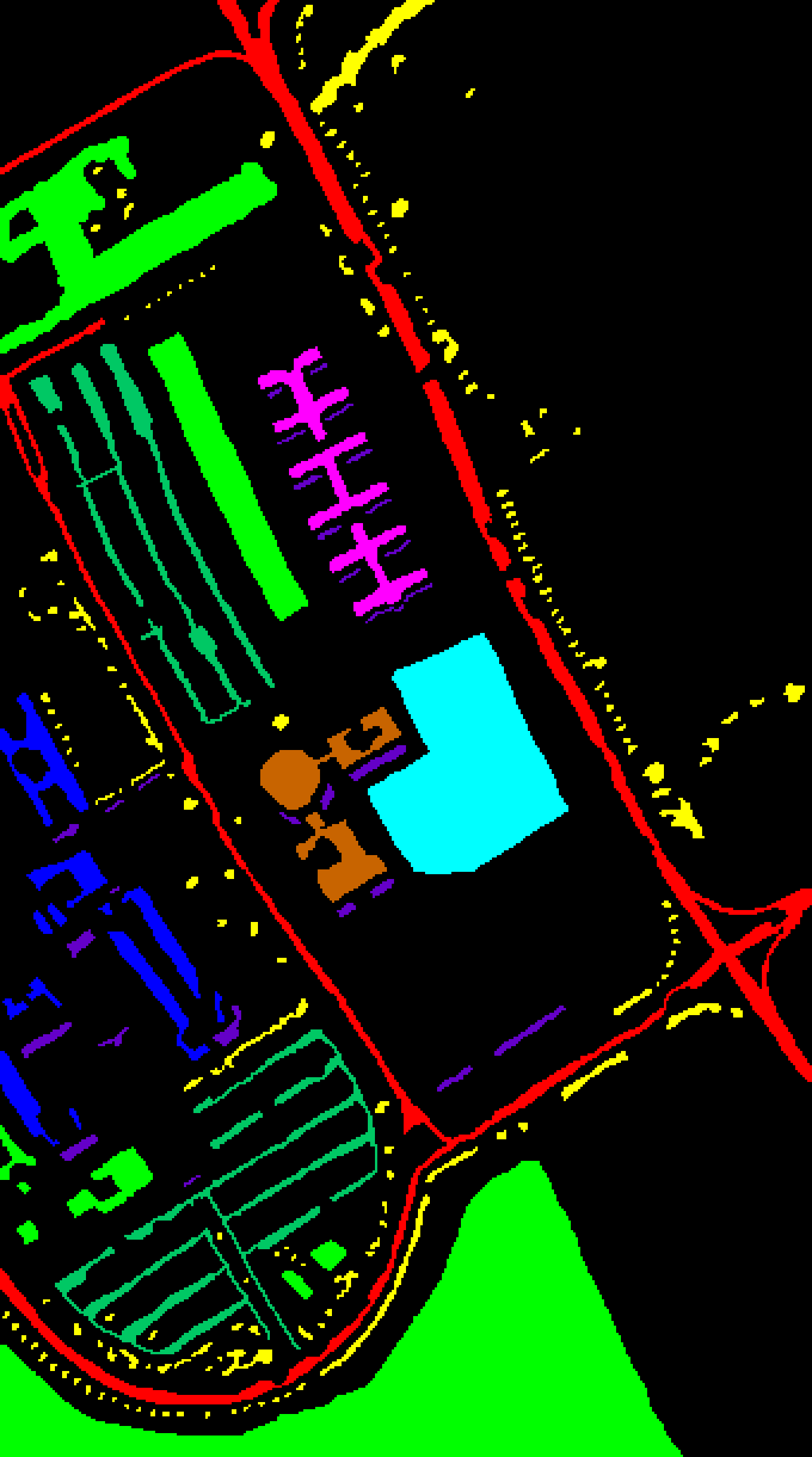}}
\hspace{3mm}
\subfloat[SVM]{\includegraphics[width=1.0in]{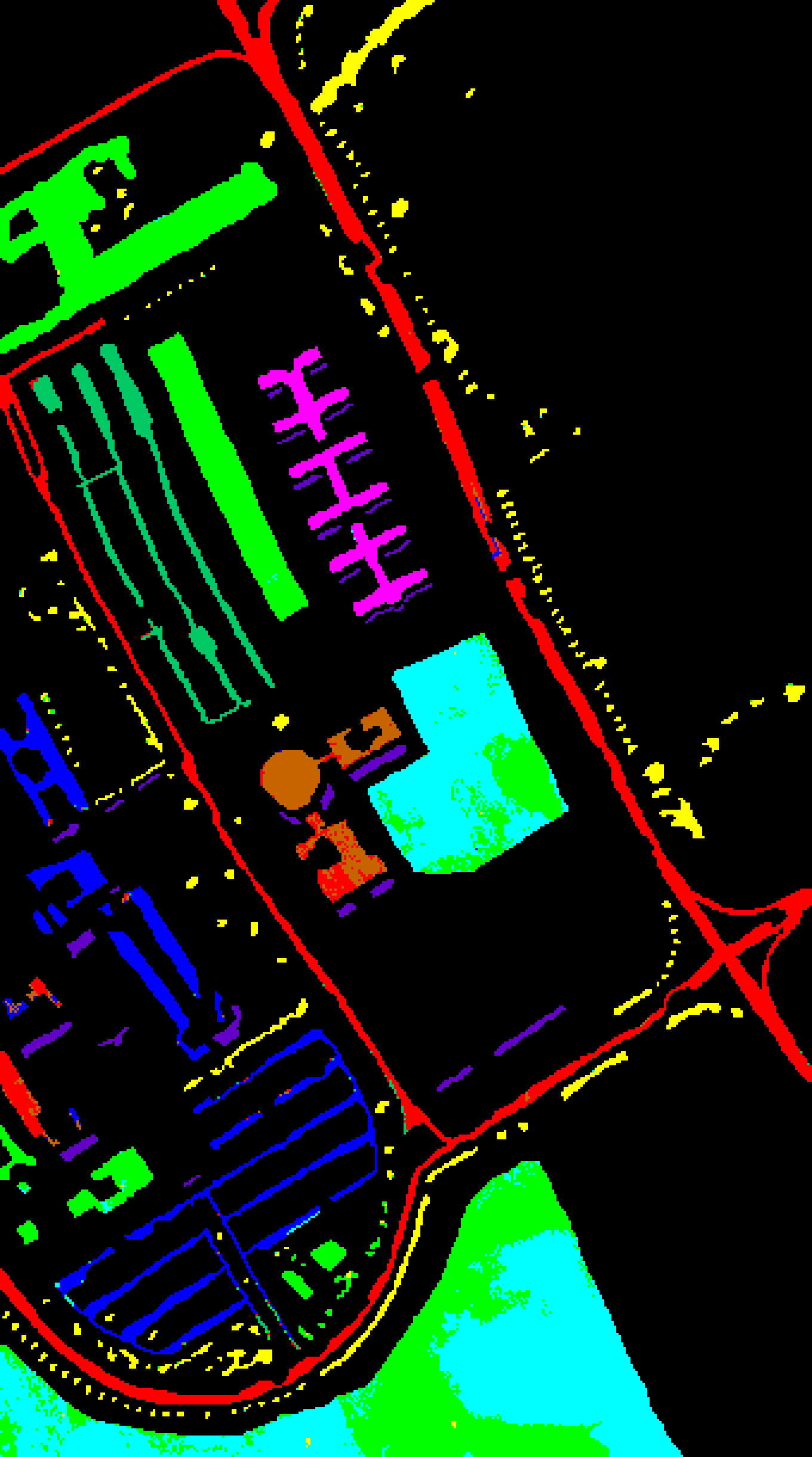}}
\hspace{3mm}
\subfloat[SpectralFormer]{\includegraphics[width=1.0in]{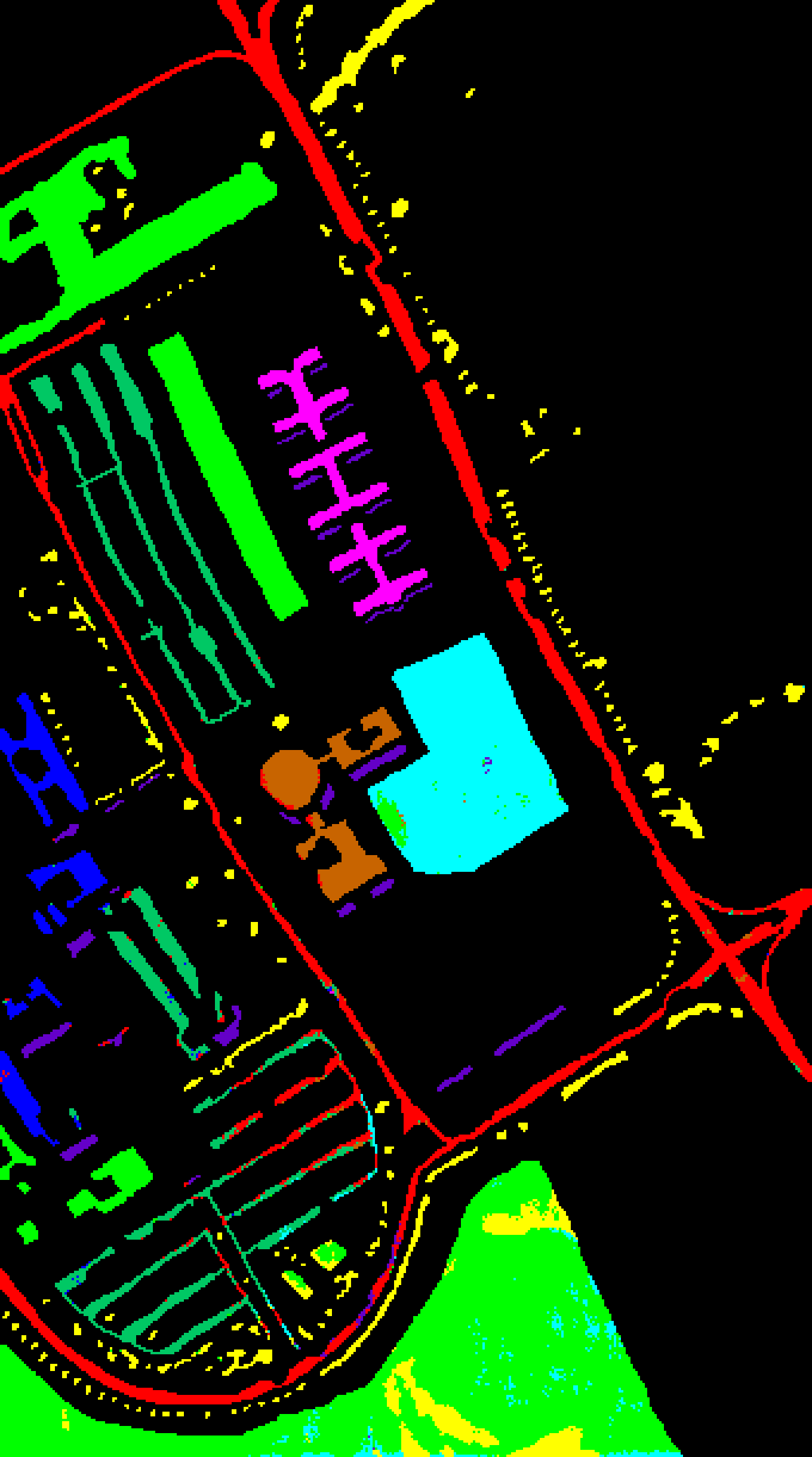}}
\hspace{3mm}
\subfloat[EMSGCN]{\includegraphics[width=1.0in]{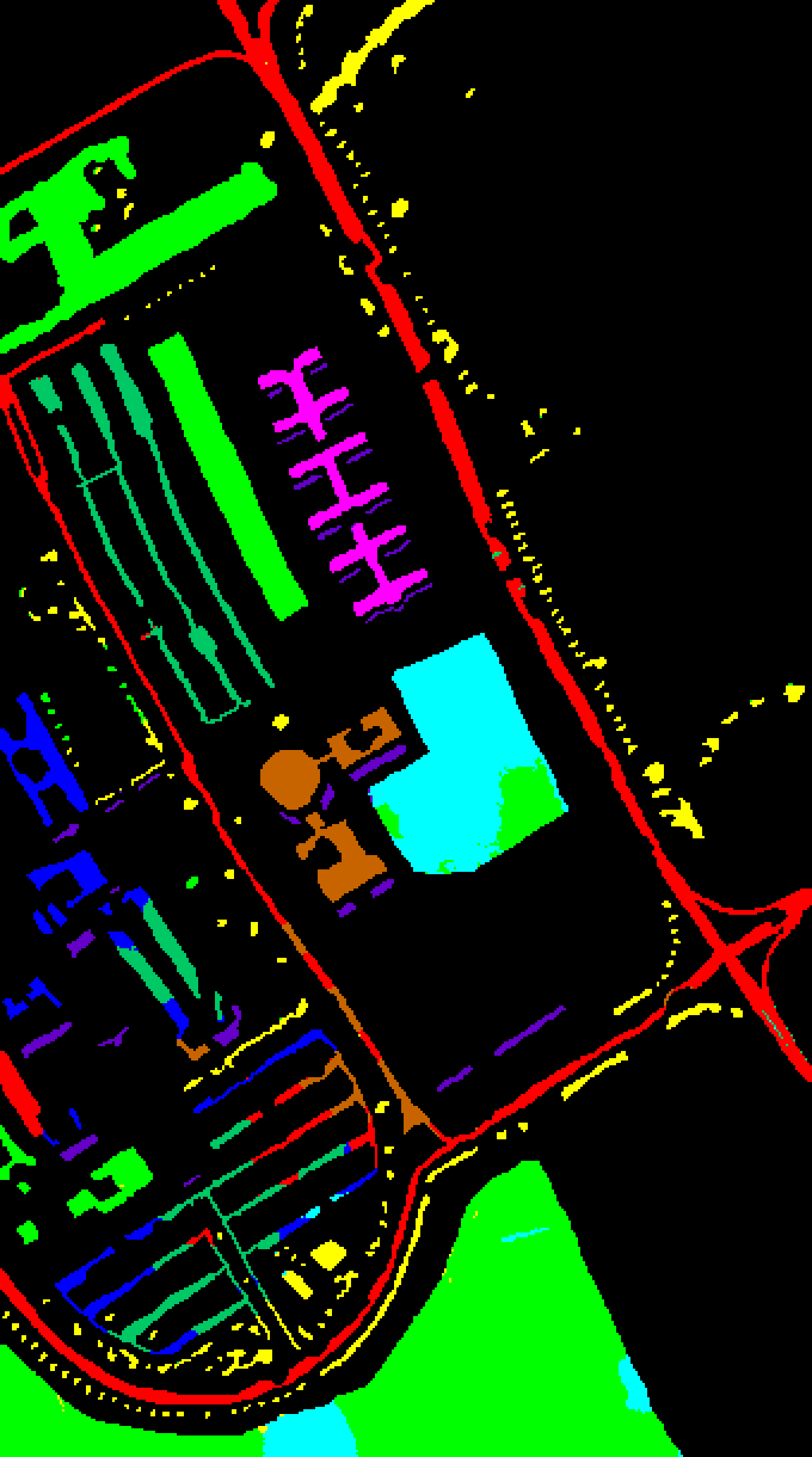}}
\hspace{3mm}
\subfloat[XDCL]{\includegraphics[width=1.0in]{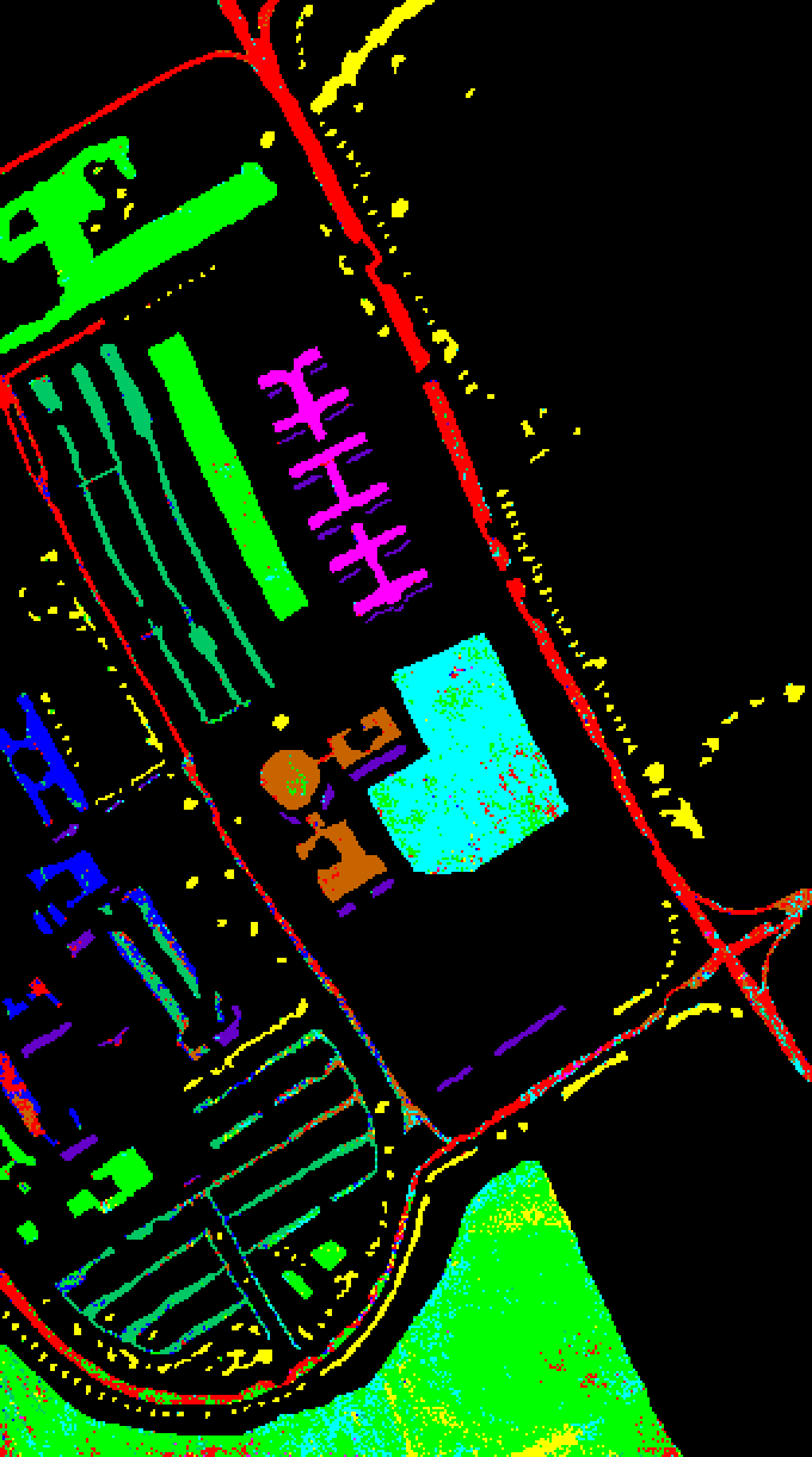}}
\hspace{3mm}
\subfloat[KnowCL]{\includegraphics[width=1.0in]{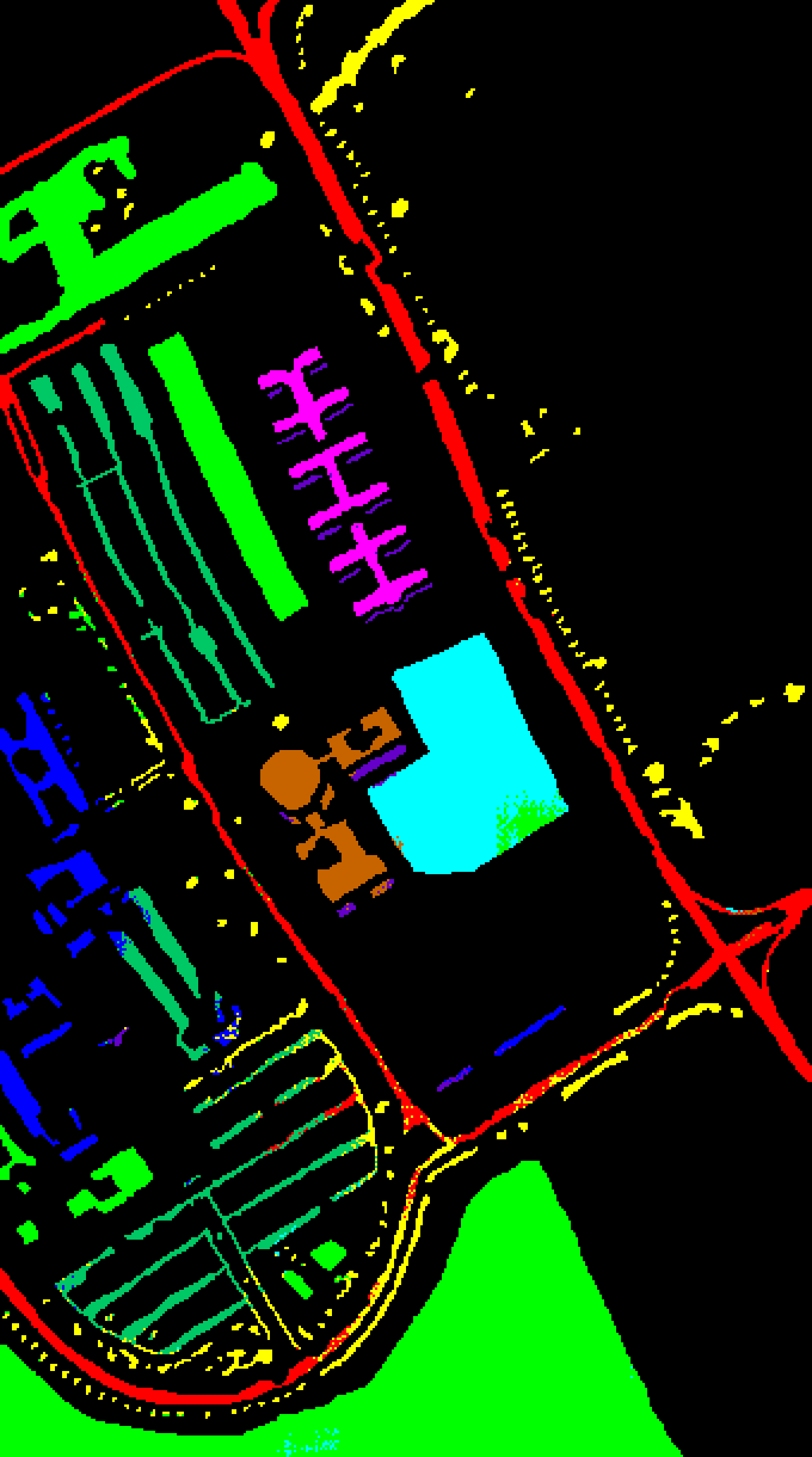}}
\hfil
\subfloat[False-color]{\includegraphics[width=1.02in]{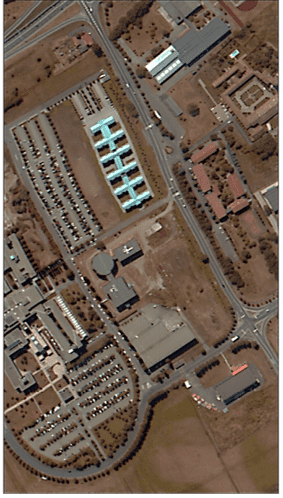}}
\hspace{3mm}
\subfloat[SVM]{\includegraphics[width=1.0in]{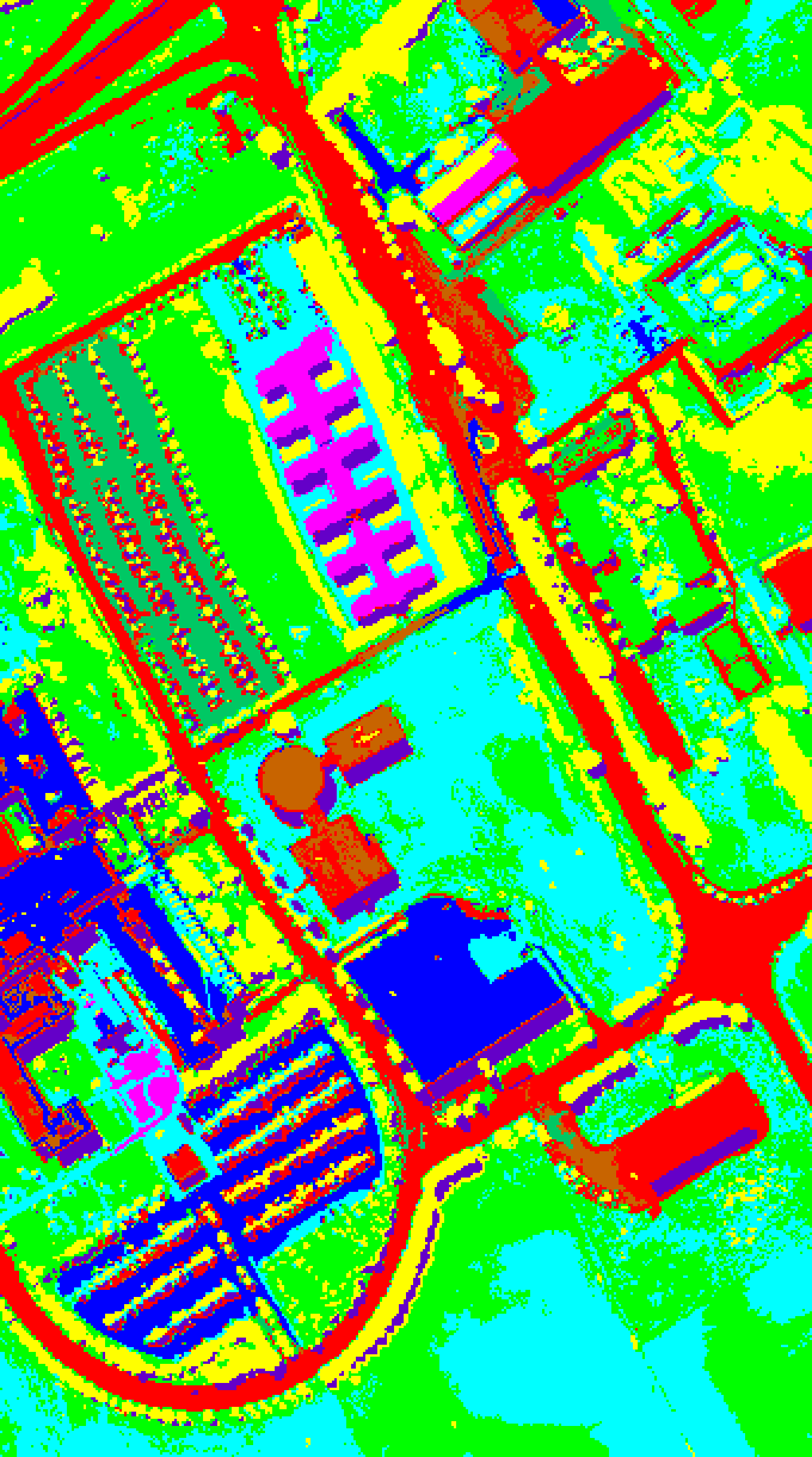}}
\hspace{3mm}
\subfloat[SpectralFormer]{\includegraphics[width=1.0in]{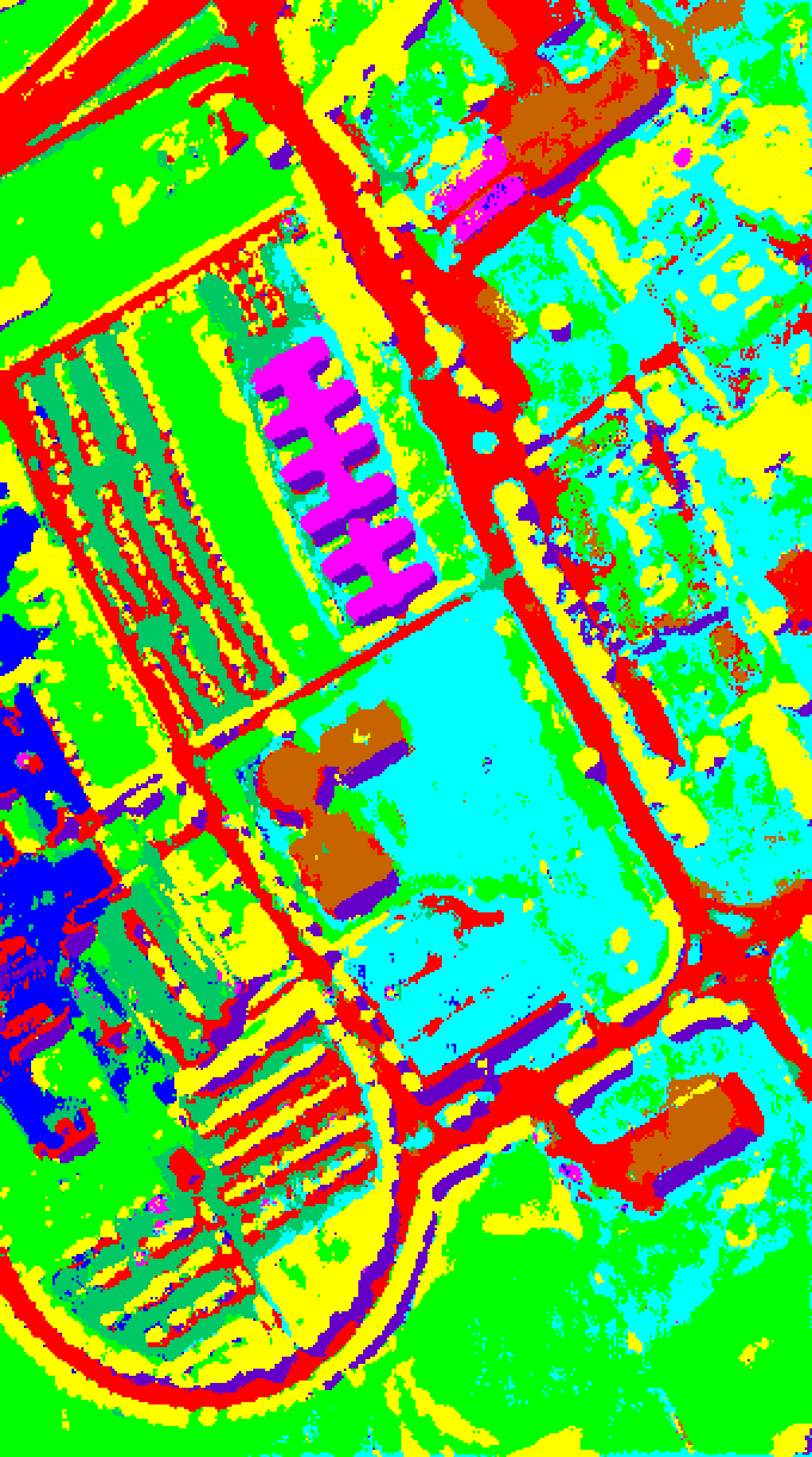}}
\hspace{3mm}
\subfloat[EMSGCN]{\includegraphics[width=1.0in]{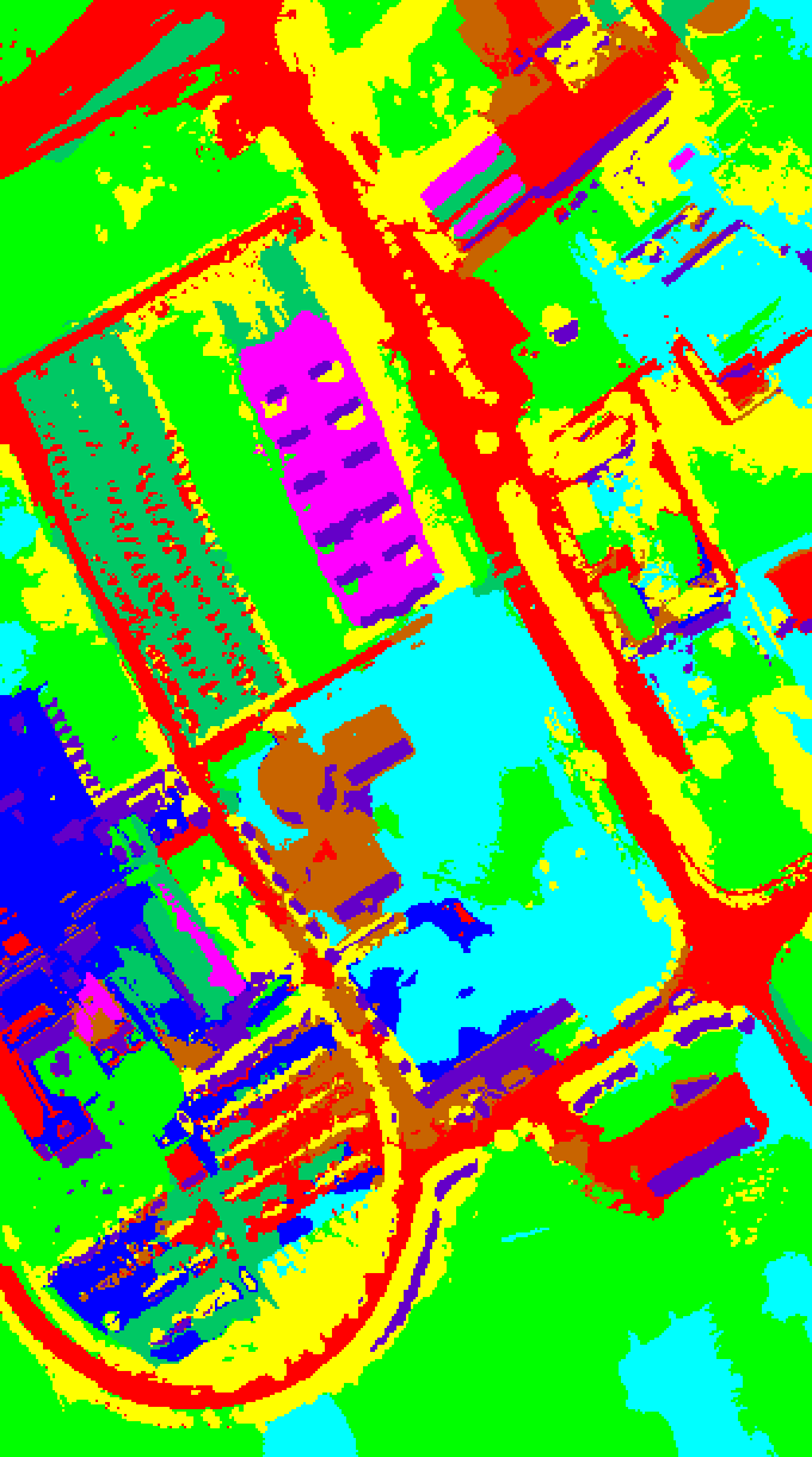}}
\hspace{3mm}
\subfloat[XDCL]{\includegraphics[width=1.0in]{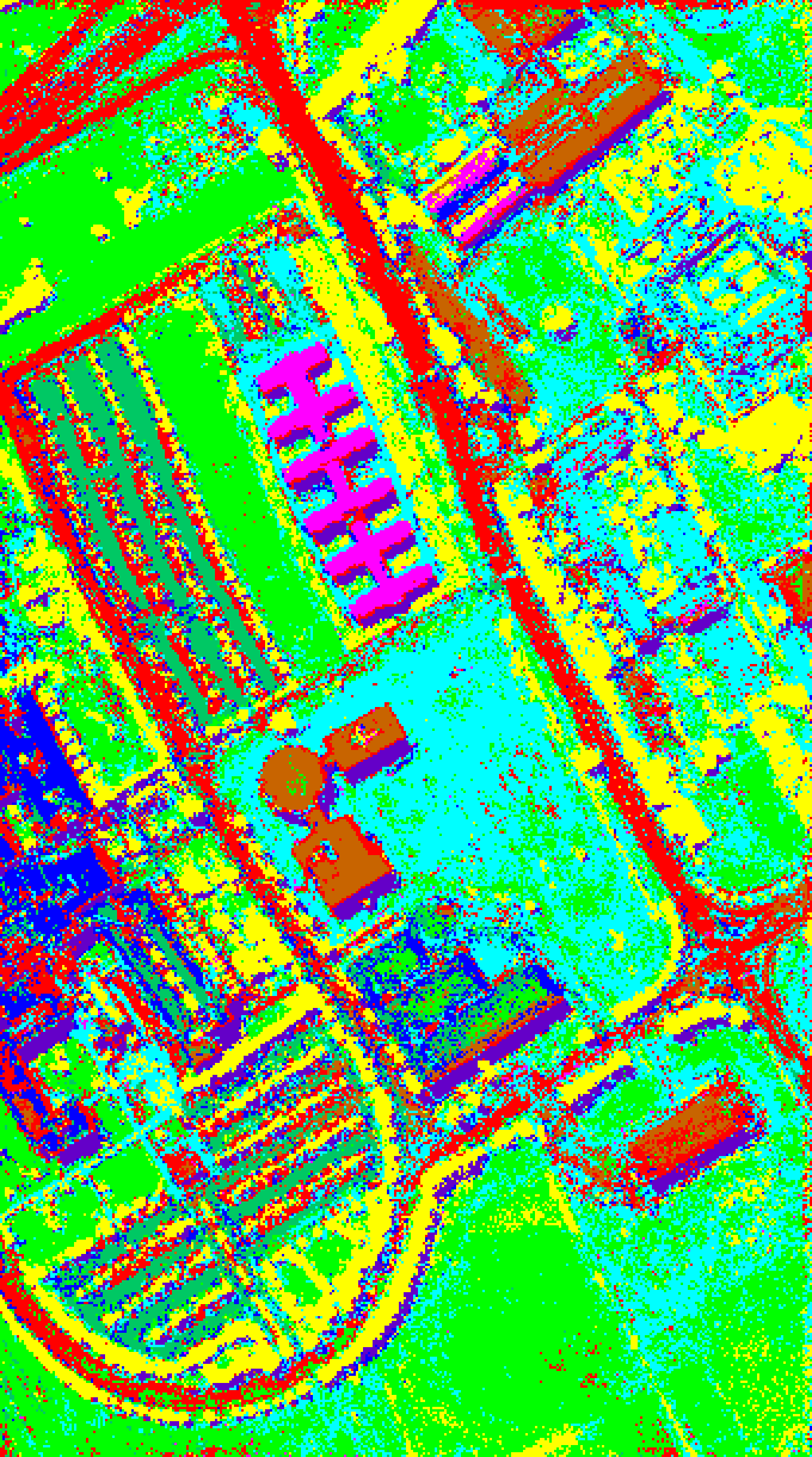}}
\hspace{3mm}
\subfloat[KnowCL]{\includegraphics[width=1.0in]{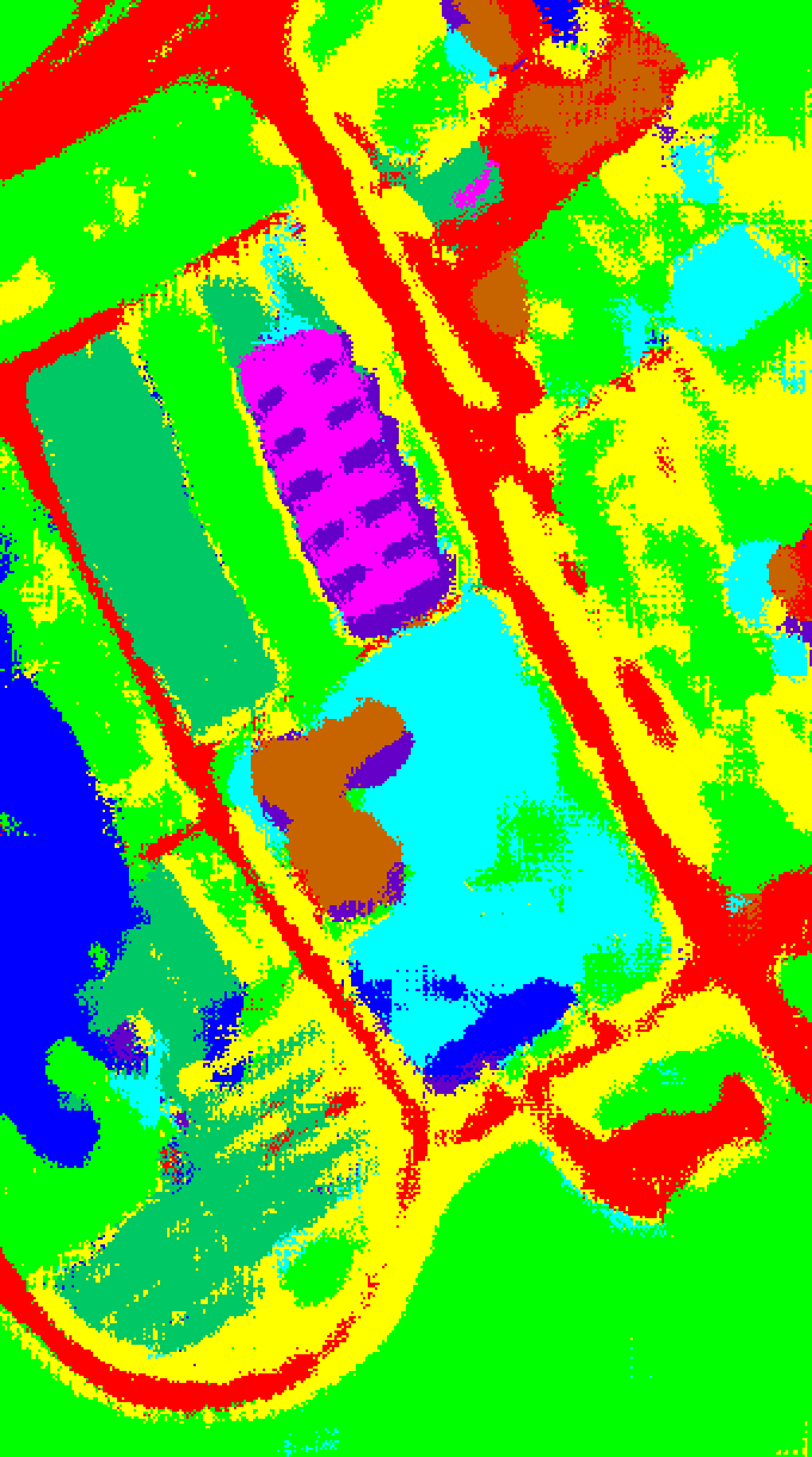}}
\caption{Ground truth (a), false-color (g), labeling area (b)-(f) and full image  (h)-(l) classification maps obtained by different models on the UP.}
\label{E_up}
\end{figure*}

\begin{table*}
\centering
\caption{Quantitative Comparison with Other Classification Methods on the Dioni Dataset.}
\label{tab:table18}
\begin{tabular}{c|c|llcccll}
\hline
&        Learning              &  Parameters& FLOPs&OA    & AA    & Kappa& Training time&Test time\\
\hline
EMP-SVM        &         -     &  -& -&77.98 & 69.26 & 72.97  & -&-\\
\hline
DFFN \cite{li2019dff}
& \multirow{5}{*}{Supervised Learning}&  435.45K& 140.31M&57.93 & 63.70 & 49.55  & 438.5s&71.79s\\
SPRN \cite{zhang2021spectral}
&                      &  176.45K& 63.43M&58.62 & 65.84 & 50.96  & 214.11s&46.96s\\
FDGC \cite{9785802}
&                      &  2.45M& 14.84M&62.27 & 75.96 & 54.16  & 101.89s&0.84s\\
SpectralFormer \cite{hong2021spectralformer}&         &  287.95K& 29.55M&74.27 & 74.69 & 68.76  & 389.69s&1.44s\\
 KnowCL-SU& &  918.09K& 19.58M
&79.37 & 78.12 &74.73 & 369.73s&1.66s\\
\hline
DMVL \cite{liu2020deep}
& \multirow{5}{*}{Unsupervised Learning}&  11.31M& 917.5M&60.62& 56.78 & 51.85  & 368.00min&2.52s\\

VAE \cite{cao2020unsupervised}
&                      &  824.03K& 3.23G&58.08 & 61.49 & 50.09  & 141.73min&13.38s\\
AAE \cite{cao2020unsupervised}
&                      &  758.37K& 3.22G&66.94 & 68.39 & 60.16  & 170.17min&14.54s\\

XDCL \cite{zhang2022cross}&                      &  6.67M& 19.65M&71.05 & 67.28 & 64.22  & 266.67s&1.03s\\
 KnowCL-US& & 585.66K& 19.25M& 75.62& 72.54& 69.95& 366.48s&1.21s\\           \hline
KnowCL(kNN)& \multirow{2}{*}{Semi-supervised Learning}  &  983.75K& 19.65M   &\textbf{82.52}& \textbf{83.18} & \textbf{78.33}  & 416.86s&1.41s\\
KnowCL(Linear evaluation)& &  983.75K& 19.65M   &81.63& 77.22& 77.42 & 17.6s&1.1s\\
\hline
\end{tabular}
\end{table*}

% Please add the following required packages to your document preamble:
% \usepackage{multirow}
\begin{table*}
\centering
\caption{Quantitative Comparison with Other Classification Methods on the DFC2018 Dataset.}
\label{tab:table19}
\begin{tabular}{c|c|llcccll}
\hline
&    Learning                  &   Parameters&FLOPs&OA    & AA    & Kappa& Training time&Test time\\
\hline
EMP-SVM        &         -    &   -&-&45.54 & 49.91 & 36.55  & -&-\\
\hline
DFFN \cite{li2019dff}
& \multirow{4}{*}{Supervised Learning}&   435.97K&4.84M&69.04 & 56.35 & 60.06  & 1487.61s&373.67s\\

SPRN \cite{zhang2021spectral}
&                      &   177.48K&63.43M&66.75 & 56.29 & 57.67  & 786.39s&237.95s\\
FDGC \cite{9785802}
&                      &   2.71M&15.22M&67.74 & 57.66 & 58.88  & 516.34s&42.08s\\

SpectralFormer \cite{hong2021spectralformer}&                      &   99.97K&4.84M&61.39 & 67.65 & 53.61  & 1870.95s&43.02s\\
 KnowCL-SU& &   934.26K&20.09M
&69.47 & 54.04 & 61.58 & 1351.20s&41.22s\\
\hline
DMVL \cite{liu2020deep}
& \multirow{4}{*}{Unsupervised Learning}&   11.31M&917.5M&59.16& 54.08 & 49.77  & 669.11min&482.82s\\

VAE \cite{cao2020unsupervised}
&                      &   824.03K&25.2M&59.55 & 57.45 & 50.75  & 160.25min&5577.63s\\
AAE \cite{cao2020unsupervised}
&                      &   758.38K&25.1M&46.56 & 53.26 & 39.09  & 190.59min&12738.13s\\

XDCL \cite{zhang2022cross}&                      &   6.6M&20.16M&37.92 & 47.58 & 30.43  & 1708.5min&41.6s\\
 KnowCL-US& & 599.77K& 19.76M
& 62.99 & 52.37 & 54.41 & 1277.37s&36.31s\\
\hline
KnowCL(kNN)& \multirow{2}{*}{Semi-supervised Learning} &   999.92K  &20.16M  &73.98& 60.6  & 66.42  & 2728.18s&43.63s\\
KnowCL(Linear Evaluation) & &   999.92K  &20.16M  &\textbf{74.08}& \textbf{68.41}& \textbf{66.67} & 70.21s&45.09s\\
\hline
\end{tabular}
\end{table*}
Deep models have many hyperparameters to be tuned, and these hyperparameters can drastically affect the performance of the model. Conventional hyperparameters such as learning rate, weight decay coefficient, number of hidden layer units, loss function, etc. have been widely and deeply studied \cite{dong2018hyperparameter}. In this study, we focus on four hyperparameters that affect this model. Crop size and batch size affect the overall performance of the model by affecting the feature extraction of the model. The number of bands and k-value of KNN are mainly affected by the dataset. Therefore, we tested the effect of crop size and batch size on the model using the Huston 2018 dataset as a benchmark. The effect of the number of bands and k-value on the model was investigated in each of the four data.

\subsubsection{Crop size}

Data augmentation affects the quality of unsupervised feature extraction. Chen et al. \cite{chen2020simple} investigated the effect of different compositions of transformations by the linear evaluation and pointed out that random cropping and random color distortion stand out. Since HSIs do not have the concept of color as in natural images, we only tested the effect of different crop sizes on the HSI classification. Table \ref{tab:table3} shows that the model performs robustly when the images are cropped to different sizes.

\subsubsection{Batch size}

Table \ref{tab:table4} shows the impact of batch size when models are trained on the DFC2018. Unlike the simCLR \cite{chen2020simple}, model performance does not improve with increasing batch size. On the one hand, it may be due to the fact that the training data involved in the training of the model are fewer, and they are taken from several regions with more similar features so global features can be learned with a smaller batch size. On the other hand, it suggests that the insertion of labeling knowledge allows the model to learn task-specific knowledge, giving the model enough power to find the optimal feature representation under a small batch size.

\subsubsection{Bands}

HSIs contain bands far beyond those of natural images. The previous work either directly input all the bands or took only 3 bands without considering the effect of different sensors and feature types on the classification performance. However, the spectral correlation affects its classification performance. PCA, a simple dimensionality reduction method is applied to remove the noise interference and extract the most discriminative feature. Table \ref{tab:table5} shows the impact of the number of principal components when the model is trained on the four datasets. We find that when the number of principal components increases, the performance of the model will have a great difference. The overall presentation is an inverse U-shape. The results manifest that different hyperspectral sensors and scenes extract different principal components to achieve the best classification performance.

\subsubsection{The value of k}

KNN is a simple but powerful non-parametric algorithm, which is often used to evaluate the capability of unsupervised feature extraction. Fig. \ref{KNN} shows kNN's evaluation results using different $k$ values for the framework. When different $k$ values are used, similar results are observed on the UP and Salinas. Different $k$ values show opposite trends on the Dinoni and DFC2018 datasets. A slightly large $k$ value will be better on Dinoni, while larger $k$ values are detrimental to DFC2018.

To explore the reason for this, we leverage the t-SNE method to visualize the features extracted by KnowCL for quantitative analysis, as shown in Fig. \ref{tsne}. We can see that due to the purity of the land cover in the first three datasets are more homogeneous and clustered, the features extracted by the model are more separated. Therefore, the model accuracy remains unchanged or improves a little as the $k$ value increases. The DFC2018 has more categories, and the extracted features are superimposed together, resulting in a lower accuracy as more neighboring nodes are taken into account. 

% We conjecture that this difference is due to the purity of the land cover in the images. The land cover classes in the first three datasets are more homogeneous and clustered. The features extracted by the model are more separated, so the model accuracy remains unchanged or improves a little as the k value increases. The DFC2018   has more categories, and the extracted features are superimposed together, resulting in a lower accuracy as more neighboring nodes are taken into account. 

\subsection{Quantitative Comparison}

\begin{figure*}[!t]
\centering
\subfloat[Ground Truth]{\includegraphics[width=1.0in]{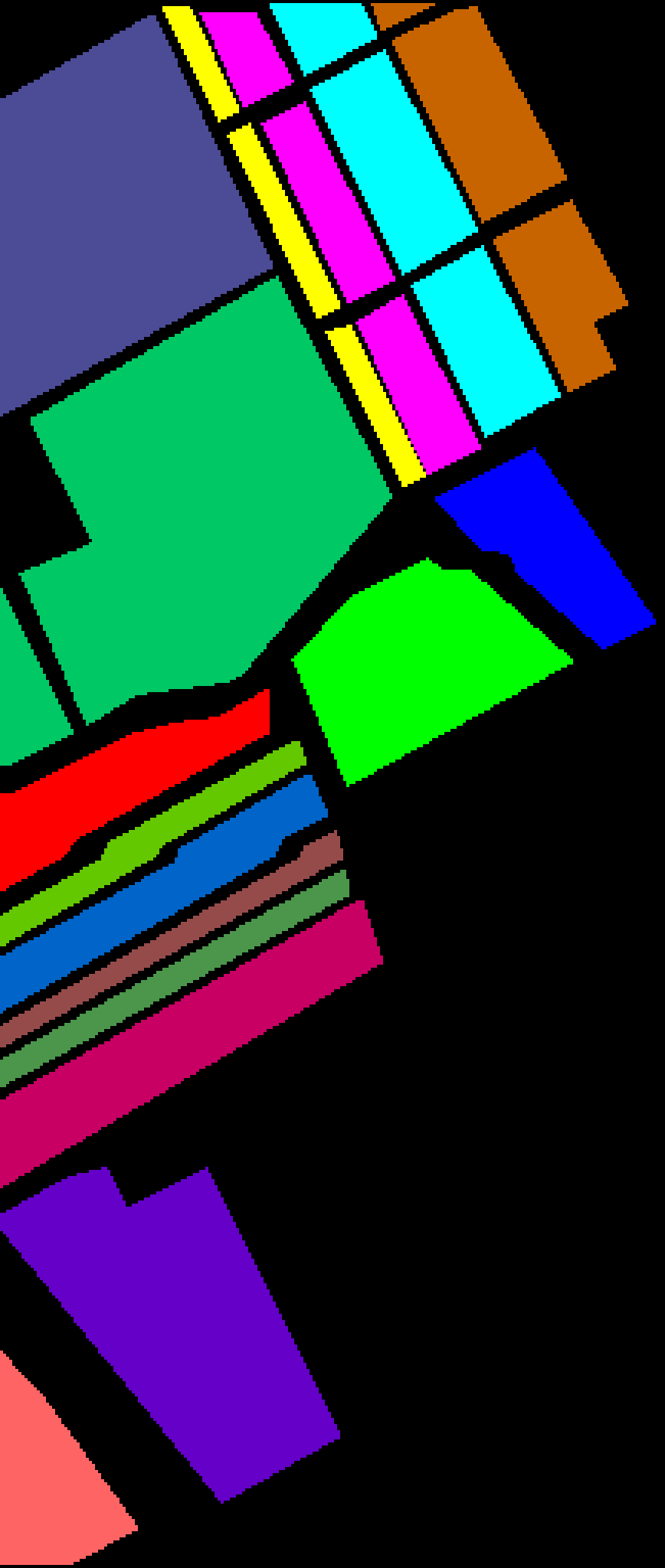}}
\hspace{3mm}
\subfloat[SVM]{\includegraphics[width=1.0in]{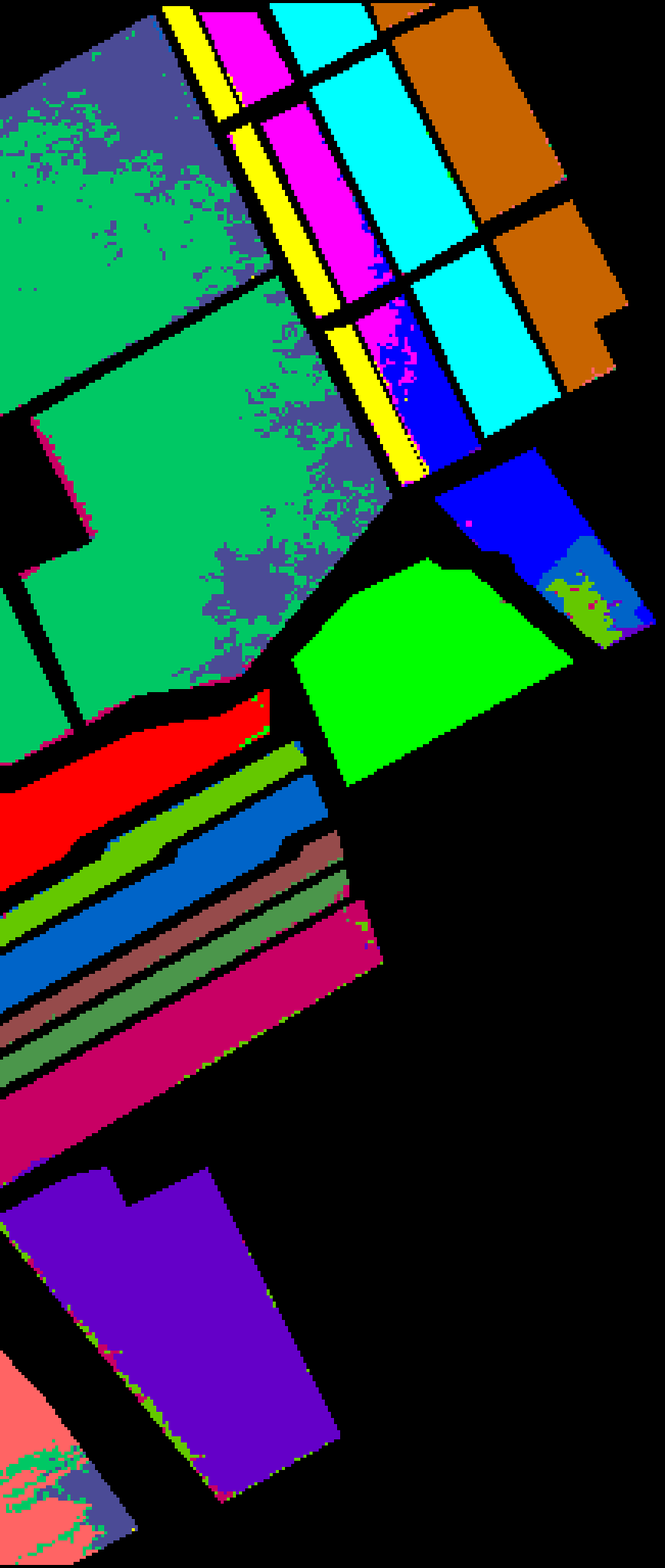}}
\hspace{3mm}
\subfloat[SpectralFormer]{\includegraphics[width=1.0in]{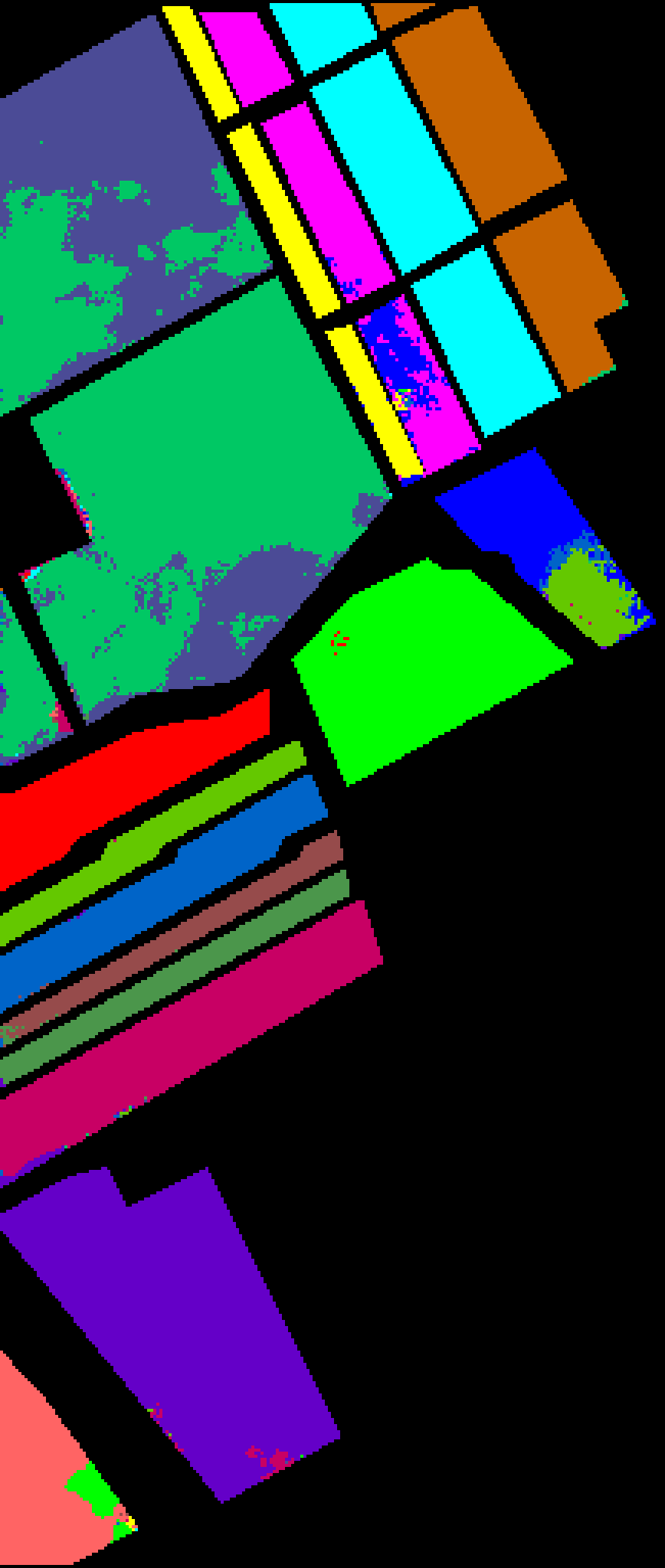}}
\hspace{3mm}
\subfloat[EMSGCN]{\includegraphics[width=1.0in]{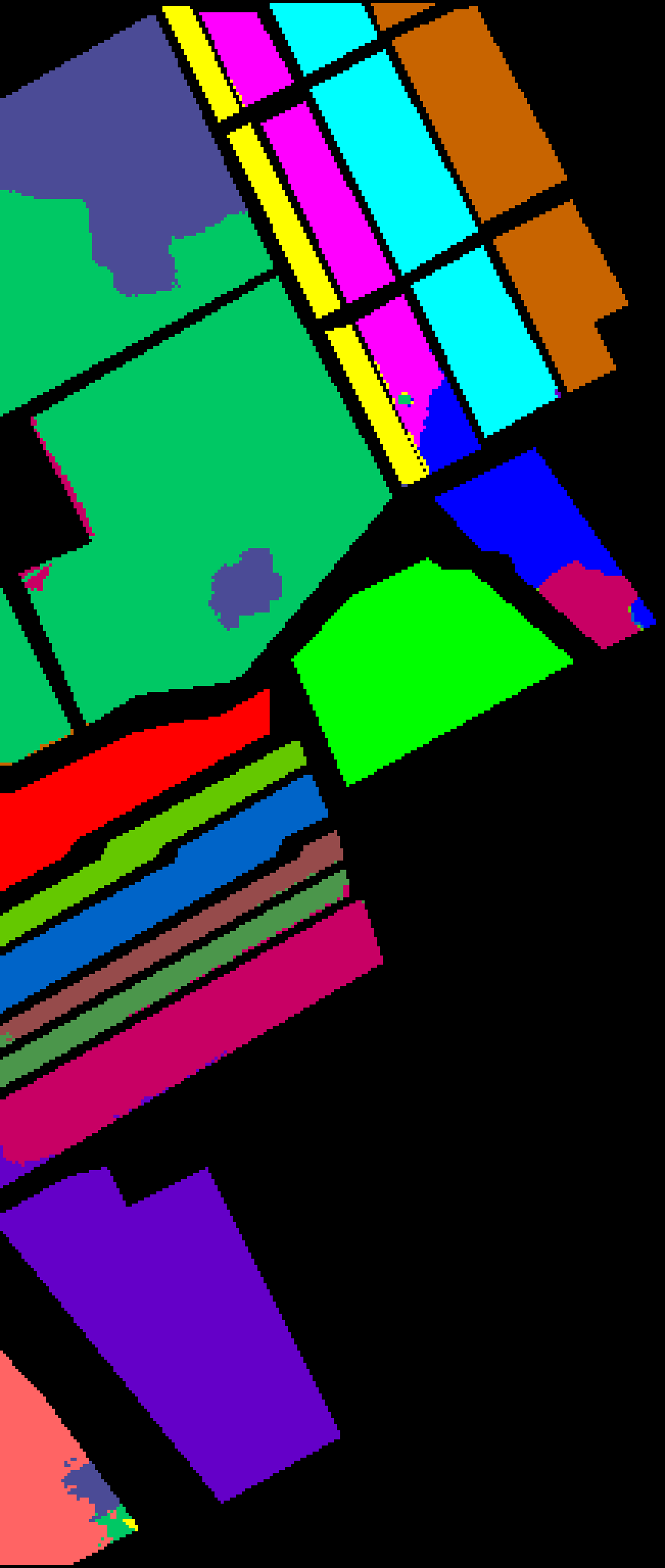}}
\hspace{3mm}
\subfloat[XDCL]{\includegraphics[width=1.0in]{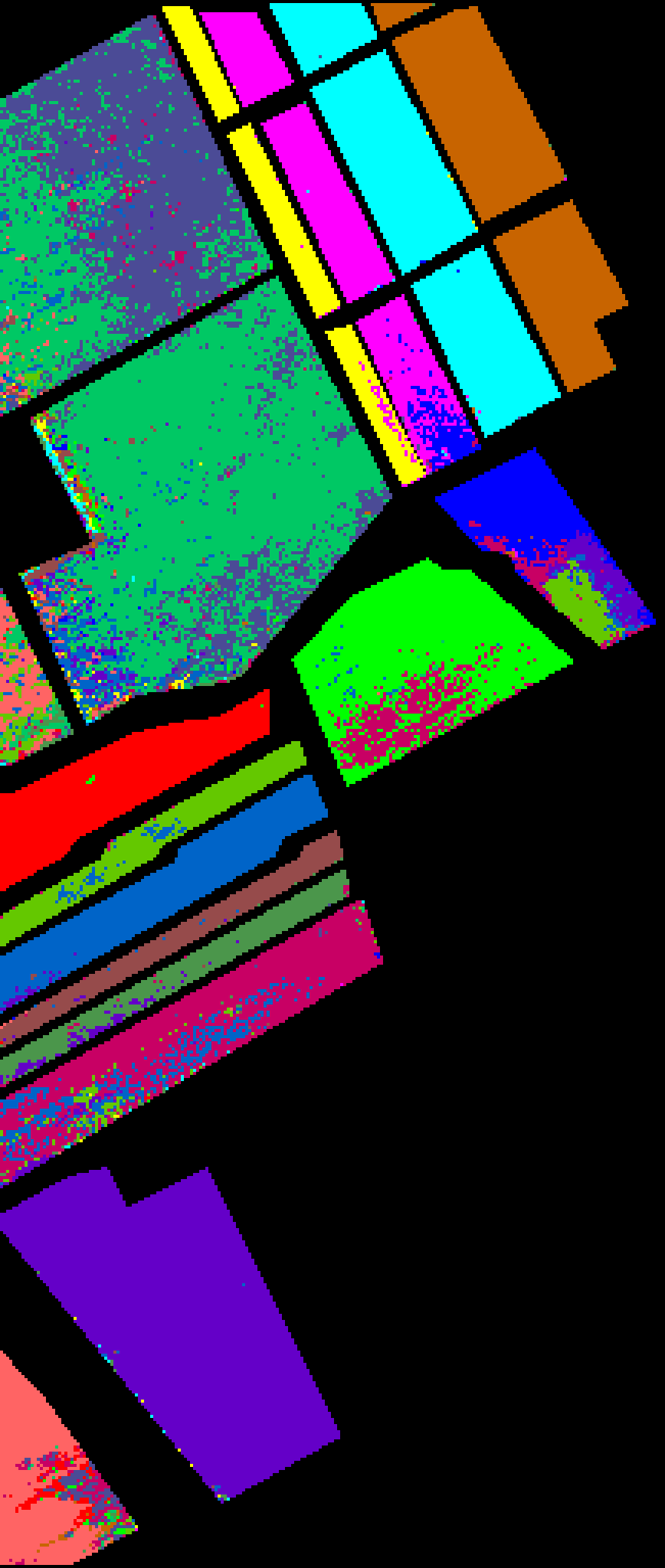}}
\hspace{3mm}
\subfloat[KnowCL]{\includegraphics[width=1.0in]{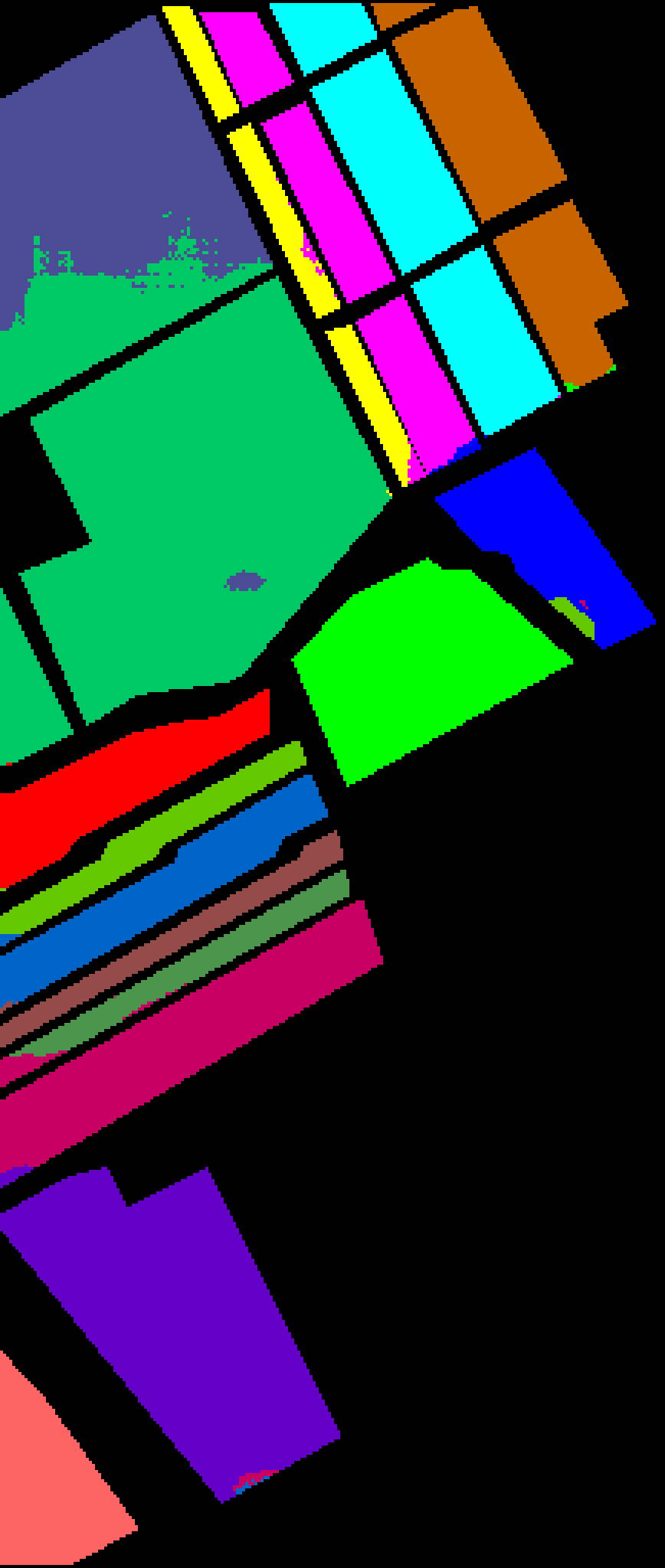}}
\hfil
\subfloat[False-color]{\includegraphics[width=1.02in]{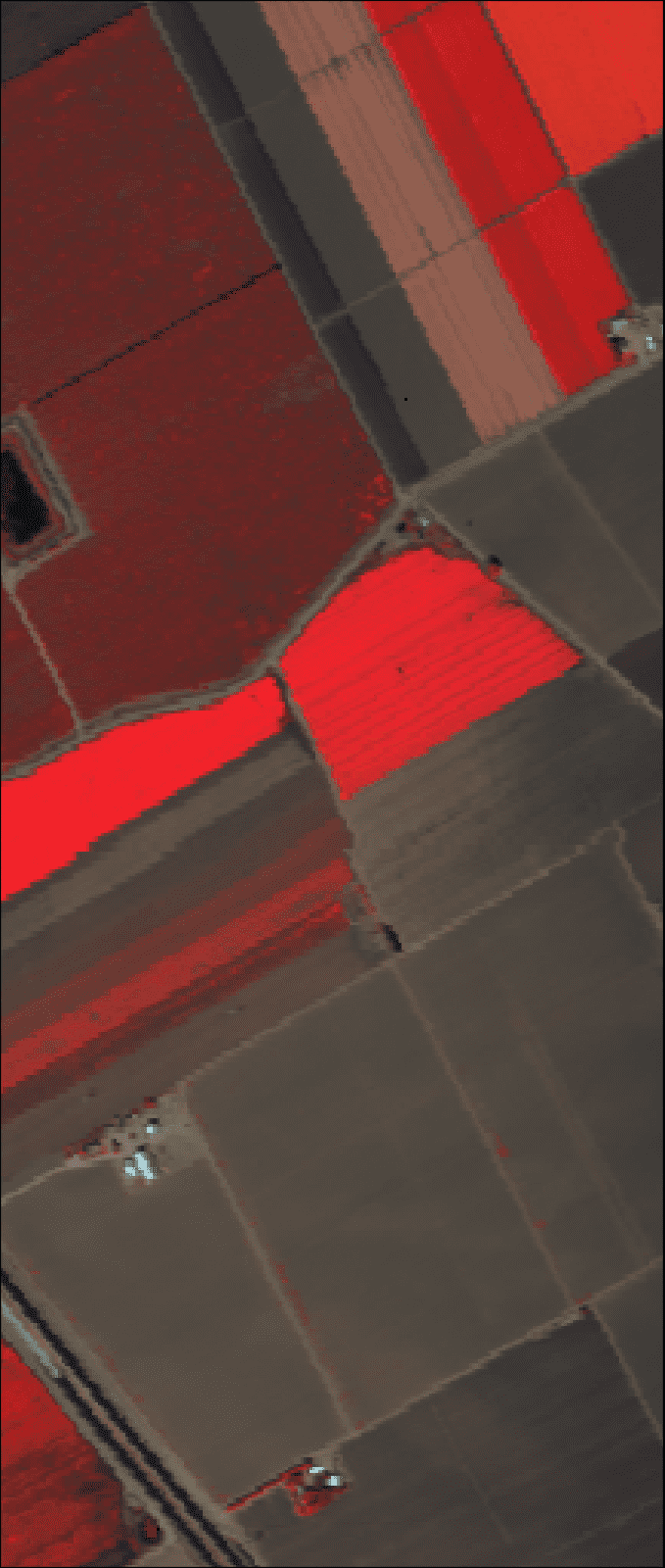}}
\hspace{3mm}
\subfloat[SVM]{\includegraphics[width=1.0in]{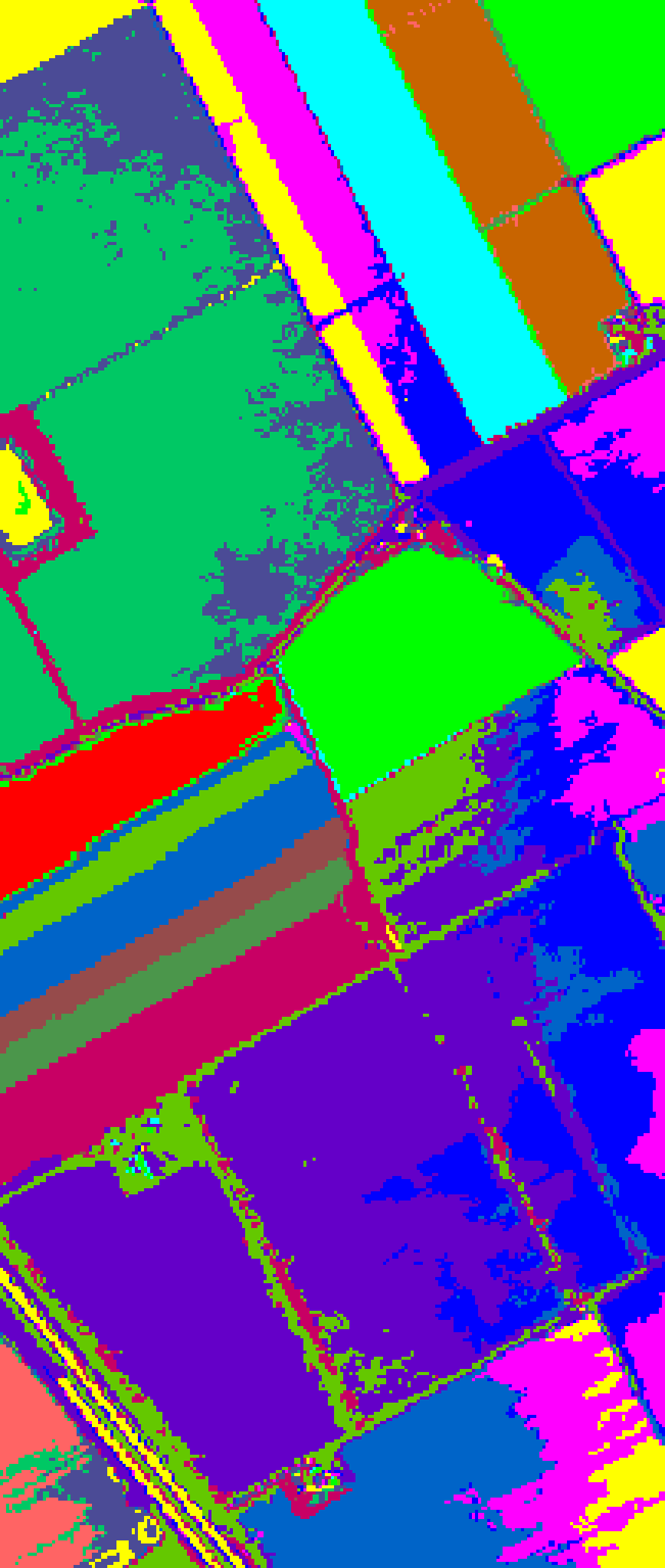}}
\hspace{3mm}
\subfloat[SpectralFormer]{\includegraphics[width=1.0in]{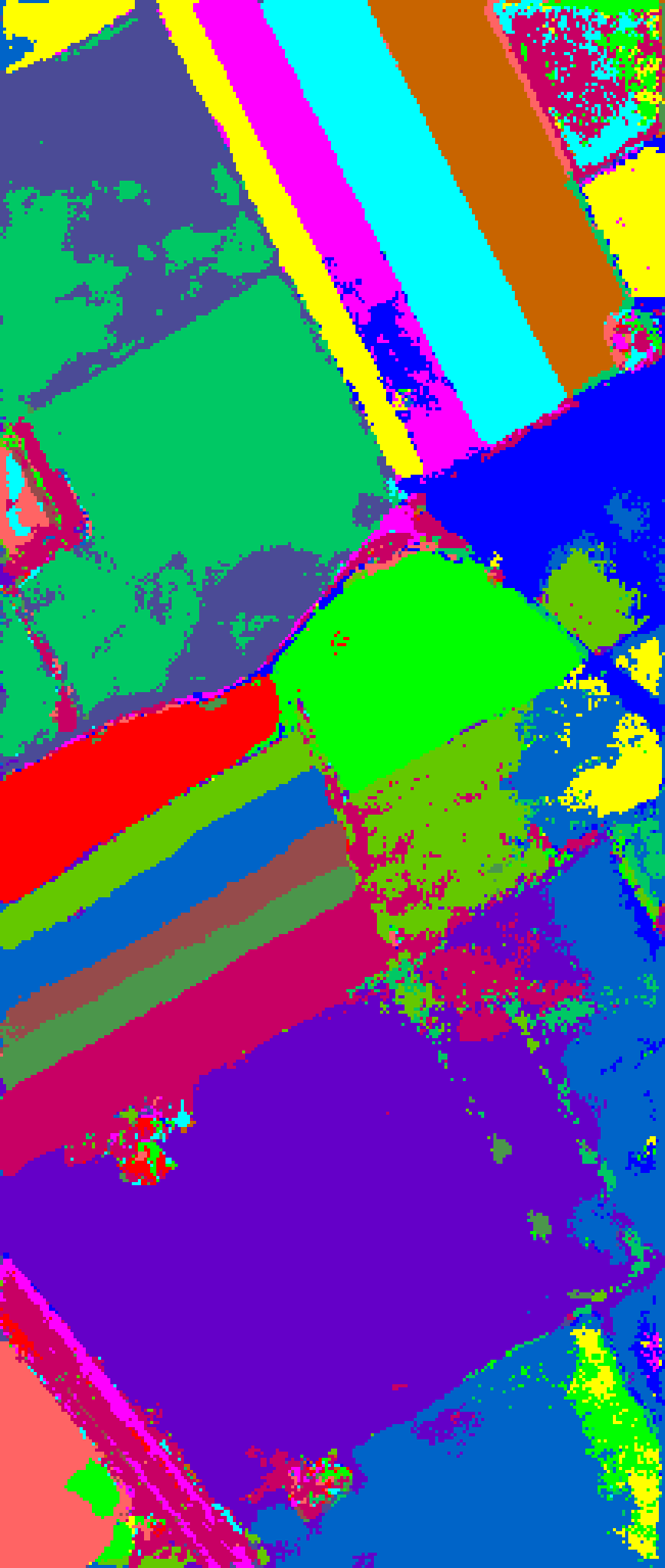}}
\hspace{3mm}
\subfloat[EMSGCN]{\includegraphics[width=1.0in]{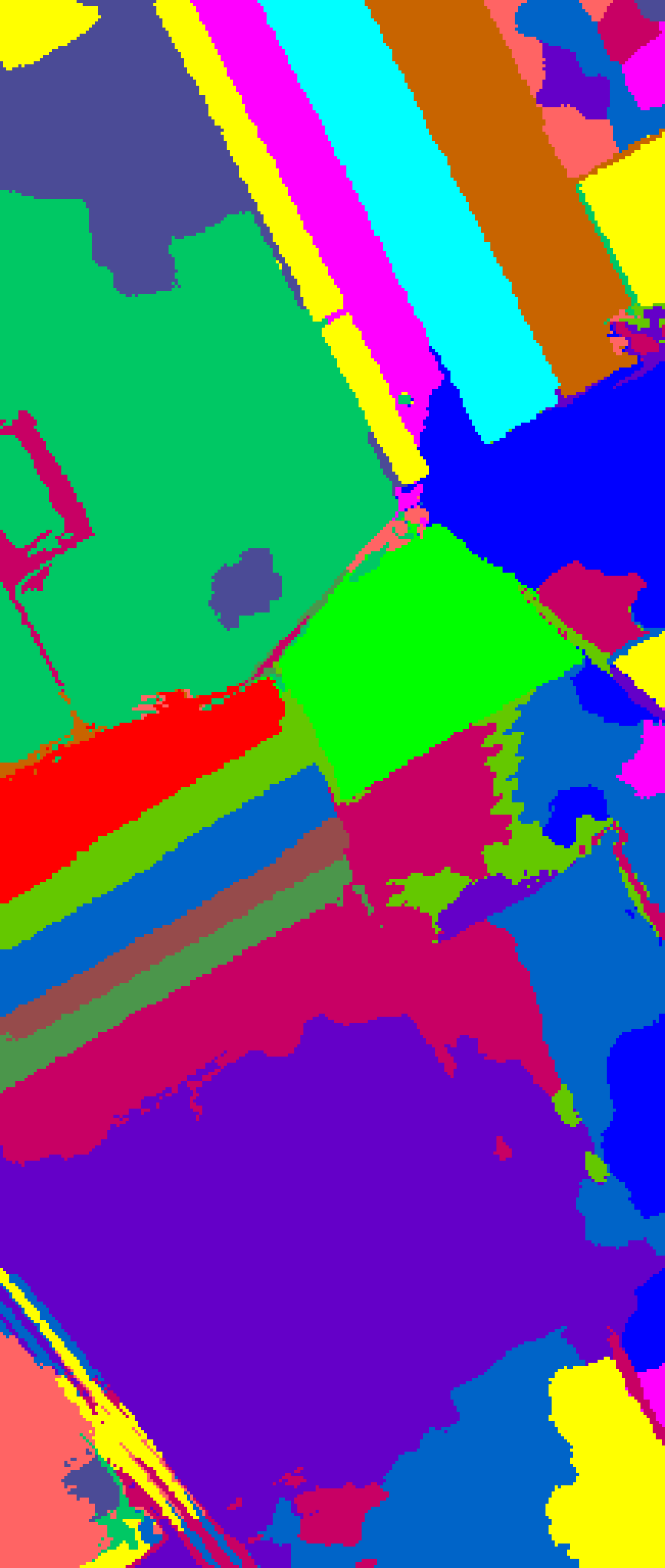}}
\hspace{3mm}
\subfloat[XDCL]{\includegraphics[width=1.0in]{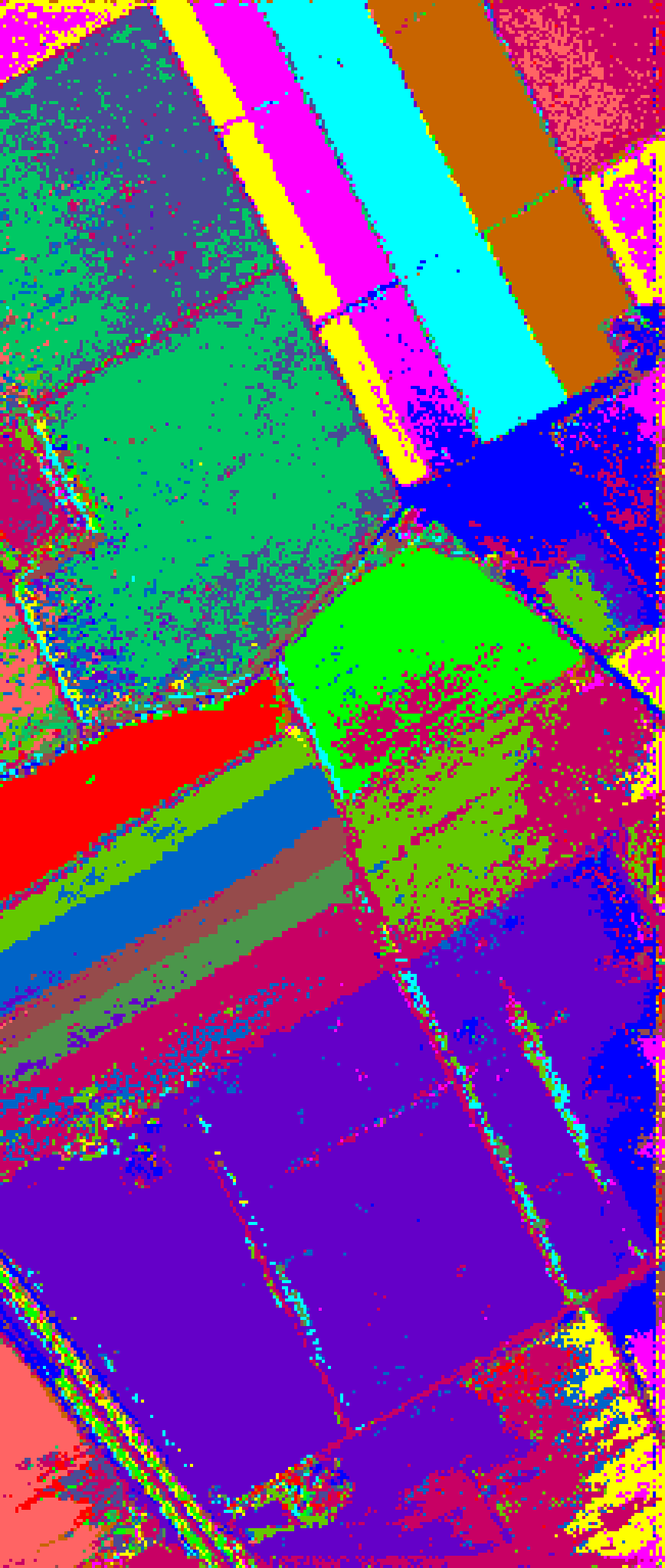}}
\hspace{3mm}
\subfloat[KnowCL]{\includegraphics[width=1.0in]{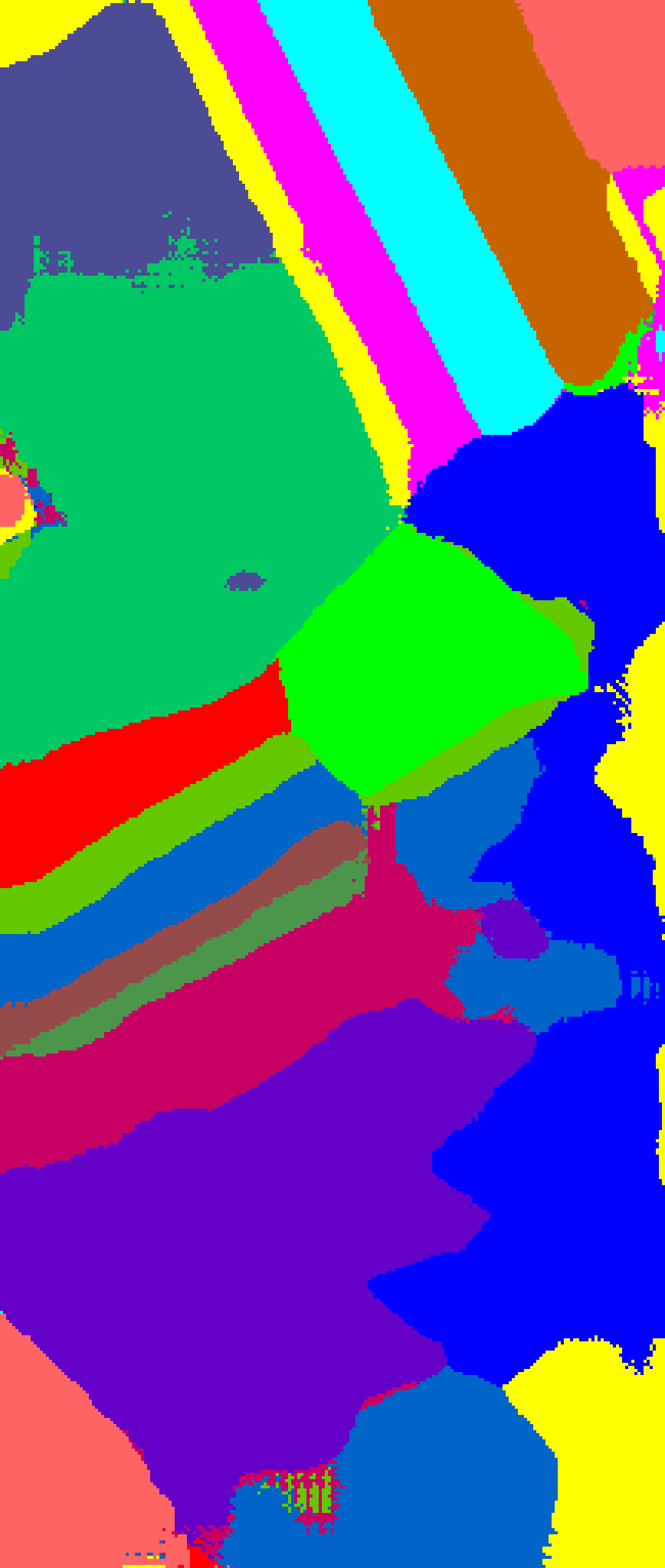}}
\caption{Ground truth (a), false-color (g), labeling area (b)-(f) and full image  (h)-(l) classification maps obtained by different models on the Salinas.}
\label{E_sa}
\end{figure*}

\begin{figure*}[!t]
\centering
\subfloat[Ground Truth]{\includegraphics[width=3.4in]{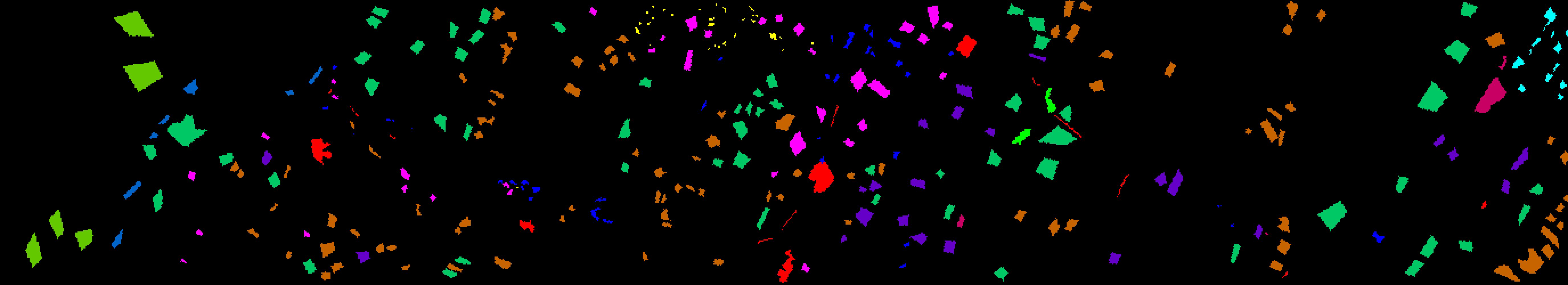}}
\hspace{1mm}
\subfloat[False-color]{\includegraphics[width=3.4in]{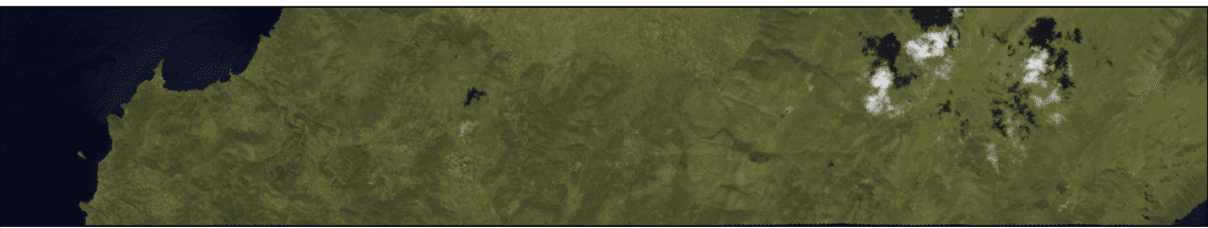}}
\hfil
\subfloat[SVM]{\includegraphics[width=3.4in]{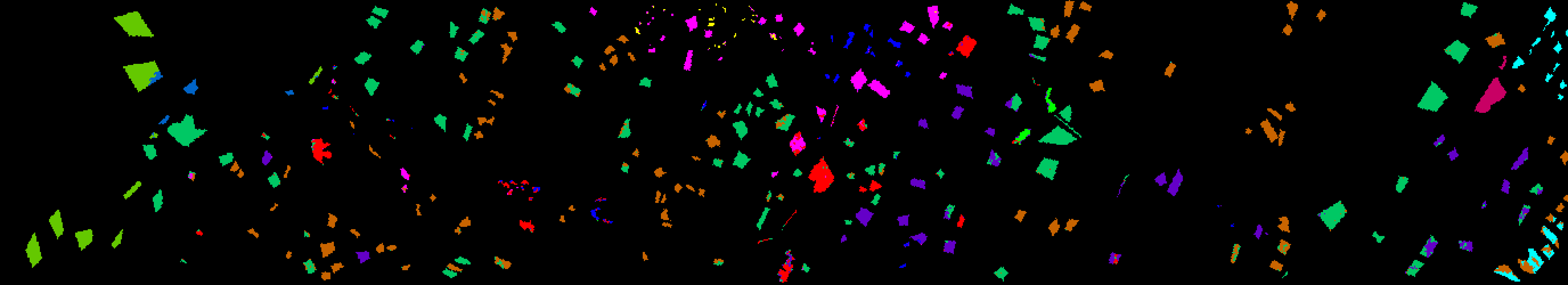}}
\hspace{1mm}
\subfloat[SVM]{\includegraphics[width=3.4in]{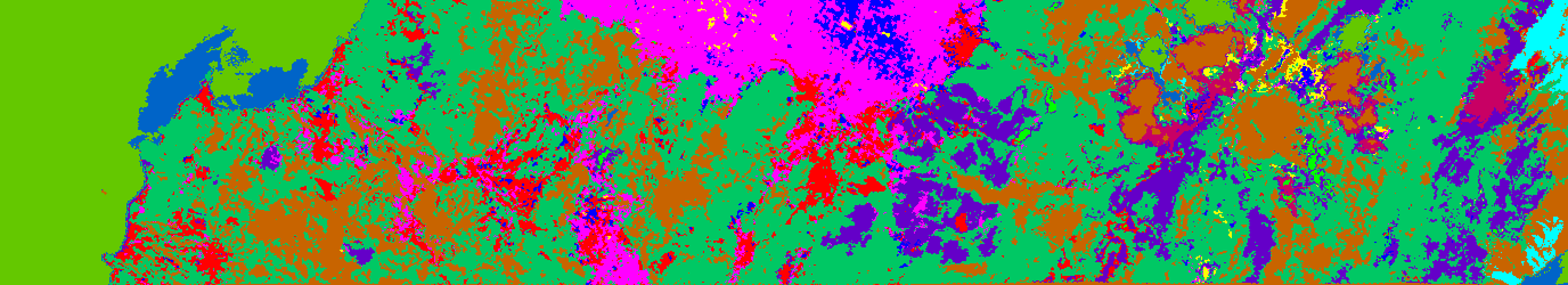}}
\hfil
\subfloat[SpectralFormer]{\includegraphics[width=3.4in]{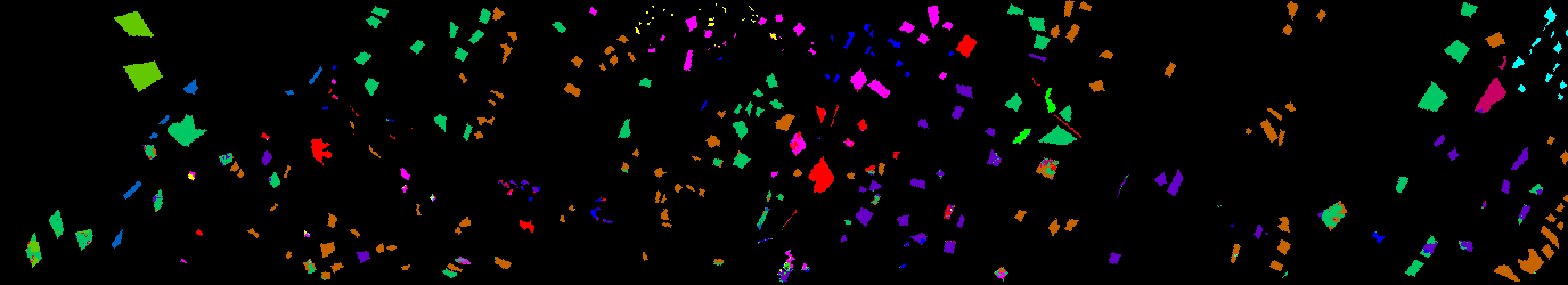}}
\hspace{1mm}
\subfloat[SpectralFormer]{\includegraphics[width=3.4in]{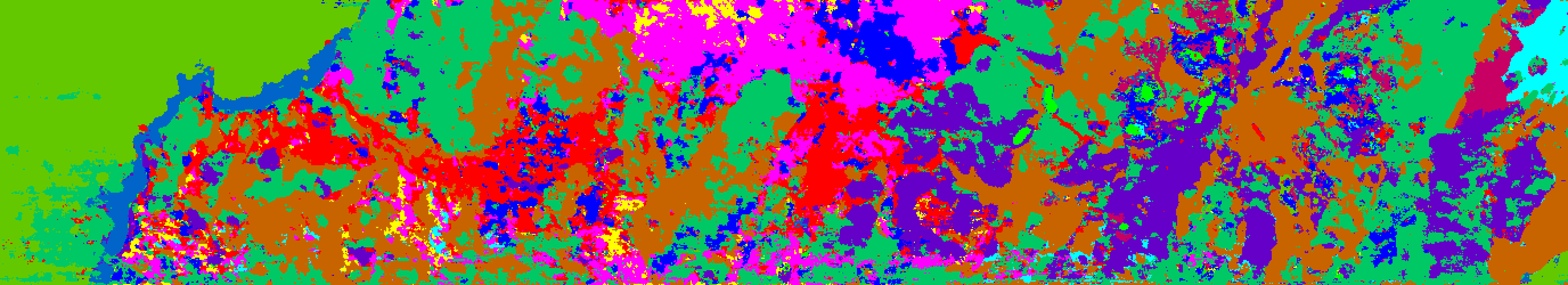}}
\hfil
\subfloat[XDCL]{\includegraphics[width=3.4in]{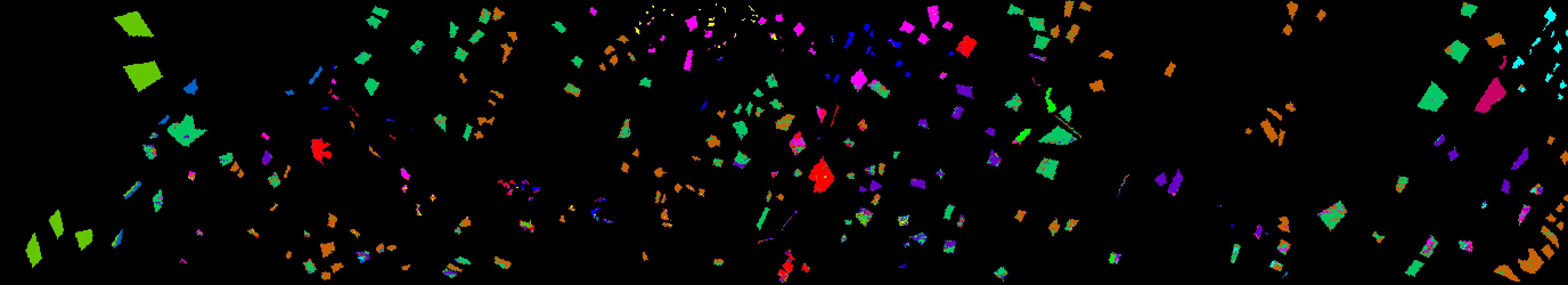}}
\hspace{1mm}
\subfloat[XDCL]{\includegraphics[width=3.4in]{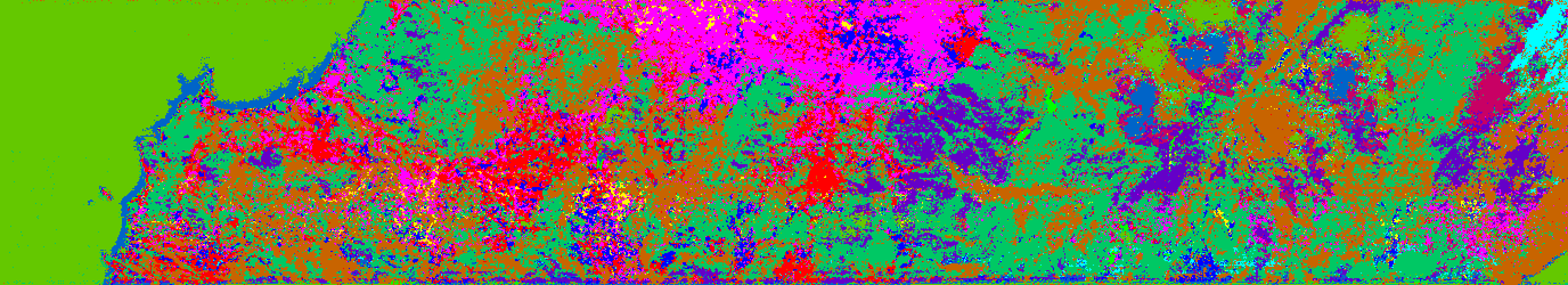}}
\hfil
\subfloat[KnowCL]{\includegraphics[width=3.4in]{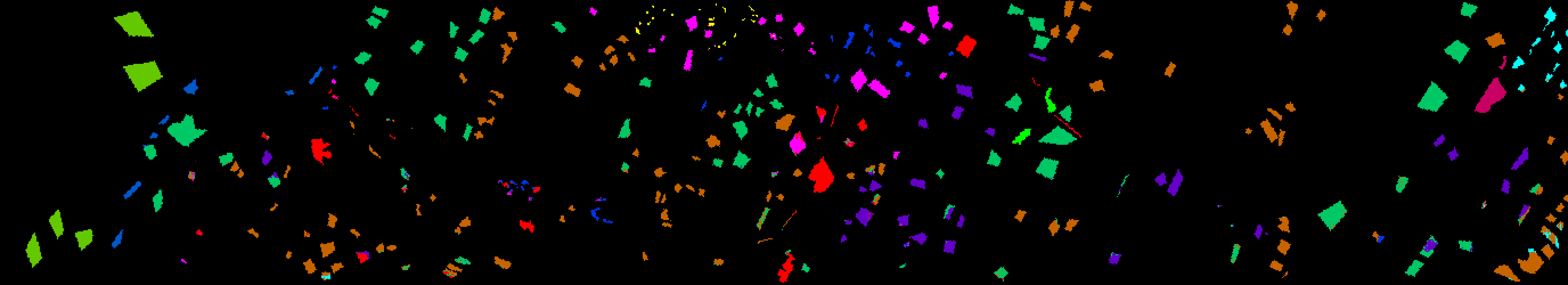}}
\hspace{1mm}
\subfloat[KnowCL]{\includegraphics[width=3.4in]{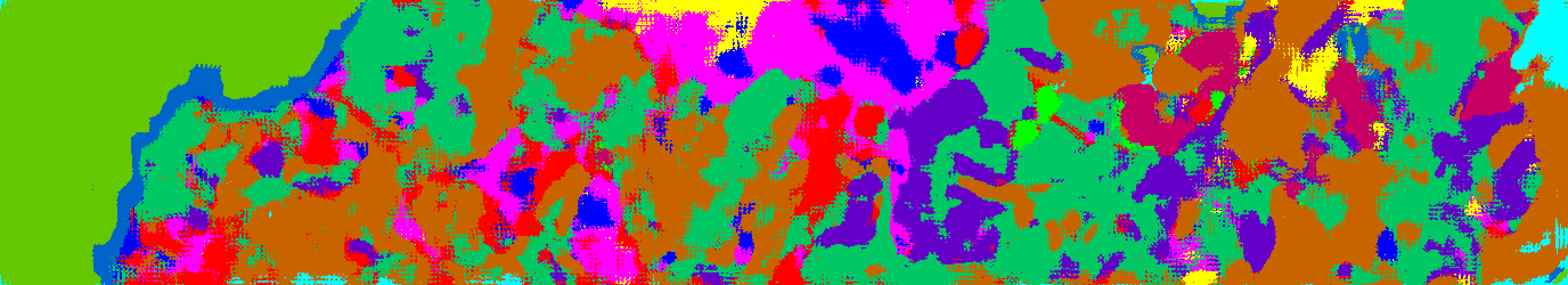}}
\caption{Ground truth (a), false-color (b), labeling area (c)-(i) and full image  (d)-(j) classification maps obtained by different models on the Dioni.}
\label{E_Dinoni}
\end{figure*}

\begin{figure*}[!t]
\centering
\subfloat[Ground Truth]{\includegraphics[width=3.3in]{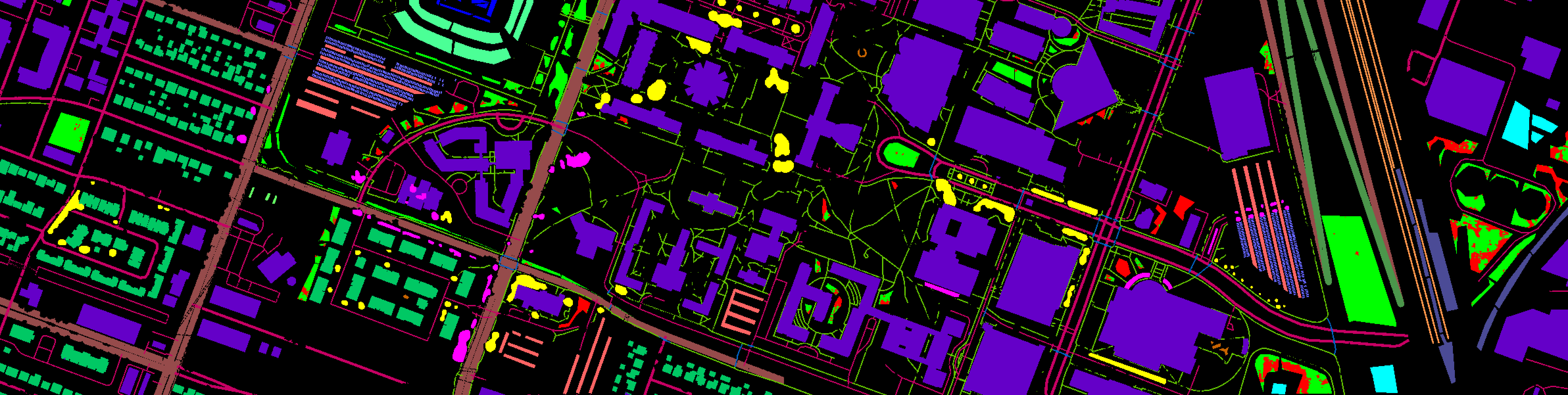}}
\hspace{1mm}
\subfloat[False-color]{\includegraphics[width=3.3in,height=0.85in]{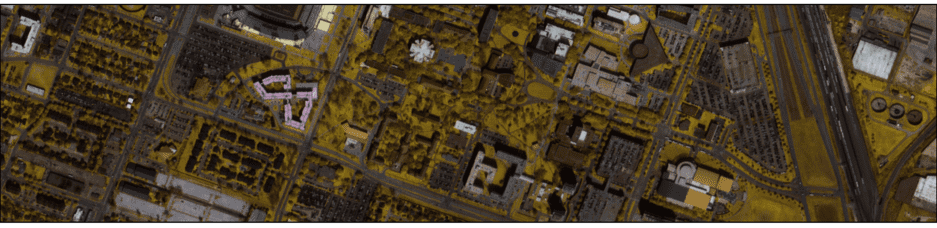}}
\hfil
\subfloat[SVM]{\includegraphics[width=3.3in]{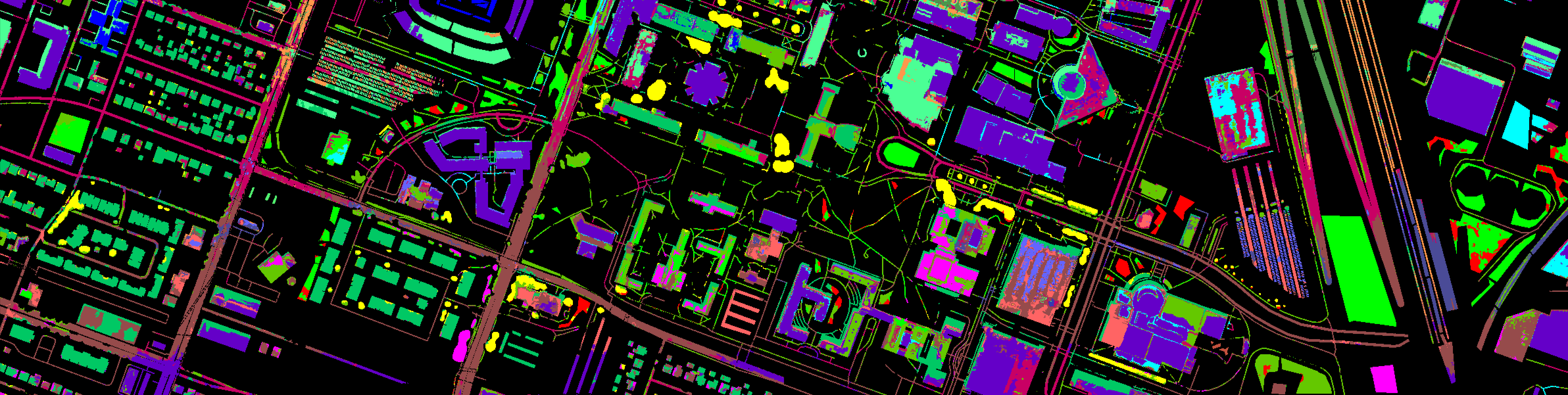}}
\hspace{1mm}
\subfloat[SVM]{\includegraphics[width=3.3in]{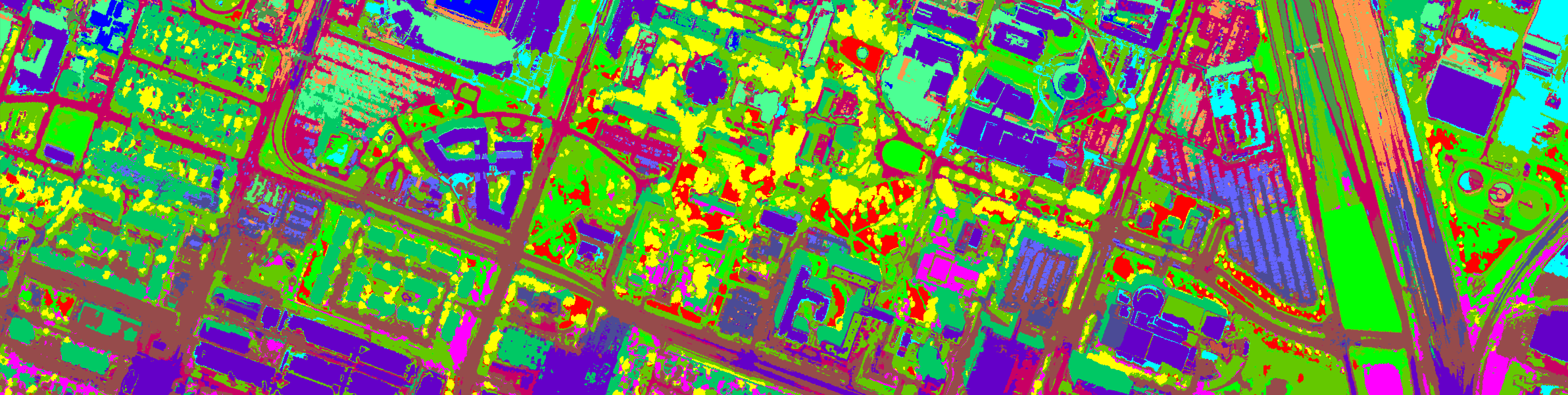}}
\hfil
\subfloat[SepctralFormer]{\includegraphics[width=3.3in]{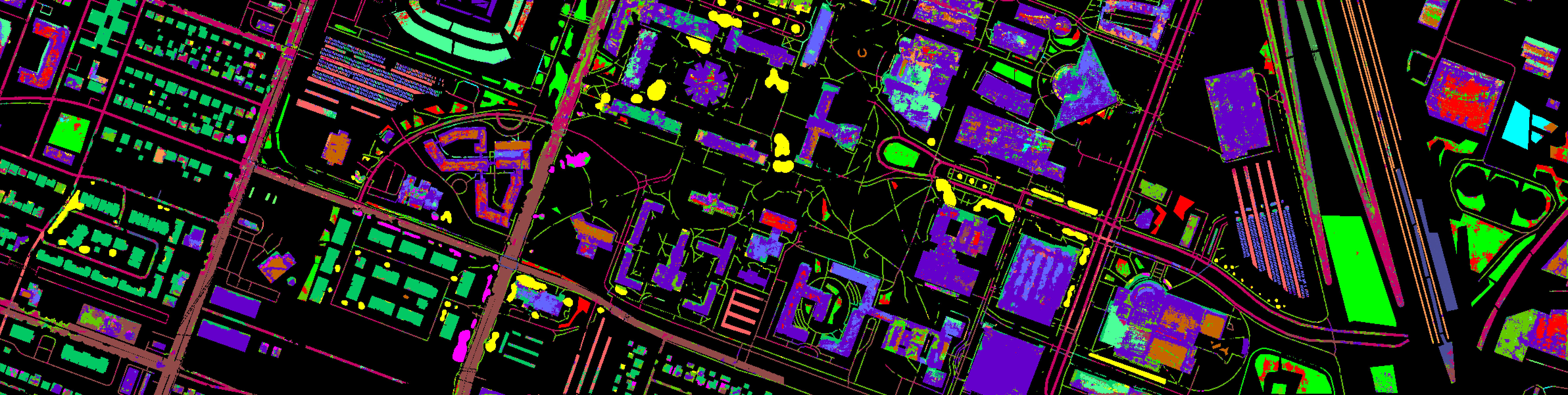}}
\hspace{1mm}
\subfloat[SepctralFormer]{\includegraphics[width=3.3in]{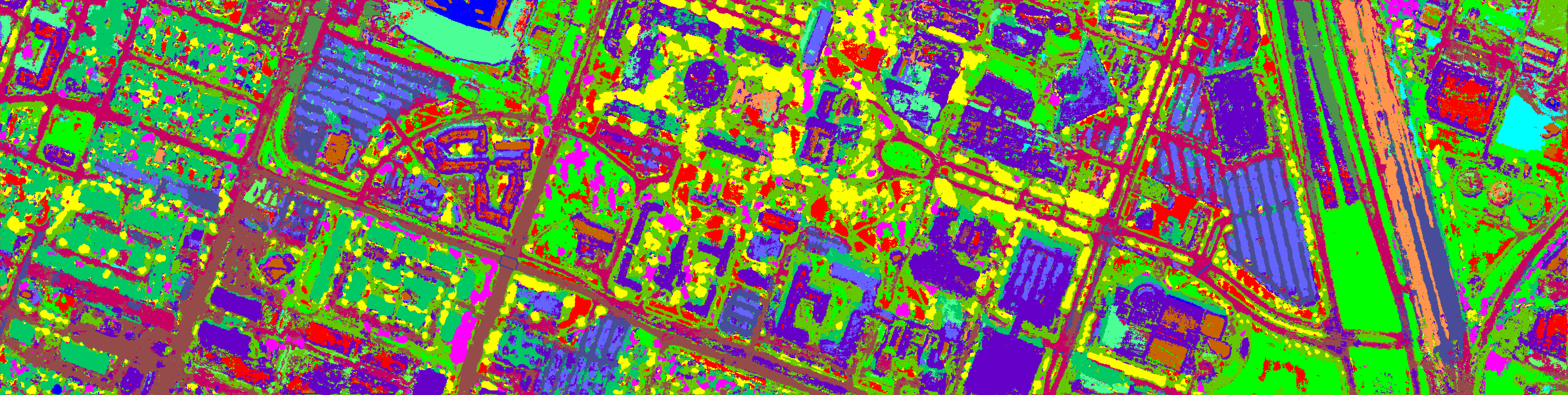}}
\hfil
\subfloat[XDCL]{\includegraphics[width=3.3in]{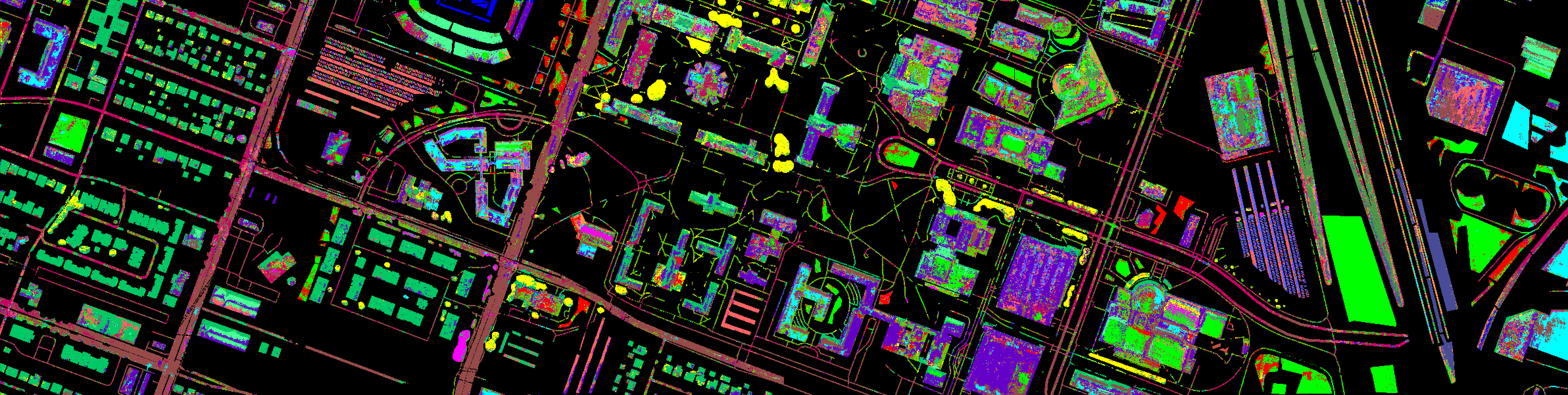}}
\hspace{1mm}
\subfloat[XDCL]{\includegraphics[width=3.3in]{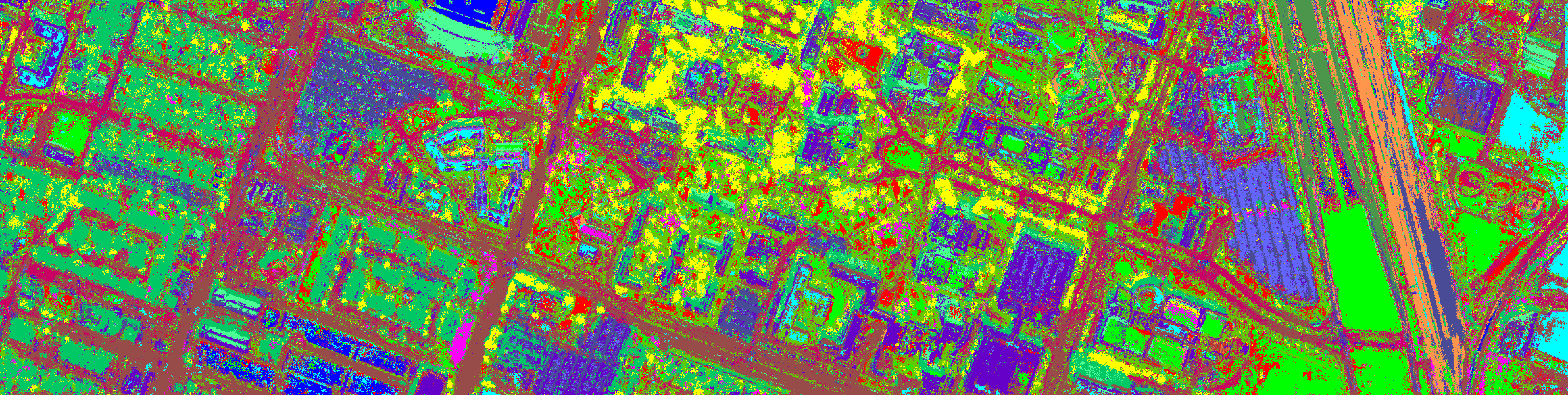}}
\hfil
\subfloat[KnowCL]{\includegraphics[width=3.3in]{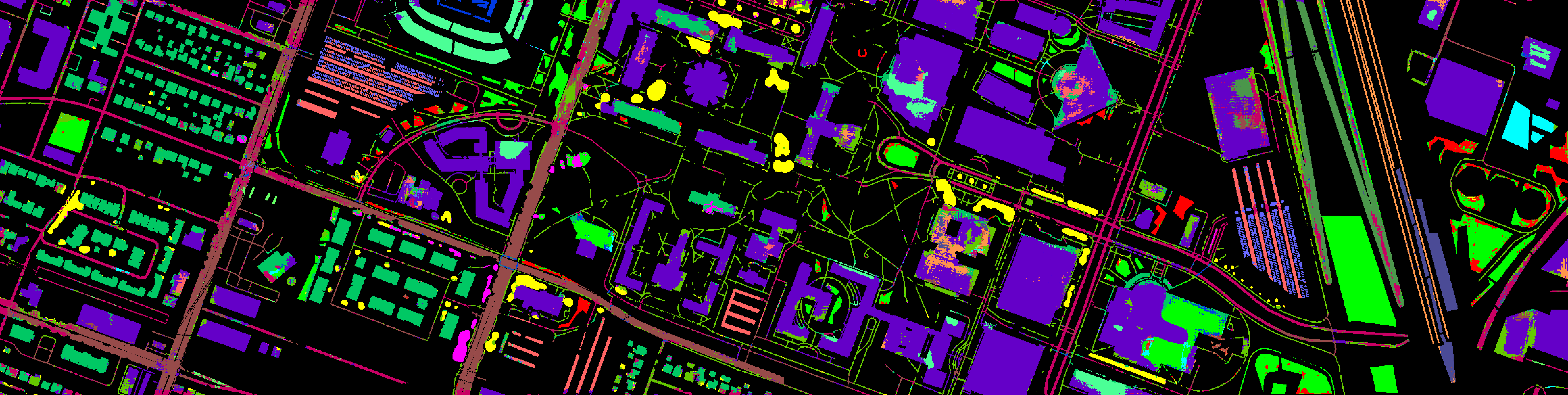}}
\hspace{1mm}
\subfloat[KnowCL]{\includegraphics[width=3.3in]{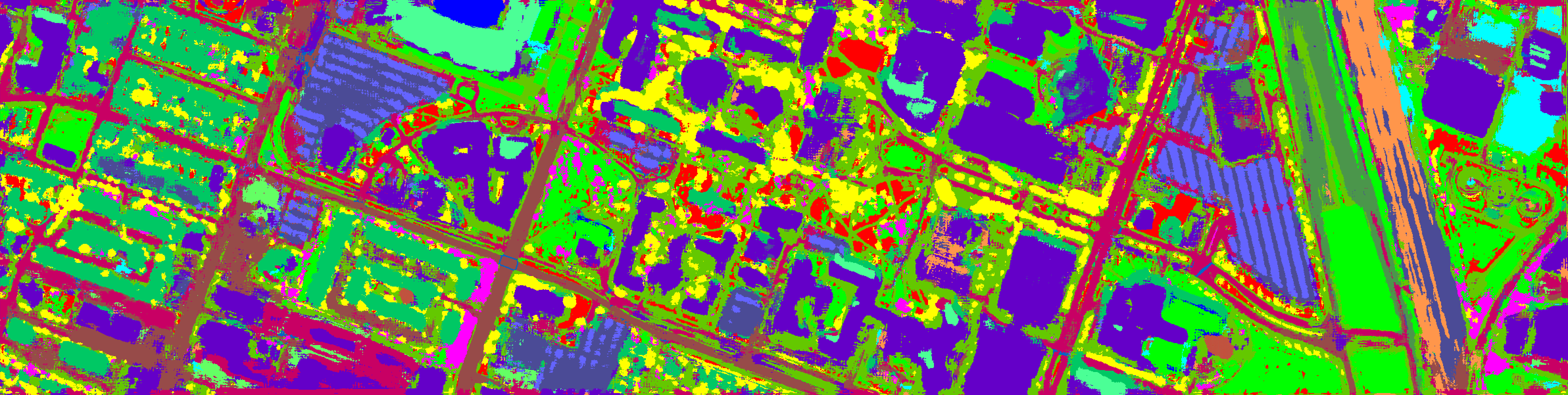}}
\caption{Ground truth (a), false-color (b), labeling area (c)-(i) and full image  (d)-(j) classification maps obtained by different models on the DFC2018.}
\label{E_houston}
\end{figure*}

% \subsection{Random vs Disjoint Comparison}
\begin{figure*}[!t]
\centering
\subfloat[HSI Patch]{\includegraphics[width=1.3in]{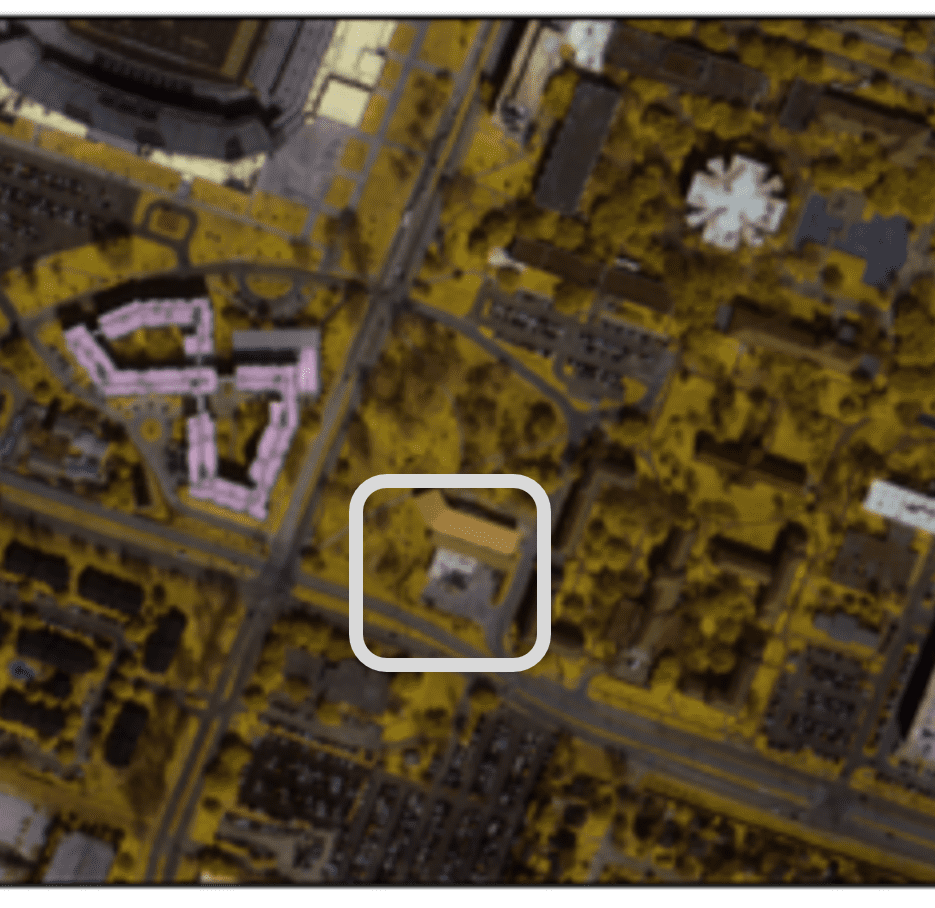}}
\hspace{2.5mm}
\subfloat[${\cal L}_{cl}$]{\includegraphics[width=1.3in]{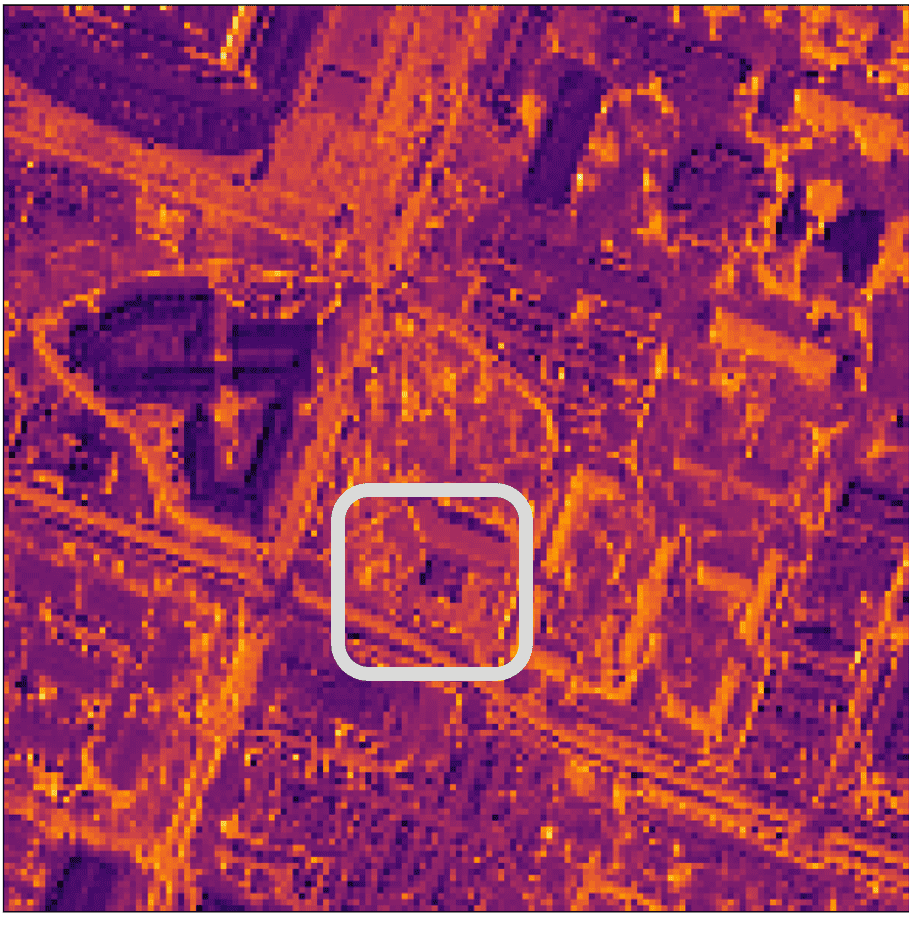}}
\hspace{2.5mm}
\subfloat[${\cal L}_{ce}$]{\includegraphics[width=1.3in]{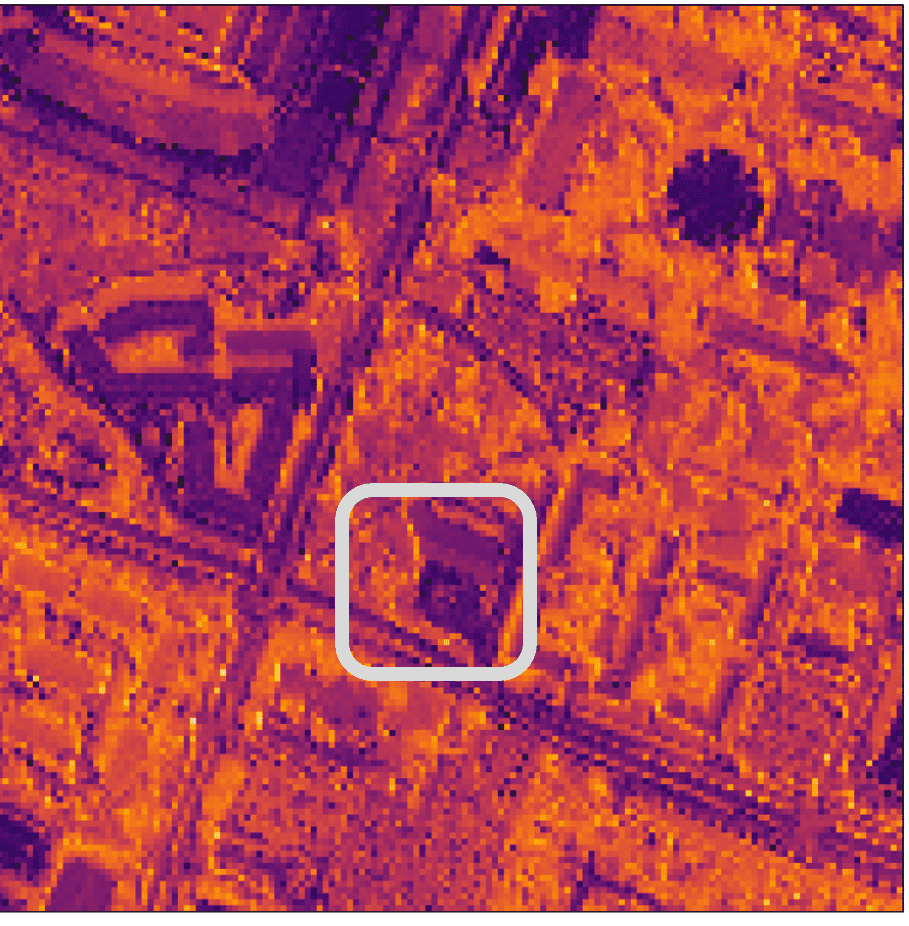}}
\hspace{2.5mm}
\subfloat[${\cal L}_{cl}+{\cal L}_{ce}$]{\includegraphics[width=1.3in]{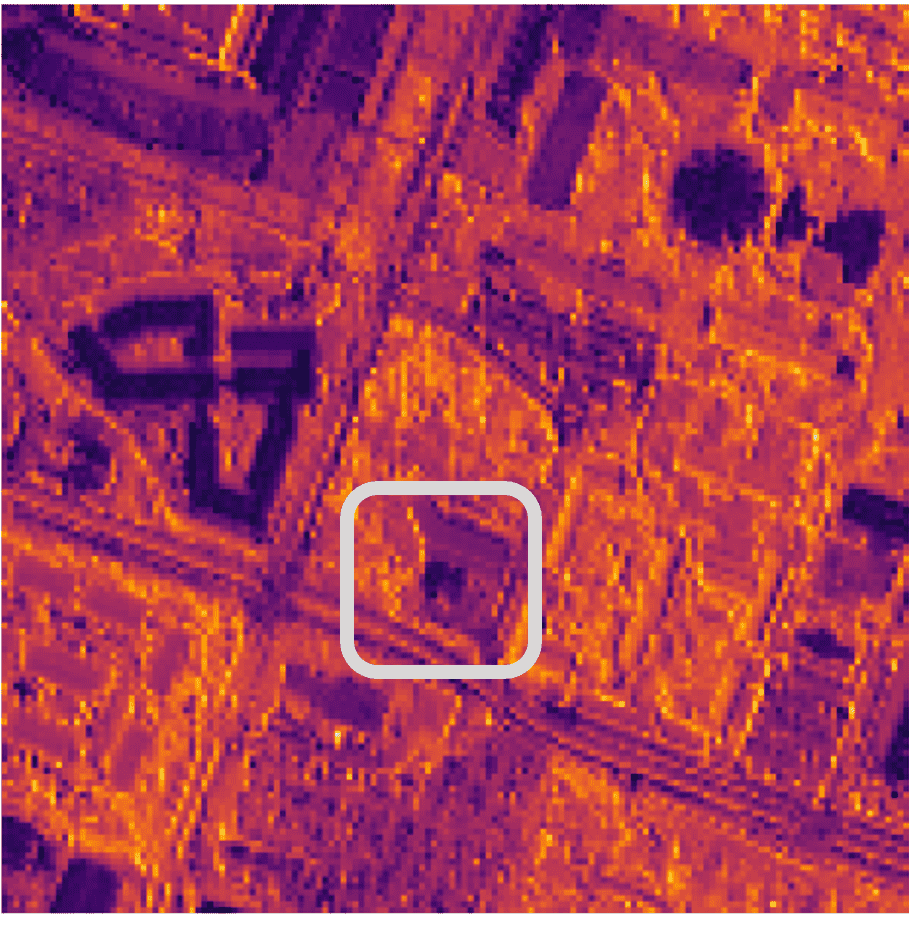}}
\hspace{2.5mm}
\subfloat[${\cal L}_{\cal T}$]{\includegraphics[width=1.3in]{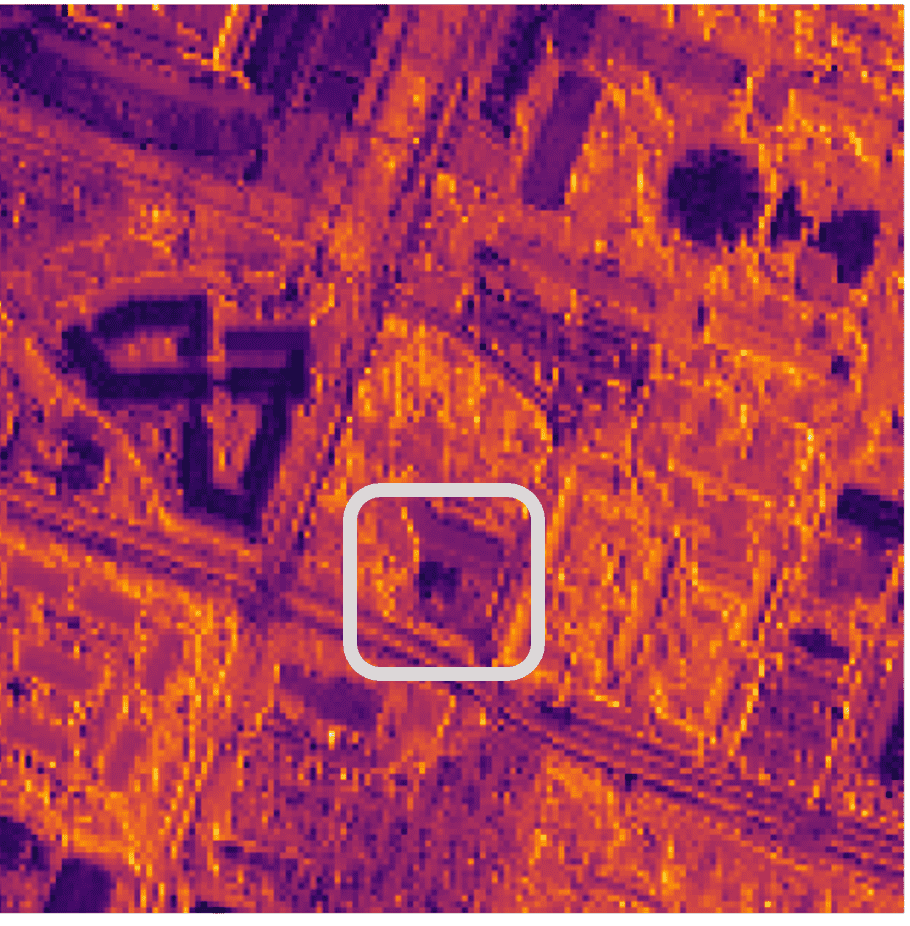}}
\hspace{2.5mm}
\caption{Visualization of different learning strategies output features obtained with ${\cal L}_{cl}$,  ${\cal L}_{ce}$,  ${\cal L}_{cl}+{\cal L}_{ce}$, and ${\cal L}_{\cal T}$.}
\label{fig_feature}
\end{figure*}

% The quantitative comparison with previous approaches on the UP, Salinas dataset, Dioni and DFC2018 are listed in Table 6, 7, 8 and 9, respectively.

The experimental results on UP are shown in Table \ref{tab:table6}. First of all, we can observe that the difference between the number of model parameters and FLOPs for HSI classification is very large. DMVL uses ResNet50 as the backbone, and its number of parameters is hundreds of times that of WFCG. Traditional semi-supervised learning inputs the whole image into memory, resulting in a dramatic increase in the FLOPs, tens or hundreds of times that of other algorithms. In terms of efficiency, semi-supervised learning is able to perform feature extraction on the whole image simultaneously, so it only takes a few tens of seconds to complete the training. Unsupervised learning, on the other hand, takes a lot of time for each epoch because it trains on every image block. Furthermore, the task-agnostic learning approach also results in poorly learned features for downstream classification tasks.

 % Constructing contrastive learning samples based on random sampling also significantly reduces the model training time. 
For supervised and unsupervised learning, KnowCL-SU and KnowCL-US achieve performance levels that are on par with or surpassing comparison methods, while semi-supervised learning KnowCL combines transformer and CL to achieve the best accuracy with balancing time and computation. Moreover, KnowCL is able to extract more fine-grained feature representations than hyperpixel-based graph learning models. Compared to the unsupervised learning model, KnowCL is able to learn task-specific feature representations using label knowledge. In addition, the performance of the linear evaluation is slightly lower than the KNN classifier, indicating that our model has fully utilized the label information in the training phase. As shown in Table \ref{tab:table17}, KnowCL also outperforms the other methods in all quality indices in Salinas. Salinas contains several crops that are uniformly distributed and easily distinguishable, and the other compared algorithms also achieve better classification results.

% Please add the following required packages to your document preamble:
% \usepackage{multirow}

\begin{table}
% \scriptsize
\tabcolsep=0.1mm  % 左右间距
\centering
\caption{Comparison under KNN, Supervied Head and Linear Classification Protocols Using Different Backbone Networks on four Datasets.}
\label{tab:table11}
\begin{tabular}{c|cccccc}
\hline
           & ViT\_hsi & ViT\_tiny     & ViT\_small & ResNet18 & ResNet50 & ResNet50*2 \\
\hline
Parameters & 534.11K  & 5.37M         & 21.43M     & 11.17M   & 23.50M   & 66.83M     \\
FLOPs      & 19.69M   & 198.23M       & 363.84M    & 423.00M  & 984.77M  & 2.78G      \\
kNN& 73.98& \textbf{74.01}         & 68.55      & 68.23    & 67.82    & 69.65      \\
Supervised head& 73.05& \textbf{74.36}         & 69.69      & 59.92    & 55.51    & 55.5       \\
Linear evaluation& 74.19    & 74.46         & 70.18      & 73.35    & 75.19    & \textbf{78.2}       \\
\hline
\end{tabular}
\end{table}

Table \ref{tab:table18} and \ref{tab:table19} show the experimental results on the Dioni and DFC2018. It is worth noting that the memory requirement of traditional semi-supervised learning methods increases exponentially with the amount of data. Conventional GPUs are no longer sufficient to support the operation of Dioni and DFC2018. Therefore, only supervised and unsupervised learning methods are used as comparison algorithms. Larger HSIs contain more diverse feature classes and produce more complex spectral combinations, leading to the accuracy of these two datasets having a large gap compared to UP and Salinas. It is in line with the performance of these algorithms when applied to the real world. We can also observe that the linear probe results are higher than the kNN classification results on the DFC2018. This can also be attributed to the complexity of the feature components of DFC2018. The large number of mixed pixels \cite{hong2018augmented} leads to poor classification performance. The best results obtained with KnowCL can substantially outperform comparative models. 

% \begin{figure*}[!t]
% \centering
% \subfloat[]{\includegraphics[width=1.0in]{GroundTruth/PaviaUtrain_gt2.png}}
% \hspace{3mm}
% \subfloat[]{\includegraphics[width=1.0in]{Experiment/PU/SVM-PaviaU0.7439776121733878a.png}}
% \hspace{3mm}
% \subfloat[]{\includegraphics[width=1.0in]{Experiment/PU/SPRN-PaviaU0.6908820438081226a.png}}
% \hspace{3mm}
% \subfloat[]{\includegraphics[width=1.0in]{Experiment/PU/FDGC-17.0366_knn_full.png}}
% \hspace{3mm}
% \subfloat[]{\includegraphics[width=1.0in]{Experiment/PU/SF-PaviaU0.png}}
% \hspace{3mm}
% \subfloat[]{\includegraphics[width=1.0in]{Experiment/PU/WFCG-PaviaU0.7901.png}}
% \hfil
% \subfloat[]{\includegraphics[width=1.02in]{GroundTruth/PaviaUtest_gt2.png}}
% \hspace{3mm}
% \subfloat[]{\includegraphics[width=1.0in]{Experiment/PU/EMSGCN-PaviaUyuan0.85637677.png}}
% \hspace{3mm}
% \subfloat[]{\includegraphics[width=1.0in]{Experiment/PU/VAE-PU0.6207875666209907a.png}}
% \hspace{3mm}
% \subfloat[]{\includegraphics[width=1.0in]{Experiment/PU/AAE-PU0.7233125646778643a.png}}
% \hspace{3mm}
% \subfloat[]{\includegraphics[width=1.0in]{Experiment/PU/XDCL-0.8226415065529022_linear_full.png}}
% \hspace{3mm}
% \subfloat[]{\includegraphics[width=1.0in]{Experiment/PU/KnowCL-19.4069_knn_full.png}}
% \caption{University of Pavia Experiments.}
% \label{pu1}
% \end{figure*}

\subsection{Visualization Comparison}

% \subsubsection{Visualizing the classification maps}

To further evaluate the classification quality of HSIs, we also examined visual perception. Due to space constraints, Figs. \ref{E_up}, \ref{E_sa}, \ref{E_Dinoni} and \ref{E_houston} show some of the algorithmic prediction results for the full image and labeled areas on the UP, Salinas, Dioni, and DFC2018, respectively. The classification results obtained by our method are more consistent with the ground truth when comparing Fig. \ref{E_up}(f) and (b)-(e), which demonstrates the effectiveness of the KnowCL.  Among the different methods, the edges of the classification map of KnowCL as shown in Fig. \ref{E_up}(l) are smoother than those of the other methods Fig. \ref{E_up}(h)-(k). This indicates that KnowCL can better capture finer spatial detail by the CL and embedded label knowledge. Although EMSGCN obtained clear classification maps by constructing superpixels, as shown in Fig \ref{E_sa}[j], this also led to more misclassification of neighboring features, such as Grapes vs. Vineyard and Fallow vs. Stubble. The result in Salinas of KnowCL contains less noise in the classification map, achieving a better visual performance. 

The same conclusion can be drawn from the visualisation results of Dioni and DFC2018 in Figs \ref{E_Dinoni} and \ref{E_houston}.  The Dioni dataset is derived from a spaceborne hyperspectral sensor, and its image quality is more susceptible to atmospheric and ground radiation compared to airborne sensors. Consequently, the classification result maps generated by many algorithms exhibit increased salt-and-pepper noise, as illustrated in Figs. \ref{E_Dinoni}(d), (f), (h). KnowCL, however, has the capability to mitigate the influence of local noise signals and produce smooth classification result maps, which is attributed to its training on global data. In the case of the DFC2018 dataset, where 1m resolution remote sensing images are acquired through an airborne hyperspectral sensor, the algorithm faces new challenges due to the complex urban environment. Notably, the limited expressive capacity of XDCL results in a substantial drop in classification accuracy on this dataset. In contrast, KnowCL, equipped with a transformer as the backbone, demonstrates strong expressive ability. The classification result aligns most closely with the genuine feature distribution on the ground.

% It is clear that the KnowCL is superior to the other methods in spatial detail when comparing Fig. 7 (I) and (h)-(k), which demonstrates the effectiveness of the KnowCL method. This can be attributed to the introduction of more global spatial context.

\subsection{Ablation Study}

We ablate the loss function to investigate the effect of its on the model. We compared our framework with the ${\cal L}_{cl}+{\cal L}_{ce}$  and ${\cal L}_{\cal T}$ loss function to reveal the effect of the proposed adaptive fusion loss function. As shown in Table  \ref{tab:table10}, when ${\cal L}_{\cal T}$ is used, the final classification results can be further improved in most cases. There is an improvement of 2.48\% and 6.09\% on the UP and Dioni, respectively. The relatively small boost on the Salinas is perhaps due to the fact that the two branches are quite different and do not complement each other. We observed a slight decrease in the accuracy of the model with adaptive fusion in DFC2018, which is possibly due to the accumulation of errors due to the lower accuracy before fusion.

% The reason is that the proposed loss function simultaneously considers the loss from the full image, while the existing methods often only consider the loss from the training data.

Fig.  \ref{fig_feature} visualizes the feature maps using supervised learning, unsupervised learning, additive fusion learning and adaptive fusion learning strategies, respectively. We selectively pick out some representative feature maps for visual comparison, where the visualization results of using the ${\cal L}_{cl}+{\cal L}_{ce}$ and  ${\cal L}_{\cal T}$ learning strategies have finer appearance than those using only one loss function.  For example, the activation maps for the region in the white box show that the area of the non-residential building based on the fusion loss has sharper edges, object contours, and texture structures. This also visually confirms the superiority of the designed framework.

\begin{table}
\centering
\caption{Quantitative Evaluation of Different Learning Strategies on the Four Datasets.}
\label{tab:table10}
\begin{tabular}{c|cccc}
\hline
& UP      & Salinas & Dioni   & DFC2018 \\
\hline
${\cal L}_{cl}+{\cal L}_{ce}$& 88.88  & 93.42  & 76.43  & \textbf{75.94}  \\
\hline
 ${\cal L}_{\cal T}$ & \textbf{91.36}      & \textbf{93.62}      & \textbf{82.52}      &73.98      \\
\hline
\end{tabular}
\end{table}

\subsection{Backbones and classifier}

There are numerous models for HSI classification at present. We try to propose a general framework that is compatible with the mainstream deep learning backbone. Table \ref{tab:table11} shows the performance of three ViT-based networks and three ResNet-based networks as well as their number of parameters under kNN, supervised head and linear classification protocols. Fig. \ref{fig_1} provides a more visual comparison of them with the other models in DFC2018. We find that the ViT-based networks have similar classification performance under the three classification strategies, while the accuracy of the ResNet-based networks becomes higher at linear evaluation as the model parameters increase. We surmise that this is because ViT is able to extract global features at each layer, so the model performance converges faster. ResNet has the local receptive fields and requires a deeper structure to adequately extract global features. Moreover, our model achieves a large advantage in all parametric quantities.

\section{Conclusion}
\label{conclusion}

HSI classification techniques have been developed over a long period. A range of models have now been developed. However, the quality of labels and the learning approach result in low accuracy of supervised and unsupervised learning methods in the real world. Traditional semi-supervised learning methods can utilize both labeled and unlabeled samples, but they lack the scalability to perform large-scale HSI classification tasks. 

In this paper, we present a universal HSI classification framework designed for addressing real-world HSI classification problems, which is compatible with supervised learning, unsupervised learning and semi-supervised learning. Among them, the semi-supervised learning classification KnowCL is newly defined to fit the requirements of real applications. An efficient contrastive learning strategy is constructed by means of labeled knowledge embedding, where the parameters, FLOPs and execution time are comparable to that of supervised learning but can fully extract features from unlabeled samples. We conduct four disjoint datasets, and a large number of experiments on them quantitatively and intuitively validate the superiority of the proposed model. Future, it is possible to adopt a pre-trained visual foundation model for HSI classification, which will greatly advance the application of HSIs in the real world.

% \section*{Acknowledgments}
% This should be a simple paragraph before the References to thank those individuals and institutions who have supported your work on this article.

% % appendix
% {\appendix[Proof of the Zonklar Equations]
% Use $\backslash${\tt{appendix}} if you have a single appendix:
% Do not use $\backslash${\tt{section}} anymore after $\backslash${\tt{appendix}}, only $\backslash${\tt{section*}}.
% If you have multiple appendixes use $\backslash${\tt{appendices}} then use $\backslash${\tt{section}} to start each appendix.
% You must declare a $\backslash${\tt{section}} before using any $\backslash${\tt{subsection}} or using $\backslash${\tt{label}} ($\backslash${\tt{appendices}} by itself
%  starts a section numbered zero.)}

% Reference
% --------------------------------------------------------------
%     You don't have to mess with anything below this line.
% --------------------------------------------------------------
\bibliographystyle{unsrt}
\bibliography{reference}

\begin{thebibliography}{10}

\bibitem{shimoni2019hypersectral}
Michal Shimoni, Rob Haelterman, and Christiaan Perneel.
\newblock Hypersectral imaging for military and security applications: Combining myriad processing and sensing techniques.
\newblock {\em IEEE Geoscience and Remote Sensing Magazine}, 7(2):101--117, 2019.

\bibitem{al2021spectral}
Suhad~Lateef Al-Khafaji, Jun Zhou, Xiao Bai, Yuntao Qian, and Alan Wee-Chung Liew.
\newblock Spectral-spatial boundary detection in hyperspectral images.
\newblock {\em IEEE Transactions on Image Processing}, 31:499--512, 2021.

\bibitem{dian2023zero}
Renwei Dian, Anjing Guo, and Shutao Li.
\newblock Zero-shot hyperspectral sharpening.
\newblock {\em IEEE Transactions on Pattern Analysis and Machine Intelligence}, 2023.

\bibitem{wu2023fully}
Chen Wu, Bo~Du, and Liangpei Zhang.
\newblock Fully convolutional change detection framework with generative adversarial network for unsupervised, weakly supervised and regional supervised change detection.
\newblock {\em IEEE Transactions on Pattern Analysis and Machine Intelligence}, 2023.

\bibitem{gao2022unsupervised}
Kuiliang Gao, Bing Liu, Xuchu Yu, and Anzhu Yu.
\newblock Unsupervised meta learning with multiview constraints for hyperspectral image small sample set classification.
\newblock {\em IEEE Transactions on Image Processing}, 31:3449--3462, 2022.

\bibitem{sohn2002supervised}
Youngsinn Sohn and N~Sanjay Rebello.
\newblock Supervised and unsupervised spectral angle classifiers.
\newblock {\em Photogrammetric Engineering and Remote Sensing}, 68(12):1271--1282, 2002.

\bibitem{benediktsson2005classification}
J{\'o}n~Atli Benediktsson, J{\'o}n~Aevar Palmason, and Johannes~R Sveinsson.
\newblock Classification of hyperspectral data from urban areas based on extended morphological profiles.
\newblock {\em IEEE Transactions on Geoscience and Remote Sensing}, 43(3):480--491, 2005.

\bibitem{huang2016remote}
Longhui Huang, Chen Chen, Wei Li, and Qian Du.
\newblock Remote sensing image scene classification using multi-scale completed local binary patterns and fisher vectors.
\newblock {\em Remote Sensing}, 8(6):483, 2016.

\bibitem{cao2018hyperspectral}
Xiangyong Cao, Feng Zhou, Lin Xu, Deyu Meng, Zongben Xu, and John Paisley.
\newblock Hyperspectral image classification with markov random fields and a convolutional neural network.
\newblock {\em IEEE Transactions on Image Processing}, 27(5):2354--2367, 2018.

\bibitem{gu2020semi}
Xiaowei Gu, Plamen~P Angelov, Ce~Zhang, and Peter~M Atkinson.
\newblock A semi-supervised deep rule-based approach for complex satellite sensor image analysis.
\newblock {\em IEEE Transactions on Pattern Analysis and Machine Intelligence}, 44(5):2281--2292, 2020.

\bibitem{guo2022deep}
Anjing Guo, Renwei Dian, and Shutao Li.
\newblock A deep framework for hyperspectral image fusion between different satellites.
\newblock {\em IEEE Transactions on Pattern Analysis and Machine Intelligence}, 2022.

\bibitem{hemae}
Kaiming He, Xinlei Chen, Saining Xie, Yanghao Li, Piotr Dollár, and Ross Girshick.
\newblock Masked autoencoders are scalable vision learners.
\newblock In {\em 2022 IEEE/CVF Conference on Computer Vision and Pattern Recognition (CVPR)}, pages 15979--15988, 2022.

\bibitem{zheng2023farseg++}
Zhuo Zheng, Yanfei Zhong, Junjue Wang, Ailong Ma, and Liangpei Zhang.
\newblock Farseg++: Foreground-aware relation network for geospatial object segmentation in high spatial resolution remote sensing imagery.
\newblock {\em IEEE Transactions on Pattern Analysis and Machine Intelligence}, 2023.

\bibitem{zhao2023exploring}
Yunqing Zhao, Chao Du, Milad Abdollahzadeh, Tianyu Pang, Min Lin, Shuicheng Yan, and Ngai-Man Cheung.
\newblock Exploring incompatible knowledge transfer in few-shot image generation.
\newblock In {\em 2023 IEEE/CVF Conference on Computer Vision and Pattern Recognition (CVPR)}, pages 7380--7391, 2023.

\bibitem{chen2020big}
Ting Chen, Simon Kornblith, Kevin Swersky, Mohammad Norouzi, and Geoffrey~E Hinton.
\newblock Big self-supervised models are strong semi-supervised learners.
\newblock {\em Advances in Neural Information Processing Systems}, 33:22243--22255, 2020.

\bibitem{chen2021exploring}
Xinlei Chen and Kaiming He.
\newblock Exploring simple siamese representation learning.
\newblock In {\em 2021 IEEE/CVF Conference on Computer Vision and Pattern Recognition (CVPR)}, pages 15745--15753, 2021.

\bibitem{wu2017semi}
Hao Wu and Saurabh Prasad.
\newblock Semi-supervised deep learning using pseudo labels for hyperspectral image classification.
\newblock {\em IEEE Transactions on Image Processing}, 27(3):1259--1270, 2017.

\bibitem{dong2022weighted}
Yanni Dong, Quanwei Liu, Bo~Du, and Liangpei Zhang.
\newblock Weighted feature fusion of convolutional neural network and graph attention network for hyperspectral image classification.
\newblock {\em IEEE Transactions on Image Processing}, 31:1559--1572, 2022.

\bibitem{liu2020cnn}
Qichao Liu, Liang Xiao, Jingxiang Yang, and Zhihui Wei.
\newblock Cnn-enhanced graph convolutional network with pixel-and superpixel-level feature fusion for hyperspectral image classification.
\newblock {\em IEEE Transactions on Geoscience and Remote Sensing}, 59(10):8657--8671, 2020.

\bibitem{9785802}
Quanwei Liu, Yanni Dong, Yuxiang Zhang, and Hui Luo.
\newblock A fast dynamic graph convolutional network and cnn parallel network for hyperspectral image classification.
\newblock {\em IEEE Transactions on Geoscience and Remote Sensing}, 60:1--15, 2022.

\bibitem{audebert2019deep}
Nicolas Audebert, Bertrand Le~Saux, and S{\'e}bastien Lef{\`e}vre.
\newblock Deep learning for classification of hyperspectral data: A comparative review.
\newblock {\em IEEE Geoscience and Remote Sensing Magazine}, 7(2):159--173, 2019.

\bibitem{liang2016sampling}
Jie Liang, Jun Zhou, Yuntao Qian, Lian Wen, Xiao Bai, and Yongsheng Gao.
\newblock On the sampling strategy for evaluation of spectral-spatial methods in hyperspectral image classification.
\newblock {\em IEEE Transactions on Geoscience and Remote Sensing}, 55(2):862--880, 2016.

\bibitem{friedl2000note}
MA~Friedl, C~Woodcock, S~Gopal, D~Muchoney, AH~Strahler, and C~Barker-Schaaf.
\newblock A note on procedures used for accuracy assessment in land cover maps derived from avhrr data.
\newblock {\em IEEE Transactions on Geoscience and Remote Sensing}, 2000.

\bibitem{hong2021spectralformer}
Danfeng Hong, Zhu Han, Jing Yao, Lianru Gao, Bing Zhang, Antonio Plaza, and Jocelyn Chanussot.
\newblock Spectralformer: Rethinking hyperspectral image classification with transformers.
\newblock {\em IEEE Transactions on Geoscience and Remote Sensing}, 60:1--15, 2021.

\bibitem{zheng2020fpga}
Zhuo Zheng, Yanfei Zhong, Ailong Ma, and Liangpei Zhang.
\newblock Fpga: Fast patch-free global learning framework for fully end-to-end hyperspectral image classification.
\newblock {\em IEEE Transactions on Geoscience and Remote Sensing}, 58(8):5612--5626, 2020.

\bibitem{zhang2021spectral}
Xiangrong Zhang, Shouwang Shang, Xu~Tang, Jie Feng, and Licheng Jiao.
\newblock Spectral partitioning residual network with spatial attention mechanism for hyperspectral image classification.
\newblock {\em IEEE Transactions on Geoscience and Remote Sensing}, 60:1--14, 2021.

\bibitem{zhang2022cross}
Suhua Zhang, Zhikui Chen, Dan Wang, and Z~Jane Wang.
\newblock Cross-domain few-shot contrastive learning for hyperspectral images classification.
\newblock {\em IEEE Geoscience and Remote Sensing Letters}, 19:1--5, 2022.

\bibitem{cao2021nonoverlapped}
Xianghai Cao, Zuji Liu, Xiangxiang Li, Qian Xiao, Jie Feng, and Licheng Jiao.
\newblock Nonoverlapped sampling for hyperspectral imagery: Performance evaluation and a cotraining-based classification strategy.
\newblock {\em IEEE Transactions on Geoscience and Remote Sensing}, 60:1--14, 2021.

\bibitem{sun2021supervised}
Hao Sun, Xiangtao Zheng, and Xiaoqiang Lu.
\newblock A supervised segmentation network for hyperspectral image classification.
\newblock {\em IEEE Transactions on Image Processing}, 30:2810--2825, 2021.

\bibitem{zhao2022superpixel}
Chunhui Zhao, Wenxiang Zhu, and Shou Feng.
\newblock Superpixel guided deformable convolution network for hyperspectral image classification.
\newblock {\em IEEE Transactions on Image Processing}, 31:3838--3851, 2022.

\bibitem{li2018independently}
Shuai Li, Wanqing Li, Chris Cook, Ce~Zhu, and Yanbo Gao.
\newblock Independently recurrent neural network (indrnn): Building a longer and deeper rnn.
\newblock In {\em 2018 IEEE/CVF Conference on Computer Vision and Pattern Recognition (CVPR)}, pages 5457--5466, 2018.

\bibitem{he2016deep}
Kaiming He, Xiangyu Zhang, Shaoqing Ren, and Jian Sun.
\newblock Deep residual learning for image recognition.
\newblock In {\em 2016 IEEE/CVF Conference on Computer Vision and Pattern Recognition (CVPR)}, pages 770--778, 2016.

\bibitem{liu2021swin}
Ze~Liu, Yutong Lin, Yue Cao, Han Hu, Yixuan Wei, Zheng Zhang, Stephen Lin, and Baining Guo.
\newblock Swin transformer: Hierarchical vision transformer using shifted windows.
\newblock In {\em 2021 IEEE/CVF Conference on Computer Vision and Pattern Recognition (CVPR)}, pages 10012--10022, 2021.

\bibitem{fu2021coded}
Ying Fu, Tao Zhang, Lizhi Wang, and Hua Huang.
\newblock Coded hyperspectral image reconstruction using deep external and internal learning.
\newblock {\em IEEE Transactions on Pattern Analysis and Machine Intelligence}, 44(7):3404--3420, 2021.

\bibitem{li2019dff}
Xian Li, Mingli Ding, and Aleksandra Pi{\v{z}}urica.
\newblock Deep feature fusion via two-stream convolutional neural network for hyperspectral image classification.
\newblock {\em IEEE Transactions on Geoscience and Remote Sensing}, 58(4):2615--2629, 2019.

\bibitem{vaswani2017attention}
Ashish Vaswani, Noam Shazeer, Niki Parmar, Jakob Uszkoreit, Llion Jones, Aidan~N Gomez, {\L}ukasz Kaiser, and Illia Polosukhin.
\newblock Attention is all you need.
\newblock {\em Advances in Neural Information Processing Systems}, 30, 2017.

\bibitem{zheng2023label}
Qinghai Zheng, Jihua Zhu, and Haoyu Tang.
\newblock Label information bottleneck for label enhancement.
\newblock In {\em 2023 IEEE/CVF Conference on Computer Vision and Pattern Recognition (CVPR)}, pages 7497--7506, 2023.

\bibitem{cao2020unsupervised}
Zeyu Cao, Xiaorun Li, and Liaoying Zhao.
\newblock Unsupervised feature learning by autoencoder and prototypical contrastive learning for hyperspectral classification.
\newblock {\em arXiv preprint arXiv:2009.00953}, 2020.

\bibitem{he2020momentum}
Kaiming He, Haoqi Fan, Yuxin Wu, Saining Xie, and Ross Girshick.
\newblock Momentum contrast for unsupervised visual representation learning.
\newblock In {\em 2020 IEEE/CVF Conference on Computer Vision and Pattern Recognition (CVPR)}, pages 9726--9735, 2020.

\bibitem{grill2020bootstrap}
Jean-Bastien Grill, Florian Strub, Florent Altch{\'e}, Corentin Tallec, Pierre Richemond, Elena Buchatskaya, Carl Doersch, Bernardo Avila~Pires, Zhaohan Guo, Mohammad Gheshlaghi~Azar, et~al.
\newblock Bootstrap your own latent-a new approach to self-supervised learning.
\newblock {\em Advances in Neural Information Processing Systems}, 33:21271--21284, 2020.

\bibitem{caron2021emerging}
Mathilde Caron, Hugo Touvron, Ishan Misra, Hervé Jegou, Julien Mairal, Piotr Bojanowski, and Armand Joulin.
\newblock Emerging properties in self-supervised vision transformers.
\newblock In {\em 2021 IEEE/CVF International Conference on Computer Vision (ICCV)}, pages 9630--9640, 2021.

\bibitem{chen2020simple}
Ting Chen, Simon Kornblith, Mohammad Norouzi, and Geoffrey Hinton.
\newblock A simple framework for contrastive learning of visual representations.
\newblock In {\em International Conference on Machine Learning}, pages 1597--1607. PMLR, 2020.

\bibitem{oord2018representation}
Aaron van~den Oord, Yazhe Li, and Oriol Vinyals.
\newblock Representation learning with contrastive predictive coding.
\newblock {\em arXiv preprint arXiv:1807.03748}, 2018.

\bibitem{tian2020makes}
Yonglong Tian, Chen Sun, Ben Poole, Dilip Krishnan, Cordelia Schmid, and Phillip Isola.
\newblock What makes for good views for contrastive learning?
\newblock {\em Advances in Neural Information Processing systems}, 33:6827--6839, 2020.

\bibitem{liu2020deep}
Bing Liu, Anzhu Yu, Xuchu Yu, Ruirui Wang, Kuiliang Gao, and Wenyue Guo.
\newblock Deep multiview learning for hyperspectral image classification.
\newblock {\em IEEE Transactions on Geoscience and Remote Sensing}, 59(9):7758--7772, 2020.

\bibitem{pan2018mugnet}
Bin Pan, Zhenwei Shi, and Xia Xu.
\newblock Mugnet: Deep learning for hyperspectral image classification using limited samples.
\newblock {\em ISPRS Journal of Photogrammetry and Remote Sensing}, 145:108--119, 2018.

\bibitem{liu2021patch}
Bing Liu and Xuchu Yu.
\newblock Patch-free bilateral network for hyperspectral image classification using limited samples.
\newblock {\em IEEE Journal of Selected Topics in Applied Earth Observations and Remote Sensing}, 14:10794--10807, 2021.

\bibitem{9745164}
Hongyan Zhang, Jiaqi Zou, and Liangpei Zhang.
\newblock Ems-gcn: An end-to-end mixhop superpixel-based graph convolutional network for hyperspectral image classification.
\newblock {\em IEEE Transactions on Geoscience and Remote Sensing}, 60:1--16, 2022.

\bibitem{liebel2018auxiliary}
Lukas Liebel and Marco K{\"o}rner.
\newblock Auxiliary tasks in multi-task learning.
\newblock {\em arXiv preprint arXiv:1805.06334}, 2018.

\bibitem{bachman2019learning}
Philip Bachman, R~Devon Hjelm, and William Buchwalter.
\newblock Learning representations by maximizing mutual information across views.
\newblock {\em Advances in Neural Information Processing Systems}, 32, 2019.

\bibitem{ye2019unsupervised}
Mang Ye, Xu~Zhang, Pong~C. Yuen, and Shih-Fu Chang.
\newblock Unsupervised embedding learning via invariant and spreading instance feature.
\newblock In {\em 2019 IEEE/CVF Conference on Computer Vision and Pattern Recognition (CVPR)}, pages 6203--6212, 2019.

\bibitem{wu2018unsupervised}
Zhirong Wu, Yuanjun Xiong, Stella~X. Yu, and Dahua Lin.
\newblock Unsupervised feature learning via non-parametric instance discrimination.
\newblock In {\em 2018 IEEE/CVF Conference on Computer Vision and Pattern Recognition (CVPR)}, pages 3733--3742, 2018.

\bibitem{dong2018hyperparameter}
Xingping Dong, Jianbing Shen, Wenguan Wang, Yu~Liu, Ling Shao, and Fatih Porikli.
\newblock Hyperparameter optimization for tracking with continuous deep q-learning.
\newblock In {\em 2018 IEEE/CVF Conference on Computer Vision and Pattern Recognition (CVPR)}, pages 518--527, 2018.

\bibitem{hong2018augmented}
Danfeng Hong, Naoto Yokoya, Jocelyn Chanussot, and Xiao~Xiang Zhu.
\newblock An augmented linear mixing model to address spectral variability for hyperspectral unmixing.
\newblock {\em IEEE Transactions on Image Processing}, 28(4):1923--1938, 2018.

\end{thebibliography}

\vspace{11pt}

\vspace{11pt}

\vfill

\end{document}